\documentclass[sigconf, nonacm]{acmart}
\usepackage{xspace}
\usepackage{enumerate}
\usepackage[ruled]{algorithm2e}
\usepackage{adjustbox}
\usepackage{multirow}
\usepackage{subfigure}
\usepackage{color}
\usepackage[table]{xcolor}
\usepackage{soul}
\usepackage{makecell}
\usepackage{balance}
\usepackage{setspace}
\usepackage{microtype}
\definecolor{MyRed}{RGB}{151,0,0}
\definecolor{Yellow}{RGB}{243,140,10}

\newcommand{\model}{\texttt{SCISE}\xspace}

\newcommand\vldbdoi{XX.XX/XXX.XX}
\newcommand\vldbpages{XXX-XXX}
\newcommand\vldbvolume{19}
\newcommand\vldbissue{1}
\newcommand\vldbyear{2026}
\newcommand\vldbauthors{\authors}
\newcommand\vldbtitle{\shorttitle} 
\newcommand\vldbavailabilityurl{URL_TO_YOUR_ARTIFACTS}
\newcommand\vldbpagestyle{plain} 

\begin{document}
\title{Breaking Structural Isolation: Scalable Graph Clustering via Community-Aware Sampling and Structural Entropy}


\author{Jingyun Zhang}
\orcid{0009-0003-7215-0040}
\affiliation{%
  \institution{Beihang University}
  \state{Beijing}
  \country{China}
}
\email{zhangjingyun@buaa.edu.cn}

\author{Hao Peng}
\authornote{Corresponding author.}
\orcid{0000-0003-0458-5977}
\affiliation{%
  \institution{Beihang University}
  \state{Beijing}
  \country{China}
}
\affiliation{%
  \institution{Hangzhou Innovation Institute, Beihang University}
  \state{Hangzhou}
  \country{China}
}
\email{penghao@buaa.edu.cn}

\author{Jianxin Li}
\affiliation{%
  \institution{Beihang University}
  \state{Beijing}
  \country{China}
}
\email{lijx@buaa.edu.cn}

\author{Angsheng Li}
\affiliation{%
  \institution{Beihang University}
  \city{Beijing}
  \country{China}
 }
\email{angsheng@buaa.edu.cn}

\author{Philip S. Yu}
\affiliation{%
  \institution{University of Illinois at Chicago}
  \city{Chicago}
  \country{USA}
 }
\email{psyu@uic.edu}

\newif\ifshowrevised
\showrevisedfalse

\ifshowrevised
  \newcommand{\revised}[1]{{\color{blue}#1}}
\else
  \newcommand{\revised}[1]{#1}
\fi

\begin{abstract}
Unsupervised graph clustering is a fundamental technique for uncovering underlying semantic patterns in large-scale networks.
Although Graph Contrastive Learning has demonstrated promising performance, existing methods often suffer from the "structural isolation" issue during mini-batch training, making it challenging to capture cohesive community structures that characterize the global topological distribution.
To address these challenges, we propose\textbf{~\model{}}, a \underline{\textbf{S}}calable unsupervised graph \underline{\textbf{C}}lustering framework that preserves structural \underline{\textbf{I}}ntegrity by synergizing community-aware sampling with constrained \underline{\textbf{S}}tructural \underline{\textbf{E}}ntropy.
Specifically, we first introduce the Structural Entropy Community Constraint operator (SECC), which optimizes structural information within a constrained solution space to mitigate community fragmentation and enhance partition cohesion. 
Second, to prevent global information loss during batch training, we design a Community-Aware Sampling Expansion (CSampE) mechanism that incorporates the community context of target nodes into sampling batches, effectively breaking structural barriers and preserving topological integrity. 
Finally, we devise a Structural Contrastive Learning (StructCL) module that refines edge weights based on intra-batch structural similarity, guiding the encoder to learn representations in a higher-order structural space.
Extensive experiments on six mainstream benchmark datasets demonstrate that ~\model{} significantly outperforms state-of-the-art algorithms, with ablation studies and robustness analyses further validating its effectiveness and reliability for real-world large-scale graphs.
\end{abstract}

\maketitle

\pagestyle{\vldbpagestyle}
\begingroup\small\noindent\raggedright\textbf{PVLDB Reference Format:}\\
\vldbauthors. \vldbtitle. PVLDB, \vldbvolume(\vldbissue): \vldbpages, \vldbyear.\\
\href{https://doi.org/\vldbdoi}{doi:\vldbdoi}
\endgroup
\begingroup
\renewcommand\thefootnote{}\footnote{\noindent
This work is licensed under the Creative Commons BY-NC-ND 4.0 International License. Visit \url{https://creativecommons.org/licenses/by-nc-nd/4.0/} to view a copy of this license. For any use beyond those covered by this license, obtain permission by emailing \href{mailto:info@vldb.org}{info@vldb.org}. Copyright is held by the owner/author(s). Publication rights licensed to the VLDB Endowment. \\
\raggedright Proceedings of the VLDB Endowment, Vol. \vldbvolume, No. \vldbissue\ %
ISSN 2150-8097. \\
\href{https://doi.org/\vldbdoi}{doi:\vldbdoi} \\
}\addtocounter{footnote}{-1}\endgroup

\ifdefempty{\vldbavailabilityurl}{}{
\vspace{.3cm}
\begingroup\small\noindent\raggedright\textbf{PVLDB Artifact Availability:}\\
The source code, data, and/or other artifacts have been made available at \url{https://github.com/SELGroup/SCISE}.
\endgroup
}

\vspace{-3mm}
\section{Introduction}
\label{sec: introduction}

Graph-structured data has become pervasive in representing complex relational information across diverse domains, such as social media, citation networks, and biological systems ~\cite{hamilton2017inductive, Thomas2017Semi}. 
As a fundamental task in network analysis, unsupervised graph clustering aims to partition nodes with high topological and attribute affinities into communities without reliance on manual labeling, thereby revealing the underlying semantic patterns and structural motifs within complex networks~\cite{fortunato2010community, xie2016unsupervised}. 
This capability is critical for numerous downstream applications, including community detection, targeted recommendation, and anomaly discovery~\cite{ying2018graph,akoglu2015graph}. 
Given the exponential growth of graph data and the prohibitive cost of human annotation, developing effective and scalable unsupervised clustering frameworks is of paramount importance for extracting meaningful structural insights from unlabeled, large-scale relational data~\cite{wu2020comprehensive,velivckovic2018deep}.\par

\begin{figure}[h]
	\centering
	\includegraphics[width=0.92\linewidth]{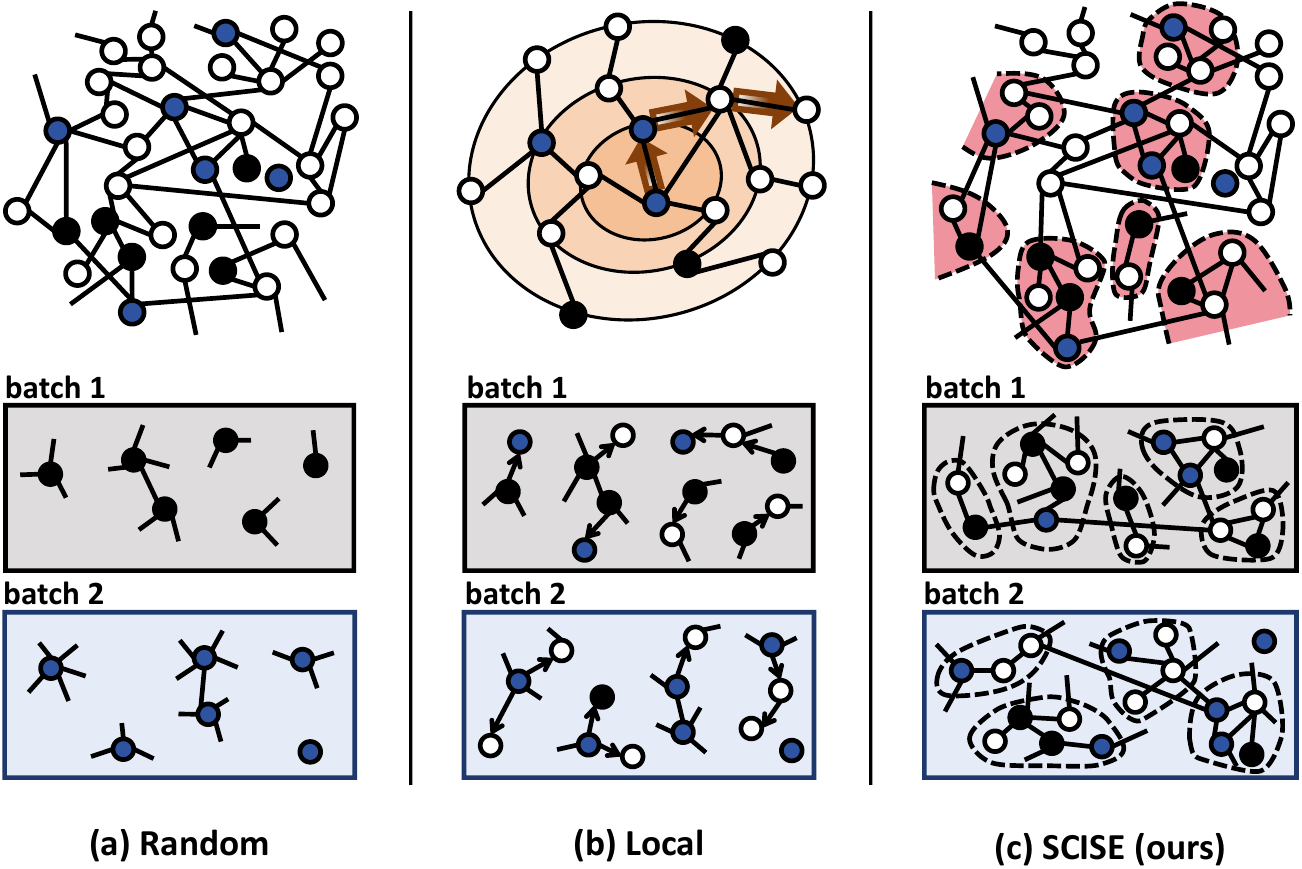}
    \vspace{-4mm}
	\caption{Comparison of sampling strategies. (a) Random: fragments communities into batches. (b) Local: limited to neighborhood context. (c) \model{} (Ours): incorporates global community context to preserve structural integrity within mini-batches.}
    \label{fig: toy}
    \vspace{-7mm}
\end{figure}

The prevailing paradigm for unsupervised graph clustering has shifted towards integrating Graph Neural Networks (GNNs) with Graph Contrastive Learning (GCL) to learn discriminative semantic embeddings~\cite{bo2020structural,li2025clustering}. 
Typically, these methods leverage message-passing mechanisms and consistency regularization to reinforce feature similarity among intra-cluster nodes~\cite{zhu2021graph}. 
However, as the scale of real-world networks grows exponentially, these frameworks encounter a formidable bottleneck: the trade-off between computational scalability and structural integrity~\cite{gao2018large,hamilton2017inductive}. \par

To handle massive graphs, existing methods predominantly adopt mini-batch training via various stochastic sampling strategies~\cite{zeng2019graphsaint}. 
Although GCL has demonstrated remarkable performance in representation learning, its efficacy on large-scale data is severely hampered by the "structural isolation" issue induced by batch-wise operations~\cite{shi2024label}. 
Since traditional samplers primarily focus on local neighborhood reachability, they inevitably sever the global community context, sequestering nodes of the same semantic cluster into disparate batches.  
Such limitations are visually depicted in Figure~\ref{fig: toy}, where conventional random sampling (a) and local neighborhood sampling (b) fail to maintain community coherence within mini-batches. 
This structural isolation inherently erodes global topological awareness, a problem further exacerbated by the inherent limitations of current technological paths: contrastive-based frameworks ~\cite{liu2022sublime,kulatilleke2025scgc,yang2023convert,shen2023neighbor} typically lack mechanisms to compensate for such sampling-induced fragmentation,  while structure-aware methods~\cite{sun2024lsenet,liu2023dink,tsitsulin2023graph} often rely on noisy, unrefined graph statistics. 
Consequently, the absence of global topological awareness makes it challenging for models to capture cohesive community structures, leading to fragmented representations that fail to reflect the underlying semantic organization.\par

To address the aforementioned challenges, we propose \textbf{~\model{}}, a \textbf{S}calable unsupervised graph \textbf{C}lustering framework designed to preserve structural \textbf{I}ntegrity through the synergy of community-aware sampling and constrained \textbf{S}tructural \textbf{E}ntropy. 
\model{} reconciles the inherent conflict between computational scalability and community cohesion through three synergistic components.
First, we introduce the Structural Entropy Community Constraint (SECC) operator to explicitly quantify and reinforce community compactness. 
Unlike traditional structural entropy metrics that often lead to excessive partitioning in unlabeled scenarios, SECC optimizes structural information within a bounded solution space constrained by the target community number. 
This allows the model to maintain minimal entropy while effectively mitigating community fragmentation.
Second, to counteract the information loss inherent in mini-batch training, we design a Community-Aware Sampling Expansion (CSampE) mechanism. 
By identifying and incorporating the global community context of target nodes into each sampling batch, CSampE dismantles structural barriers between batches and restores the topological continuity required for stable optimization.
Finally, we devise a Structural Contrastive Learning (StructCL) method to guide the encoder toward a more discriminative representation space. 
Diverging from conventional paradigms that rely on noisy raw topology, StructCL refines higher-order edge weights based on latent structural similarities among intra-batch nodes. 
These refined affinities serve as structural guidance to maximize consistency among similar nodes, ultimately yielding high-quality embeddings that are partitioned into final clusters.\par

To evaluate the effectiveness of the proposed ~\model{}, we conduct extensive experiments on three medium-scale datasets (Photo, Computer, Pubmed) and three large-scale datasets (Ogbn-arxiv, Reddit, Ogbn-products). 
Quantitative results yield several critical insights:
First, comprehensive performance analysis demonstrates that ~\model{} achieves overall superior performance compared to ten state-of-the-art baselines, confirming its consistent efficacy across both medium and large-scale scenarios. 
Second, rigorous ablation studies dissect the synergistic effects of our modules, validating that SECC, CSampE, and StructCL are each indispensable for preserving structural integrity. 
Finally, the stability analysis assesses the model's invariance to hyper-parameter variations, underscoring its robustness and deployment in real-world, large-scale systems.
The main contributions of this work are summarized as follows.\par

$\bullet$
\textbf{Scalable Framework (\model{}):} We propose ~\model{}, an unsupervised graph clustering framework that effectively harmonizes the conflict between computational scalability and structural integrity through the synergy of community-aware sampling and constrained structural entropy.

$\bullet$
\textbf{Community-Aware Sampling Expansion (CSampE):} We develop the CSampE mechanism to bridge "structural isolation" by adaptively incorporating global community contexts into mini-batches, thereby restoring topological continuity in mini-batch training.

$\bullet$
\textbf{Structural Entropy Community Constraint (SECC):} We introduce the SECC operator to explicitly quantify community compactness within a constrained space, significantly mitigating partition fragmentation and ensuring global community integrity.

$\bullet$
\textbf{Extensive Validation and Insights:} Through extensive evaluations on six benchmark datasets, we demonstrate the overall superior performance and robustness of SCISE, while identifying the synergistic and complementary effects of its core components.

\begin{figure*}[t]
	\centering
	\includegraphics[width=0.9\linewidth]{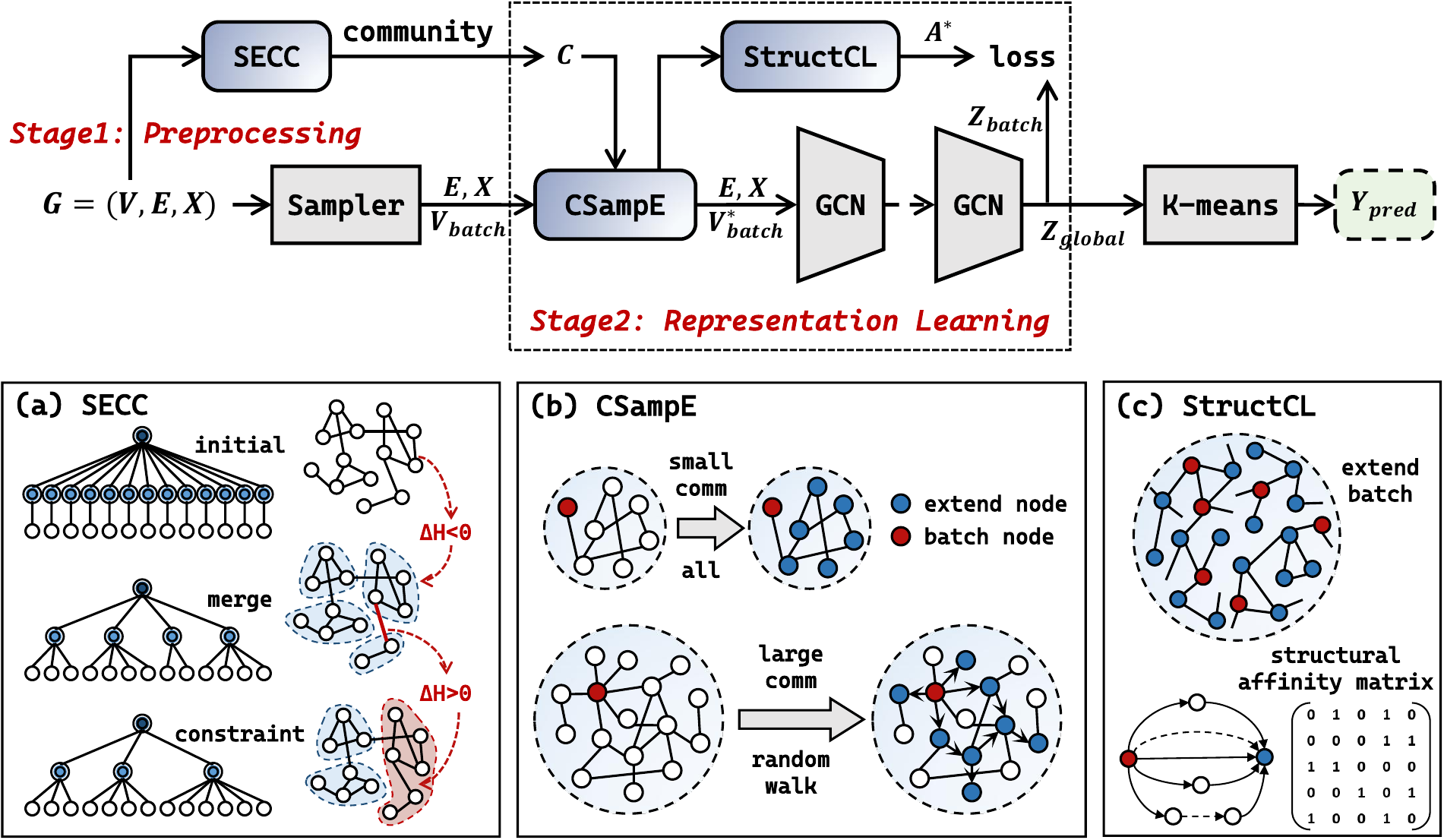}
    \vspace{-3mm}
	\caption{Overall framework of \model{}.}
	\label{fig: framework}
    \vspace{-5mm}
\end{figure*}

\vspace{-4mm}
\section{Related Work}
\label{sec: related_work}
\vspace{-2mm}
\subsection{Unsupervised Graph Clustering}
Unsupervised graph clustering evolves from traditional statistical methods to sophisticated deep learning paradigms. 
Systematic surveys~\cite{liu2022survey, ren2024deep} establish comprehensive taxonomies, categorizing the field into GNN-based and contrastive learning-based approaches while analyzing the trade-offs between generative and discriminative models. 
Within the GNN paradigm, DMoN~\cite{tsitsulin2023graph} achieves end-to-end community detection via modularity maximization, while DeepCut~\cite{aflalo2023deepcut} and RGC~\cite{liu2023reinforcement} extend these techniques to unsupervised image segmentation and reinforcement learning-based automated cluster number determination, respectively. 
Recently, contrastive learning has emerged as a dominant trend to enhance discriminative power; for instance, SCGC~\cite{kulatilleke2025scgc} simplifies contrastive architectures through neighbor-oriented loss, while CCGC~\cite{yang2023cluster} and DCGL~\cite{chen2024deep} leverage clustering-oriented guidance to refine representation learning. 
Parallel to these algorithmic advancements, structural entropy introduces a robust information-theoretic perspective for capturing graph hierarchies, with methods like USER~\cite{wang2023user} and the Lorentz-based LSEnet~\cite{sun2024lsenet} pushing the boundaries of accuracy and parameter-free clustering.
Furthermore, applying structural entropy to unsupervised graph clustering~\cite{zhang2025dese} opens new avenues for synergizing structural information theory with neural networks.\par

\vspace{-3mm}
\subsection{Structural Information Theory}
Structural Information Theory has been validated in many applications. 
Introducing structural entropy in neural networks captures the underlying topological connectivity and reduces random interference~\cite{wang2023user,zhang2025dese}. 
The hierarchical nature of the structure entropy encoding tree provides new methods for hierarchical structure pooling in graph neural network~\cite{wu2022structural}, unsupervised image segmentation~\cite{xie2025hierarchical,huang2026sector}, dimension estimation~\cite{yang2023minimum}, state abstraction~\cite{zeng2023hierarchical,zeng2026structural} in reinforcement learning, social bot detection~\cite{peng2024unsupervised,zeng2025proactive}, unsupervised social event detection~\cite{cao2024hierarchical}, and recommendation systems~\cite{zhang2025enhanced}.
Additionally, reconstructing the graph structure on the hierarchical encoding tree suppresses edge noise and enhances the learning ability of the graph structure~\cite{zou2024structural}. 
Furthermore, modifying the network structure based on minimizing structural entropy achieves maximum deception of community structure~\cite{liu2019rem}. 
Similarly, SEGA~\cite{wu2023sega} and SEPC~\cite{huang2025structural} improve the performance of graph contrastive learning, guided by the principle of structural entropy minimization.
Based on the homogeneous graph structure entropy, studys of multi-relational graph SE~\cite{cao2024multi} and hypergraph further SE~\cite{zhang2025enhanced,zeng2026hyperbolic} extend the structural information theory, making it suitable for more complex scenarios. \par

\vspace{-3mm}
\subsection{Large Graph Learning}
Large-scale graph learning has evolved to address the scalability bottlenecks of traditional Graph Neural Networks (GNNs). 
As foundational works, GraphSAGE~\cite{hamilton2017inductive} introduces neighbor sampling for inductive representation learning, while VR-GCN~\cite{chen2018stochastic} adopts variance reduction to stabilize stochastic training. 
To overcome the memory explosion in deep GNNs, Cluster-GCN~\cite{chiang2019cluster} leverages graph clustering to restrict neighborhood search within dense subgraphs. 
Further refinements in sampling strategies, such as GraphSAINT~\cite{zeng2019graphsaint}, utilize subgraph sampling to decouple the number of layers from sampling complexity. 
Parallel to these developments, model simplification has become a key trend; SGC~\cite{wu2019simplifying} removes non-linearities to fold GCN into a linear filter.
Recently, the research frontier has shifted toward scaling Graph Transformers to large graphs.
GraphGPS~\cite{rampavsek2022recipe} decouples local and global mechanisms, while SGFormer~\cite{wu2023sgformer} and Exphormer~\cite{shirzad2023exphormer} employ simplified global attention or sparse frameworks to mitigate quadratic complexity. \par

\vspace{-3mm}
\section{Preliminary}
\label{sec: preliminary}
\begin{definition}[\textbf{Unsupervised Graph Clustering}] 
The primary objective of unsupervised graph clustering is to partition nodes into cohesive groups by exclusively exploiting intrinsic topological structures and nodal attributes, in the absence of any ground-truth labels.
Formally, let $G = (V, E, \mathbf{X})$ denote an undirected homogeneous graph, where $V = \{v_1, v_2, \dots, v_N\}$ is the set of $N$ nodes and $E \subseteq V \times V$ represents the set of $M$ edges.
The nodal properties are encapsulated in a feature matrix $\mathbf{X} \in \mathbb{R}^{N \times f}$, where $f$ denotes the dimensionality of the attribute space.
The connectivity patterns within $G$ are characterized by a symmetric adjacency matrix $\mathbf{A} \in \{0, 1\}^{N \times N}$, where $\mathbf{A}_{ij} = 1$ indicates the existence of an edge between nodes $v_i$ and $v_j$.
Given the graph $G$, the objective is to learn a mapping function $f: \{V, \mathbf{A}, \mathbf{X}\} \to \mathbf{Y}$ that assigns each node to one of $K$ disjoint clusters $\mathbf{Y} \in \{1, \dots, K\}^N$, where $K$ is a pre-defined or inferred number of clusters.
\end{definition}

\vspace{-3mm}
\begin{definition}[\textbf{Structural Entropy}]
Structural information theory~\cite{li2016structural} is originally proposed for measuring the structural information contained within a graph. 
Specifically, this theory aims to calculate the structural entropy of the homogeneous graph $G=(V, E)$, which reflects its uncertainty when undergoing hierarchical division. 
The structural information of the homogeneous graph $G$ determined by the encoding tree $\mathcal{T}$ is defined as:
\begin{eqnarray} 
    H^\mathcal{T}(G)=-\sum_{\alpha \in \mathcal{T},\alpha \neq \lambda}\frac{g_\alpha}{vol(G)}\log\frac{vol(\alpha)}{vol(\alpha^-)},
\label{eq: H_T}
\end{eqnarray}
where $vol(G)$ is the sum of the degrees of all nodes in the graph $G$. 
Each vertex in the encoding tree $\mathcal{T}$ corresponds to a node subset $T_{\alpha}$  in the graph $G$.
$vol(\alpha)$ is the volume of $T_{\alpha}$ and is the sum of the degrees of all nodes in the subset $T_{\alpha}$. 
$\alpha^-$ is the parent vertex of vertex $\alpha$ in the encoding tree.
$g_{\alpha}$ is the sum of weights of all edges from node subset $T_{\alpha}$ to node subset $V/ T_{\alpha}$, which can be understood as the total weight of the edges from the nodes outside the node subset $T_{\alpha}$ to the nodes inside $T_{\alpha}$, or the total weight of the cut edges.
$\frac{g_\alpha}{vol(G)}$ represents the probability that the random walk enters $T_{\alpha}$.
The structural entropy $H(G)$ of graph $G$ is the minimum $H^{\mathcal{T}}(G)$. 
Let $\mathcal{T}_h$ be encoding trees whose height is not greater than $h$, then the $h$-dimensional structural entropy of $G$ is defined as $H_h(G)=min \{ H^{\mathcal{T}_h}(G) \}$.
\end{definition}

\vspace{-3mm}
\section{Methodology}
\label{sec: methodology}
This section presents a detailed formalization of the proposed scalable unsupervised graph clustering framework, \textbf{\model{}}. 
As illustrated in Figure~\ref{fig: framework}, \model{} comprises three strategic components: Structural Entropy Community Constraint (SECC), which optimizes structural entropy minimization under community-cardinality constraints to generate cohesive partitions;
Community-aware Sampling Extension (CSampE), which mitigates the "structural isolation" of mini-batches by performing community-aware context expansion; 
and Structural Contrastive Learning (StructCL), which executes contrastive alignment on a higher-order topology reconstructed via structural similarity.
Specifically, the pipeline begins with SECC (Section~\ref{subsec: SECC}) as a preprocessing stage, leveraging a constrained structural entropy operator to partition the raw graph into dense community anchors. 
During the representation learning stage, CSampE (Section~\ref{subsec: CSampE}) augments sampled nodes with their corresponding intra-community neighbors to restore topological integrity. 
Subsequently, StructCL (Section~\ref{subsec: StructCL}) refines the batch-wise subgraph by re-weighting edges based on structural similarities approximated through random walks.
Finally, the Overall Optimization (Section~\ref{subsec: optimization}) integrates these modules into a unified objective to refine latent embeddings, followed by K-means clustering to produce the final partitions.
A detailed complexity analysis of \model{} is provided in Section~\ref{subsec: complexity_analysis}.\par

\vspace{-3mm}
\subsection{Structural Entropy Community Constraint (SECC)}
\label{subsec: SECC}

Applying structural information theory to graph structure learning has demonstrated substantial advantages across various downstream tasks, primarily due to its capacity to reveal the hierarchical topological semantics of nodes. 
However, current optimization paradigms for structural entropy predominantly rely on a greedy merge operator—iteratively merging node pairs until no further reduction in structural entropy is possible.
This unconstrained, locally greedy approach is highly susceptible to over-segmentation and frequently traps the model in local optima. 
As illustrated in the case study in Section~\ref{sec: case_study}, conventional methods often yield a plethora of fragmented, micro-communities (e.g., sizes of 2, 3, or 4). 
In ideal real-world scenarios, however, community partitions are expected to be cohesive, characterized by an abundance of medium-sized communities and a minimal presence of extreme-sized (excessively large or small) communities.\par

\setlength{\textfloatsep}{5pt plus 2pt minus 2pt}
\setlength{\intextsep}{5pt plus 2pt minus 2pt}
\begin{algorithm}[h]
\setstretch{0.55}
\LinesNumbered
    \KwIn{Graph $G=(V, E)$, target community count $N_{\text{comm}}$, merge speed $p \in (0, 1]$.}
    
    \KwOut{Community partition $\mathcal{C} = \{c_1, c_2, \dots, c_{N_{\text{comm}}}\}$.}
    
    Initialize $\mathcal{C} = \{\{v_1\}, \{v_2\}, \dots, \{v_N\}\}$; \\
    
    Calculate $\text{vol}(G) = \sum_{v \in V} d_v$; \\
    
    \While{$|\mathcal{C}| > N_{\text{comm}}$}{
        $\mathcal{E}_{cand} = \{ (c_i, c_j) \mid c_i, c_j \in \mathcal{C}, i < j, \exists e_{u,w} \in E \text{ s.t. } v_u \in c_i, v_w \in c_j \}$; \\
        
        \ForEach{$(c_i, c_j) \in \mathcal{E}_{cand}$}{
            Calculate $\Delta H(c_i,c_j)$ via Eq. \eqref{eq: merge}; \\
        }
        
        Sort $\mathcal{E}_{cand}$ in descending order of $\Delta H$; \\
        
        $P_{pos} = \{ (c_i, c_j) \in  \mathcal{E}_{cand} \mid \revised{\Delta H(c_i,c_j) < 0} \}$; \\
    
        \tcp{Forced Convergence}
        \If{$P_{pos}$ is empty}{
            $P_{pos} = \mathcal{E}_{cand}$; \\
        }
        $OP =|P_{pos}|$; \\
        
        $k = \min \left( p \cdot OP, |\mathcal{C}| - N_{\text{comm}} \right)$; \\
        
        Select top $k$ pairs from $P_{pos}$ and merge them\;
        Update $\mathcal{C}$ and $|\mathcal{C}| = |\mathcal{C}| - k$\;
    }
    \Return $\mathcal{C}$
    
    \caption{Structural Entropy Community Constraint}
    \label{algo: SECC}
\end{algorithm}

To address this bottleneck, we impose a community cardinality constraint, denoted as $N_{\text{comm}}$, on the greedy structural entropy minimization process. 
Consequently, the termination criterion for the merge operator shifts from "the absence of entropy-decreasing node pairs" to "convergence to the target community count $N_{\text{comm}}$".
Specifically, the process initializes with all $N$ nodes forming individual singleton communities, denoted as $\mathcal{C} = \{c_1, c_2, \dots, c_N\}$, where inter-community connections correspond to the original edge set $E$. 
For each edge $e_{i,j} \in E$, we calculate the change in structure entropy, $\Delta \text{SE}(i,j)$, resulting from the merger of $c_i$ and $c_j$:
\begin{equation}
\footnotesize
\begin{aligned}
\Delta H(c_i,c_j) = & -\frac{g_{c_i \cup c_j}}{vol(G)} \log \frac{vol(c_i \cup c_j)}{vol(G)} + \sum_{v_t \in c_i \cup c_j} \left( -\frac{d_{v_t}}{vol(G)} \log \frac{d_{v_t}}{\text{vol}(c_i \cup c_j)} \right) \\
& - \left( -\frac{g_{c_i}}{\text{vol}(G)} \log \frac{\text{vol}(c_i)}{\text{vol}(G)} + \sum_{v_t \in c_i} \left( -\frac{d_{v_t}}{\text{vol}(G)} \log \frac{d_{v_t}}{\text{vol}(c_i)} \right) \right) \\
& - \left( -\frac{g_{c_j}}{\text{vol}(G)} \log \frac{\text{vol}(c_j)}{\text{vol}(G)} + \sum_{v_t \in c_j} \left( -\frac{d_{v_t}}{\text{vol}(G)} \log \frac{d_{v_t}}{\text{vol}(c_j)} \right) \right),
\end{aligned}
\label{eq: merge}
\end{equation}
where $vol(\cdot)$ denotes the volume of the community, $g(\cdot)$ represents the cut edges, and $d_{v_t}$ signifies the node degree. 
Subsequently, we sort the entropy variations of all candidate edges in increasing order, retaining $OP$ pairs where $\Delta H(c_i,c_j) < 0$. 
To control the pace and efficiency of the agglomeration, we introduce a hyperparameter $p$ ($0 < p \leq 1$), executing the merge operation only on the top $p$ proportion of these candidate pairs. 
Following one iteration, a new community set $\mathcal{C}' = \{c_1, c_2, \dots, c_{N'}\}$ is formed, where $N' = N - p \times OP$.
The core innovation of the constraint lies in its ability to circumvent premature convergence.
So, before termination, if no node pairs satisfy $\Delta H(c_i,c_j) < 0$, yet the current number of communities has not reached $N_{\text{comm}}$, the algorithm relaxes the strict entropy-descent condition. 
It forces further merges—regardless of temporary structural entropy increases—until the total number of communities is strictly limited to $N_{\text{comm}}$ (detailed in Algorithm~\ref{algo: SECC}). 
This mechanism effectively breaks out of local optima and yields highly compact partitions by compelling nodes into exactly $N_{\text{comm}}$ cohesive groups, while ensuring global structural integrity. 
\revised{The unconstrained minimization algorithm can be used to estimate a feasible upper bound for the community range. 
As demonstrated in Section~\ref{sec: hyperparameter}, \model{} is highly insensitive to the precise value of $N_{comm}$, and therefore no exhaustive search for an optimal community number is required in practice.}\par

\vspace{-3mm}
\subsection{Community-aware Sampling Extension (CSampE)}
\label{subsec: CSampE}
To maintain computational efficiency on large-scale graphs, models typically rely on stochastic node sampling to construct mini-batches. 
However, this inherent randomness frequently severs deep-seated community ties. 
Nodes belonging to the same cohesive cluster may be partitioned into disparate batches, preventing the model from capturing the full topological evolution of a community within a single gradient update—a phenomenon we define as Structural Isolation. 
Conventional node-centric paradigms (e.g., neighbor sampling) often neglect long-range community constraints, leading to fragmented subgraphs where intra-batch nodes are topologically isolated.
This fragmentation hinders the model from learning the cohesive hierarchical semantics intrinsic to the data.\par

CSampE bridges this gap by adaptively expanding the initial batch $\mathcal{V}_{batch}$ into structurally dense collectives based on underlying community memberships. 
By shifting the sampling paradigm from "node expansion" to "community alignment," CSampE identifies the corresponding community $c(v) \in \mathcal{C}$ for each seed node $v \in \mathcal{V}_{batch}$ via the partitions generated by SECC. 
This mechanism pulls nodes from the same community and their critical topological structures into a unified sampling space, forcing the model to observe complete community semantics in each training step.
Specifically, for each node $v \in \mathcal{V}_{batch}$ associated with community $c(v)$, we introduce a cardinality threshold $\theta$ to govern the expansion strategy:\par

Case I: Structural Restoration for Small Communities ($|c(v)| \leq \theta$). 
For small-scale communities, we execute a full-community expansion to ensure the local topology is entirely restored within a single training step. 
The expanded set $\mathcal{V}_{ext}$ is defined as:
\begin{equation}
    \mathcal{V}_{ext} = \{u \mid \exists v \in \mathcal{V}_{batch}, u \in c(v) \},
    \label{eq: extend_small}
\end{equation}\par

Case II: Backbone Extraction for Large Communities ($|c(v)| > \theta$). 
In expansive communities, full inclusion would trigger "batch explosion" and excessive memory overhead. 
To distill the most representative features, we execute $w_l$ random walks of length $w_t$ starting from node $v$ within $c(v)$. 
Let $f(u; v)$ denote the frequency of node $u$ being visited by node $v$ within $c(v)$.
We identify the structural backbone by filtering nodes that exceed the mean visitation frequency:
\begin{equation}
    \mathcal{V}_{ext} = \{ u \mid \exists v \in \mathcal{V}_{batch} \text{ s.t. } u \in c(v) \text{ and } f(u; v) > \bar{f}_v \},
    \label{eq: extend_large}
\end{equation}
where $\bar{f}_v = \frac{1}{|\mathcal{R}_v|} \sum_{w \in \mathcal{R}_v} f(w; v)$ denotes the mean visitation frequency associated with seed $v$, and $\mathcal{R}_v$ represents the set of all nodes reachable from seed $v$ within the community via random walks.
The final expanded batch, $\mathcal{V}^*_{batch} = \mathcal{V}_{batch} \cup \mathcal{V}_{ext}$, transforms a collection of isolated points into a coherent structural field. 
This dual-track strategy enables the model to leverage dense community-aware signals while maintaining an effectively managed batch size, thereby enhancing the stability of gradient updates and the robustness of learned hierarchical representations.

\vspace{-3mm}
\subsection{Structural Contrastive Learning (StructCL)}
\label{subsec: StructCL}
While the CSampE module successfully assembles structurally dense mini-batches, efficiently translating these multi-hop topological patterns into the representation space remains a significant challenge. 
A naive approach would be to treat all node pairs within the same community as positive anchors. 
However, for large-scale graphs, even with the reduced scale of a mini-batch, this would lead to a quadratic explosion ($\mathcal{O}(|\mathcal{V}^*_{batch}|^2)$) in the number of positive pairs, incurring prohibitive computational overhead. 
To quantify community-level structural relations without inflating complexity, we propose Structural Contrastive Learning (StructCL).\par

To efficiently capture community-level structural relations, we construct a structural affinity matrix $\mathbf{A}^*$ based on the visitation counts of localized random walks. 
For each node $v_i \in \mathcal{V}^*_{batch}$, we execute $w_t$ random walks of length $w_l$ within the graph. Let $Count(v_j; v_i)$ denote the total frequency of node $v_j$ being visited in walks originating from $v_i$. 
The structural affinity matrix $\mathbf{A}^* \in \mathbb{R}^{B \times B}$ (where $B = |\mathcal{V}^*_{batch}|$) is defined as:
\begin{equation}
\mathbf{A}^*_{ij} =
\begin{cases}
Count(v_j; v_i), & \text{if } i \neq j \text{ and } v_j \in \text{RW}(v_i) 
\\ 0, & \text{otherwise}
\end{cases},
\label{eq: batch_A}
\end{equation}
This sparse matrix effectively transforms the batch into a weighted structural subgraph.
Within this subgraph, the weights $\mathbf{A}^*_{ij}$ serve as confidence scores for topological co-occurrence: a higher frequency indicates a more stable structural bond. 
By selectively focusing on these frequently visited "structural anchors," StructCL effectively captures the core community semantics while avoiding the computational redundancy of all-to-all community matching.\par

To align the neural similarity of embeddings with the distilled structural affinities, for any node pair $(v_i, v_j)$ with $\mathbf{A}^*_{ij} > 0$, we define the consistency loss as a multi-positive InfoNCE objective:
\begin{equation}
\mathcal{L}_{struct} = - \frac{1}{B} \sum_{v_i \in \mathcal{V}^*_{batch}} \frac{1}{|\mathcal{P}_i|} \sum_{v_j \in \mathcal{P}_i} \log \frac{\exp(\text{sim}(\mathbf{z}_i, \mathbf{z}_j) / \tau)}{\sum_{v_k \neq v_i} \exp(\text{sim}(\mathbf{z}_i, \mathbf{z}_k) / \tau)},
\label{eq: loss}
\end{equation}
where $\mathcal{P}_i = \{v_j \mid \mathbf{A}^*_{ij} > 0\}$ is the sparse set of structural neighbors for node $v_i$. 
By minimizing $\mathcal{L}_{struct}$, the model adaptively encodes dense intra-community signals into the representation space. 
This self-supervised mechanism ensures that the learned embeddings are inherently endowed with community-topology awareness, effectively bridging the "structural isolation" gap while maintaining a bounded computational complexity suitable for large-scale graph analysis.\par

\vspace{-1mm}
\begin{algorithm}[h]
\caption{Overall execution procedure of~\model{}.}
\label{algo: SCISE}
\setstretch{0.55}
\LinesNumbered
    \KwIn{Graph with feature $G=(V, E)$, features $\mathcal{X}$, adjacency matrix $A$, target community count $N_{\text{comm}}$, merge speed $p \in (0, 1]$, expansion threshold $\theta$, random walk params $(w_l, w_t)$, batch size $B$, target clusters $K$.}
    \KwOut{Cluster assignments $\mathcal{Y}$.}

    \tcp{Stage 1: Preprocessing community division}
    $\mathcal{C} \leftarrow \text{SECC}(G, N_{\text{comm}}, p)$ via Algorithm~\ref{algo: SECC}; \\
    
    Construct community lookup mapping $c(v)$ for each $v \in V$ from $\mathcal{C}$; \\

    \tcp{Stage 2: Structure-Aware Representation Learning}
    
    \For{epoch = 1, 2, 3,...}{

        \ForEach{mini-batch $\mathcal{V}_{batch} \subset V$}{
        
            Initialize $\mathcal{V}_{ext} \leftarrow \emptyset$;\\

            \ForEach{$v \in \mathcal{V}_{batch}$}{
                \If{$|c(v)| \leq \theta$}{
                    \tcp{Case I: Full Expansion}
                    $\mathcal{V}_{ext} \leftarrow \mathcal{V}_{ext} \cup \{ u \mid u \in c(v) \}$ via Eq.~\ref{eq: extend_small};\\
                }
                \Else{
                    \tcp{Case II: Backbone Extraction}
                    Perform $w_t$ random walks of length $w_l$ within $c(v)$;\\
                    
                    Update $\mathcal{V}_{ext}$ via Eq.~\ref{eq: extend_large};\\ 
                }
            }
            
            $\mathcal{V}^*_{batch} = \mathcal{V}_{batch} \cup \mathcal{V}_{ext}$;\\

            \tcp{StructCL Optimization}
            Compute structural affinity $\mathbf{A}^*$ via Eq.~\ref{eq: batch_A};\\
            
            $\mathbf{Z}_{batch} = \text{GCN}(\mathcal{X}[\mathcal{V}^*_{batch}], A)$;\\

            calculate $\mathcal{L}_{struct}$ via Eq.~\ref{eq: loss};\\

            $\mathcal{L}_{struct}$ backward;\\

        }

    }


    $\mathbf{Z}_{global} = \emptyset$;\\
    
    \ForEach{mini-batch $\mathcal{V}_{batch} \subset V$}{
    
        $\mathbf{Z}_{batch} = \text{GCN}(\mathcal{X}[\mathcal{V}_{batch}], A)$;\\
        
        $\mathbf{Z}_{global} = \mathbf{Z}_{global} \cup \mathbf{Z}_{batch}$\;
    }
    
    $\mathcal{Y} = \text{K-means}(\mathbf{Z}_{global}, K)$;\\

    \Return $\mathcal{Y}$
\end{algorithm}

\vspace{-3mm}
\subsection{Overall Optimization}
\label{subsec: optimization}
The overall execution procedure of ~\model{} is summarized in Algorithm \ref{algo: SCISE}. 
Our framework adopts a two-stage paradigm consisting of community division preprocessing and structure-aware representation learning. 
Specifically, the SECC module is first invoked to obtain a stable community structure mapping $\mathcal{C}$, which circumvents the high computational overhead of recalculating communities at each training step. 
During the training phase, for each mini-batch $\mathcal{V}_{batch}$, the CSampE module adaptively expands the batch with community nodes $\mathcal{V}_{extend}$ based on the pre-acquired community scales. 
This enables the subsequent StructCL objective to distill community-level semantic signals into the GCN encoder without incurring the "neighbor explosion" problem. 
Subsequently, the node features are processed through a $GCN$ backbone to learn the node embeddings $\mathbf{Z}_{batch}$. 
Finally, a single-pass K-means clustering is performed on the consolidated global embeddings $\mathbf{Z}_{global}$ to obtain the final partitions $\mathcal{Y}_{pred}$.\par

\vspace{-4mm}
\subsection{Complexity Analysis}
\label{subsec: complexity_analysis}
To demonstrate the scalability of \model{}, we analyze its computational complexity across two primary stages: offline structural prior acquisition and structure-aware representation learning.
First, the complexity of the SECC module is governed by the construction of the structural entropy encoding tree. 
Given a graph $G$ with $N$ nodes and $M$ edges, when the parallel merge rate is $p$, the maximum number of rounds required for hierarchical merging is $\log_{1-p} \frac{1-p}{N-1}$. 
Consequently, the time complexity for constructing the structural prior is $O(M \cdot \log_{1-p} \frac{1-p}{N-1})$, which can be simplified to $O(M \log N)$. 
Since SECC is performed as a one-time preprocessing step, it does not add to the iterative overhead of the training phase.\par

In the online training phase, the complexity for each mini-batch $\mathcal{V}_{batch}$ consists of three components: CSampE expansion, neural encoding, and StructCL optimization. 
For each batch, the adaptive expansion via either full community retrieval or localized random walks incurs a complexity of $O(B \cdot w_l \cdot w_t)$, where $B$ is the initial batch size, and $(w_l, w_t)$ are the random walk parameters. 
The forward pass through the GCN has a complexity of $O(|\mathcal{V}^*_{batch}| \cdot f \cdot d)$, where $f$ and $d$ denote the input and embedding dimensions, respectively. 
Calculating the structural contrastive loss involves pairwise similarity computations within the expanded batch, resulting in $O(|\mathcal{V}^*_{batch}| \cdot w_l \cdot w_t)$. 
Given that $|\mathcal{V}^*_{batch}|$ is bounded within a small constant multiple of $B$, the total time complexity per iteration during the representation learning stage is $O(B \cdot( w_lw_t + fd))$.
Extending this to a full training epoch, the overall complexity remains $O(N \cdot( w_lw_t + fd))$.\par

                

\begin{table}[t]
    \centering
    \caption{Statistics of datasets.}
    \label{tab: dataset}
    \vspace{-4mm}
    \small
    \renewcommand\arraystretch{1.1}
    \setlength{\tabcolsep}{1.0mm}
    \begin{tabular}{l|ccccc}
        \toprule
        \textbf{Dataset} & \textbf{\#Node} & \textbf{\#Edge} & \textbf{\#Feature} & \textbf{\#Cluster} &\revised{\textbf {Degree}}\\
        \hline

        Photo & 7,650 & 119,081 & 745 & 8 & \revised{31.13} \\
        Computers & 13,752 & 245,861 & 767 & 10 & \revised{35.76} \\
        Pubmed & 19,717 & 44,324 & 500 & 3 & \revised{4.50} \\

        \hline
        
        Ogbn-arxiv & 169,343 & 1,166,243 & 128 & 40 & \revised{13.77} \\
        Reddit & 232,965 & 114,615,892 & 602 & 41 & \revised{491.98} \\
        Ogbn-products & 2,449,029 & 61,859,140 & 100 & 47 & \revised{50.52} \\
                
        \bottomrule
    \end{tabular}
    \vspace{-2mm}
\end{table}

\vspace{-3mm}
\section{Experiments}
\label{sec: experiments}

In this section, we conduct empirical experiments to demonstrate the effectiveness of the proposed framework~\model{}. 
We aim to answer five research questions as follows:
\textbf{Q1: Effectiveness.} How does \model{} perform in unsupervised graph clustering tasks compared with state-of-the-art baselines (Section~\ref{sec: cluster})?
\textbf{Q2: Ablation Study.} How do the SECC, CSampE, and StructCL modules influence the performance of ~\model{}, and how do they interact with each other (Section~\ref{sec: ablation})?
\textbf{Q3: Stability Analysis.} How do key hyperparameters impact the performance of ~\model{}, and what is the degree of model sensitivity to these parameters (Section~\ref{sec: hyperparameter})?
\revised{\textbf{Q4: Robustness Against Noise.} How robust is \model{} to noisy community priors (Section~\ref{sec: noise})?}
\revised{\textbf{Q5: Robustness Against Sparsity.} Can \model{} effectively alleviate structural isolation and maintain clustering performance when graph connectivity is severely degraded (Section~\ref{sec: sparsity})?}
\textbf{Q6: Case Study.} How do the community number ($N_{comm}$) constraints and merge speed ($p$) balance structural entropy and community cohesion (Section~\ref{sec: case_study})?
\textbf{Q7: Visualization.} Can ~\model{} learn discriminative node representations that produce well-separated cluster distributions in a low-dimensional space (Section~\ref{sec: visualization})?

\begin{table*}[t]
    \centering
    \caption{Comparison of the NMI, ARI, ACC, and F1 across different methods on six datasets. The best results are bolded, and the second-best results are underlined.}
    \label{tab: results}
    \vspace{-4mm}
    \renewcommand\arraystretch{1.1}
    \setlength{\tabcolsep}{0.3mm}
    \begin{tabular}{c|cccc|cccc|cccc}
        \toprule
        \multirow{2}*{Method} 
        & \multicolumn{4}{c|}{\multirow{1}*{\textbf{Photo}}} 
        & \multicolumn{4}{c|}{\multirow{1}*{\textbf{Computers}}} 
        & \multicolumn{4}{c}{\multirow{1}*{\textbf{Pubmed}}} \\
        
        \cline{2-13}
        
        & {\textbf{NMI}} & {\textbf{ARI}} & {\textbf{ACC}} & \multicolumn{1}{c|}{\textbf{F1}} 
        & {\textbf{NMI}} & {\textbf{ARI}} & {\textbf{ACC}} & \multicolumn{1}{c|}{\textbf{F1}} 
        & {\textbf{NMI}} & {\textbf{ARI}} & {\textbf{ACC}} & \multicolumn{1}{c}{\textbf{F1}} \\

        \hline

        DGI
        & 33.70$_{\pm 0.85}$ & 22.10$_{\pm 0.55}$ & 43.00$_{\pm 1.10}$ & 35.20$_{\pm 0.90}$ 
        & 42.00$_{\pm 1.05}$ & 30.60$_{\pm 0.85}$ & 47.90$_{\pm 1.20}$ & 39.00$_{\pm 1.00}$ 
        & 32.20$_{\pm 0.80}$ & 29.20$_{\pm 0.75}$ & 65.70$_{\pm 1.65}$ & 65.40$_{\pm 1.60}$ \\

        S3GC
        & 59.80$_{\pm 0.47}$ & 56.10$_{\pm 0.73}$ & 75.20$_{\pm 0.58}$ & 72.90$_{\pm 0.69}$ 
        & 56.00$_{\pm 0.62}$ & 43.80$_{\pm 0.41}$ & 58.80$_{\pm 0.76}$ & 47.50$_{\pm 0.55}$ 
        & \underline{33.30$_{\pm 0.79}$} & \underline{34.50$_{\pm 0.64}$} & \underline{71.30$_{\pm 0.48}$} & \underline{70.30$_{\pm 0.71}$}\\

        SUBLIME
        & 56.91$_{\pm 0.43}$ & 45.11$_{\pm 0.98}$ & 64.95$_{\pm 0.61}$ & 61.24$_{\pm 0.25}$
        & 45.55$_{\pm 0.73}$ & 24.89$_{\pm 0.15}$ & 40.94$_{\pm 0.88}$ & 34.64$_{\pm 0.49}$
        & 18.68$_{\pm 0.32}$ & 16.16$_{\pm 0.67}$ & 58.33$_{\pm 0.59}$ & 58.69$_{\pm 0.91}$\\

        CONVERT
        & 63.04$_{\pm 0.34}$ & 55.20$_{\pm 1.18}$ & 74.24$_{\pm 0.67}$ & 69.13$_{\pm 1.05}$ 
        & 48.44$_{\pm 1.42}$ & 34.25$_{\pm 0.53}$ & 51.30$_{\pm 0.91}$ & 42.17$_{\pm 0.28}$
        & 28.19$_{\pm 0.79}$ & 26.59$_{\pm 1.36}$ & 63.29$_{\pm 0.45}$ & 62.29$_{\pm 1.12}$\\

        SCGC
        & 54.96$_{\pm 3.49}$ & 45.66$_{\pm 3.58}$ & 64.90$_{\pm 3.33}$ & 52.02$_{\pm 5.26}$ 
        & 47.41$_{\pm 1.18}$ & 46.06$_{\pm 2.91}$ & 61.16$_{\pm 2.22}$ & 38.07$_{\pm 3.24}$ 
        & 26.05$_{\pm 2.21}$ & 25.27$_{\pm 3.51}$ & 65.12$_{\pm 2.21}$ & 61.49$_{\pm 3.40}$\\

        \hline
        
        DMoN
        & 63.38$_{\pm 0.46}$ & 52.41$_{\pm 0.73}$ & 73.83$_{\pm 0.58}$ & 71.05$_{\pm 0.79}$
        & 49.30$_{\pm 0.64}$ & 31.82$_{\pm 0.41}$ & 49.13$_{\pm 0.77}$ & 39.29$_{\pm 0.52}$
        & 29.80$_{\pm 0.71}$ & 24.52$_{\pm 0.55}$ & 61.99$_{\pm 0.69}$ & 62.17$_{\pm 0.44}$\\

        Dink-Net
        & \underline{74.36$_{\pm 0.35}$} & \textbf{68.40$_{\pm 0.98}$} & \textbf{81.71$_{\pm 0.61}$} & \underline{73.92$_{\pm 1.14}$}
        & 42.10$_{\pm 0.78}$ & 24.85$_{\pm 1.64}$ & 45.17$_{\pm 1.12}$ & 40.38$_{\pm 1.93}$
        & 29.93$_{\pm 1.47}$ & 29.29$_{\pm 0.56}$ & 67.29$_{\pm 1.89}$ & 66.96$_{\pm 1.03}$ \\

        LSEnet
        & 58.94$_{\pm 0.63}$ & 47.80$_{\pm 0.88}$ & 65.46$_{\pm 0.54}$ & 45.80$_{\pm 0.91}$ 
        & 55.03$_{\pm 0.79}$ & 42.15$_{\pm 1.02}$ & 53.27$_{\pm 0.47}$ & 22.41$_{\pm 0.68}$
        & 29.67$_{\pm 0.42}$ & 27.90$_{\pm 0.36}$ & 58.98$_{\pm 0.58}$ & 59.00$_{\pm 0.61}$\\

        MAGI
        & 71.60$_{\pm 0.76}$ & 61.50$_{\pm 0.33}$ & 79.00$_{\pm 1.07}$ & 72.90$_{\pm 0.88}$
        & \textbf{59.20$_{\pm 1.19}$} & \underline{46.20$_{\pm 0.54}$} & \underline{62.00$_{\pm 0.90}$} & \underline{57.40$_{\pm 0.37}$}
        & 29.37$_{\pm 0.25}$ & 29.25$_{\pm 0.40}$ & 66.23$_{\pm 0.23}$ & 65.11$_{\pm 0.20}$\\

        DeSE
        & 68.94$_{\pm 0.28}$ & 62.49$_{\pm 0.12}$ & \underline{80.54$_{\pm 0.12}$} & 73.67$_{\pm 0.03}$
        & 52.10$_{\pm 0.15}$ & 45.64$_{\pm 0.16}$ & 58.78$_{\pm 0.31}$ & 43.17$_{\pm 0.23}$
        & 30.82$_{\pm 0.17}$ & 29.47$_{\pm 0.15}$ & 67.47$_{\pm 0.21}$ & 67.36$_{\pm 0.19}$\\

        \hline

        \rowcolor{gray!20}
        \rule[-1.5ex]{0pt}{4ex} \textbf{\model{}} 
        & \textbf{74.53$_{\pm 0.14}$} & \underline{64.51$_{\pm 0.29}$} & 79.71$_{\pm 0.15}$ & \textbf{76.22$_{\pm 0.12}$}

        & \underline{58.46$_{\pm 0.24}$} & \textbf{51.78$_{\pm 0.80}$} & \textbf{67.32$_{\pm 0.36}$} & \textbf{59.48$_{\pm 0.84}$} 
        
        & \textbf{35.77$_{\pm 0.02}$} & \textbf{36.33$_{\pm 0.02}$} & \textbf{72.38$_{\pm 0.01}$} & \textbf{71.83$_{\pm 0.01}$}\\

        \hline

        \textbf{Improv.(\%)}
        & $\uparrow$ 0.23\% & - & - & $\uparrow$ 3.11\%
        & - & $\uparrow$ 12.07\% & $\uparrow$ 8.58\% & $\uparrow$ 3.63\%
        & $\uparrow$ 7.42\% & $\uparrow$ 5.31\% & $\uparrow$ 1.52\% & $\uparrow$ 2.18\% \\

        \bottomrule

        \multirow{2}*{Method} 
        & \multicolumn{4}{c|}{\multirow{1}*{\textbf{Ogbn-arxiv}}} 
        & \multicolumn{4}{c|}{\multirow{1}*{\textbf{Reddit}}} 
        & \multicolumn{4}{c}{\multirow{1}*{\textbf{Ogbn-products}}}\\
        
        \cline{2-13}
        
        & {\textbf{NMI}} & {\textbf{ARI}} & {\textbf{ACC}} & \multicolumn{1}{c|}{\textbf{F1}} 
        & {\textbf{NMI}} & {\textbf{ARI}} & {\textbf{ACC}} & \multicolumn{1}{c|}{\textbf{F1}} 
        & {\textbf{NMI}} & {\textbf{ARI}} & {\textbf{ACC}} & \multicolumn{1}{c}{\textbf{F1}} \\

        \hline

        DGI
        & 41.20$_{\pm 1.20}$ & 22.30$_{\pm 0.80}$ & 31.40$_{\pm 0.95}$ & 23.00$_{\pm 0.70}$ 
        & 30.60$_{\pm 0.90}$ & 17.00$_{\pm 0.60}$ & 22.40$_{\pm 0.75}$ & 18.30$_{\pm 0.55}$ 
        & 46.70$_{\pm 1.40}$ & 17.40$_{\pm 0.65}$ & 32.00$_{\pm 1.00}$ & 19.20$_{\pm 0.60}$ \\

        S3GC
        & 46.30$_{\pm 0.58}$ & 27.00$_{\pm 1.42}$ & 35.00$_{\pm 0.91}$ & 23.00$_{\pm 1.16}$
        & 80.70$_{\pm 1.27}$ & 74.50$_{\pm 0.63}$ & 73.60$_{\pm 1.08}$ & 56.00$_{\pm 0.75}$ 
        & 53.60$_{\pm 0.84}$ & \underline{23.00$_{\pm 0.35}$} & 40.20$_{\pm 0.56}$ & 25.00$_{\pm 0.92}$\\

        SUBLIME
        & \multicolumn{4}{c|}{OOM}
        & \multicolumn{4}{c|}{OOM}
        & \multicolumn{4}{c}{OOM}\\

        CONVERT
        & \multicolumn{4}{c|}{OOM}
        & \multicolumn{4}{c|}{OOM}
        & \multicolumn{4}{c}{OOM}\\

        SCGC
        & \multicolumn{4}{c|}{OOM}
        & \multicolumn{4}{c|}{OOM}
        & \multicolumn{4}{c}{OOM}\\

        \hline
        
        DMoN
        & 35.60$_{\pm 0.26}$ & 12.70$_{\pm 0.97}$ & 25.00$_{\pm 0.58}$ & 19.00$_{\pm 1.14}$ 
        & 62.80$_{\pm 1.08}$ & 50.20$_{\pm 0.34}$ & 52.90$_{\pm 0.86}$ & 26.00$_{\pm 0.41}$ 
        & 42.80$_{\pm 0.72}$ & 13.90$_{\pm 1.19}$ & 30.40$_{\pm 0.29}$ & 21.00$_{\pm 0.95}$\\

        Dink-Net
        & 43.73$_{\pm 0.23}$ & \underline{35.22$_{\pm 0.41}$} & \underline{43.68$_{\pm 0.35}$} & \underline{26.92$_{\pm 0.48}$} 
        & 78.91$_{\pm 0.39}$ & 71.34$_{\pm 0.26}$ & 76.03$_{\pm 0.44}$ & 67.95$_{\pm 0.31}$ 
        & 50.78$_{\pm 0.50}$ & 21.08$_{\pm 0.34}$ & 41.09$_{\pm 0.28}$ & 25.15$_{\pm 0.42}$\\

        LSEnet
        & \multicolumn{4}{c|}{OOM}
        & \multicolumn{4}{c|}{OOM}
        & \multicolumn{4}{c}{OOM}\\

        MAGI
        & \underline{46.90$_{\pm 0.29}$} & 31.00$_{\pm 1.12}$ & 38.80$_{\pm 0.68}$ & 26.60$_{\pm 1.34}$ 
        & \underline{87.50$_{\pm 1.46}$} & \underline{90.70$_{\pm 0.57}$} & \underline{91.10$_{\pm 0.91}$} & \underline{85.30$_{\pm 0.24}$}
        & \underline{55.10$_{\pm 0.83}$} & 21.50$_{\pm 1.27}$ & \underline{42.50$_{\pm 0.42}$} & \underline{27.60$_{\pm 1.05}$}\\

        DeSE
        & \multicolumn{4}{c|}{OOM}
        & \multicolumn{4}{c|}{OOM}
        & \multicolumn{4}{c}{OOM}\\

        \hline

        \rowcolor{gray!20}
        \rule[-1.5ex]{0pt}{4ex} \textbf{\model{}} 
        & \textbf{48.42$_{\pm 0.08}$} & \textbf{37.66$_{\pm 0.47}$} & \textbf{44.13$_{\pm 0.59}$} & \textbf{31.52$_{\pm 0.41}$}
        & \textbf{87.92$_{\pm 0.12}$} & \textbf{90.97$_{\pm 0.08}$} & \textbf{91.75$_{\pm 0.24}$} & \textbf{86.48$_{\pm 0.41}$} 
        & \textbf{55.61$_{\pm 0.02}$} & \textbf{25.65$_{\pm 0.67}$} & \textbf{43.88$_{\pm 0.39}$} & \textbf{27.82$_{\pm 0.10}$}\\

        \hline

        \textbf{Improv.(\%)}
        & $\uparrow$ 3.25\% & $\uparrow$ 6.92\% & $\uparrow$ 1.04\% & $\uparrow$ 17.08\%
        & $\uparrow$ 0.48\% & $\uparrow$ 0.30\% & $\uparrow$ 0.71\% & $\uparrow$ 1.39\%
        & $\uparrow$ 0.93\% & $\uparrow$ 11.50\% & $\uparrow$ 3.24\% & $\uparrow$ 0.78\% \\

        \bottomrule
    \end{tabular}
    \vspace{-2mm}
\end{table*}

\begin{figure}[t]
	\centering
	\includegraphics[width=0.95\linewidth]{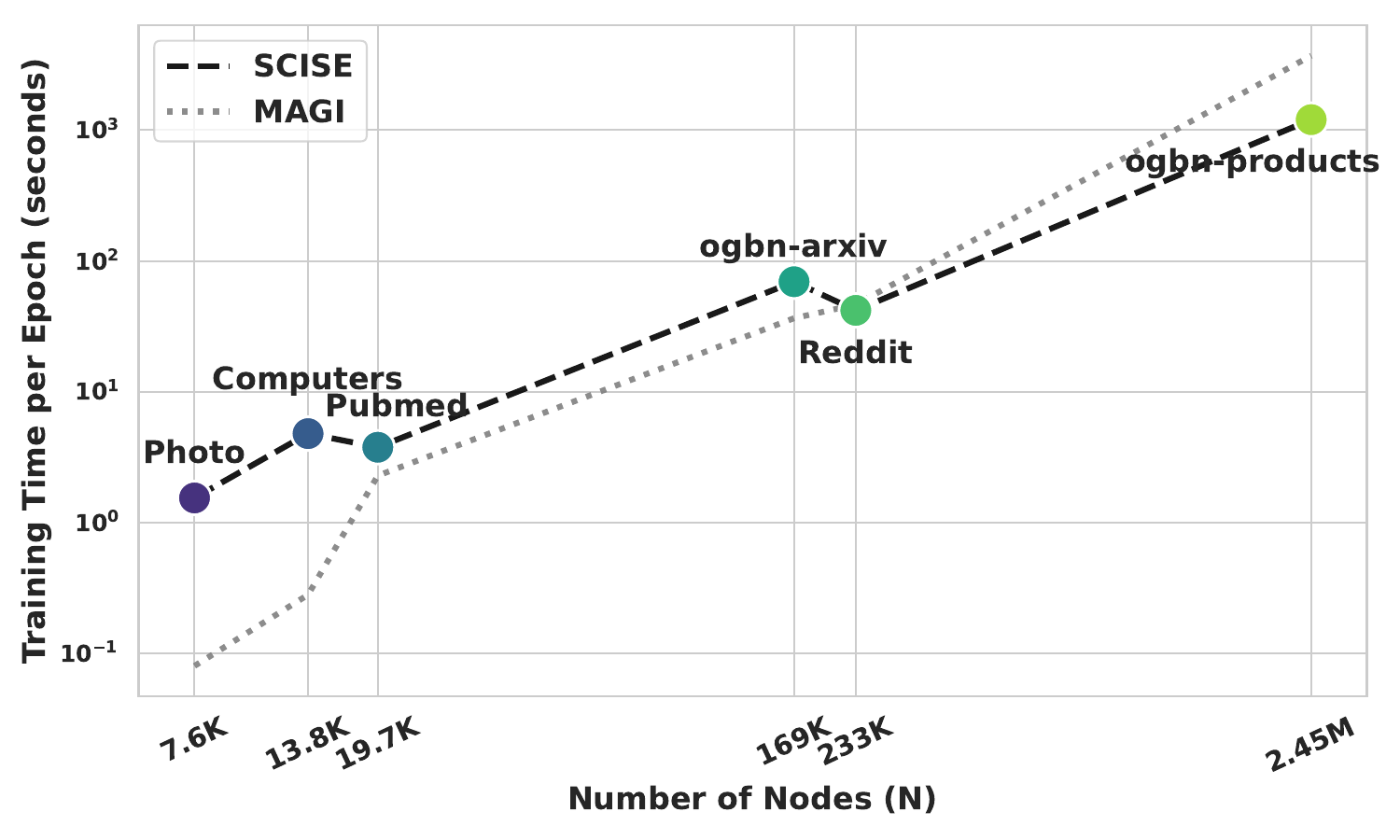}
    \vspace{-5mm}
	\caption{Runtime comparison between \model{} and MAGI.}
	\label{fig: time}
    \vspace{-7mm}
\end{figure}

\vspace{-2mm}
\subsection{Experiment Setup}
\textbf{Datasets.}
We conduct experiments on six benchmark datasets: Photo~\cite{shchur2018pitfalls}, Computers~\cite{shchur2018pitfalls}, Pubmed~\cite{sen2008pubmed}, Ogbn-arxiv~\cite{hu2020ogbn}, Reddit~\cite{hamilton2017inductive}, and Ogbn-products~\cite{hu2020ogbn}.
Details of datasets are summarized in Table~\ref{tab: dataset}.\par

\noindent \textbf{Baselines.}
For graph clustering, we mainly compare~\model{} with ten state-of-the-art baselines, including DGI~\cite{velivckovic2018deep}, S3GC~\cite{devvrit2022s3gc}, SUBLIME~\cite{liu2022sublime}, CONVERT~\cite{yang2023convert}, DMoN~\cite{tsitsulin2023graph}, Dink-Net~\cite{liu2023dink}, LSEnet~\cite{sun2024lsenet}, MAGI~\cite{liu2024magi}, DeSE~\cite{zhang2025dese}, and SCGC~\cite{kulatilleke2025scgc}.\par

\noindent \textbf{Evaluation Metrics.}
We evaluate the accuracy and consistency of graph clustering with four metrics.
NMI (Normalized Mutual Information) evaluates how well the predicted clusters match the true clusters in terms of information shared.
ARI (Adjusted Rand Index) assesses the similarity between the predicted and true cluster assignments, adjusting for random chance.
ACC (Accuracy) measures the proportion of nodes correctly assigned to their true clusters.
F1 Score evaluates the balance between precision and recall in cluster assignments.\par

\vspace{-3mm}
\subsection{Overall Performance}
\label{sec: cluster}

\begin{table*}[t]
    \centering
    \caption{Ablation study on six datasets. \colorbox{red!20}{Red cells} indicate a significant performance degradation compared to the full model.}
    \label{tab: ablation}
    \vspace{-3mm}
    \renewcommand\arraystretch{1.1}
    \setlength{\tabcolsep}{0.2mm}
    \begin{tabular}{c|cccc|cccc|cccc}
        \toprule
        \multirow{2}*{Variation} 
        & \multicolumn{4}{c|}{\multirow{1}*{\textbf{Photo}}} 
        & \multicolumn{4}{c|}{\multirow{1}*{\textbf{Computers}}} 
        & \multicolumn{4}{c}{\multirow{1}*{\textbf{Pubmed}}}\\
        
        \cline{2-13}

        & {\textbf{NMI}} & {\textbf{ARI}} & {\textbf{ACC}} & \multicolumn{1}{c|}{\textbf{F1}} 
        & {\textbf{NMI}} & {\textbf{ARI}} & {\textbf{ACC}} & \multicolumn{1}{c|}{\textbf{F1}} 
        & {\textbf{NMI}} & {\textbf{ARI}} & {\textbf{ACC}} & \multicolumn{1}{c}{\textbf{F1}} \\

        \hline

        \rowcolor{gray!20}
        \textbf{\model{}} 
        & 74.53$_{\pm 0.14}$ & 64.51$_{\pm 0.29}$ & 79.71$_{\pm 0.15}$ & 76.22$_{\pm 0.12}$
        & 58.46$_{\pm 0.24}$ & 51.78$_{\pm 0.80}$ & 67.32$_{\pm 0.36}$ & 59.48$_{\pm 0.84}$
        & 35.77$_{\pm 0.02}$ & 36.33$_{\pm 0.02}$ & 72.38$_{\pm 0.01}$ & 71.83$_{\pm 0.01}$\\

        \hline

        w/o SECC

        & 74.41$_{\pm 0.14}$ & 64.51$_{\pm 0.10}$ & 79.74$_{\pm 0.07}$ & 76.16$_{\pm 0.29}$ 
        & 58.11$_{\pm 0.23}$ & 51.07$_{\pm 0.38}$ & 67.05$_{\pm 0.26}$ & 58.97$_{\pm 0.13}$ 
        & 35.59$_{\pm 0.01}$ & 36.20$_{\pm 0.02}$ & 72.31$_{\pm 0.01}$ & 71.72$_{\pm 0.01}$ \\

        w/o CSampE
        
        & 73.62$_{\pm 0.71}$ & 63.79$_{\pm 0.43}$ & 79.48$_{\pm 0.14}$ & 75.44$_{\pm 0.93}$
        & \cellcolor{red!20}49.00$_{\pm 0.04}$ & \cellcolor{red!20}45.12$_{\pm 0.08}$ & 64.67$_{\pm 0.16}$ & 58.41$_{\pm 0.06}$ 
        & 35.87$_{\pm 0.02}$ & 36.12$_{\pm 0.01}$ & 72.19$_{\pm 0.01}$ & 71.68$_{\pm 0.02}$ \\
        
        w/o StructCL
        
        & 74.79$_{\pm 0.13}$ & 64.72$_{\pm 0.09}$ & 79.82$_{\pm 0.05}$ & 76.31$_{\pm 0.06}$
        & 55.69$_{\pm 0.06}$ & \cellcolor{red!20}39.12$_{\pm 0.10}$ & \cellcolor{red!20}55.72$_{\pm 0.06}$ & \cellcolor{red!20}52.03$_{\pm 0.02}$ 
        & \cellcolor{red!20}28.59$_{\pm 0.02}$ & \cellcolor{red!20}30.32$_{\pm 0.03}$ & \cellcolor{red!20}68.73$_{\pm 0.02}$ & \cellcolor{red!20}67.48$_{\pm 0.04}$ \\

        {\footnotesize w/o SECC \& StructCL}
        
        & 74.77$_{\pm 0.13}$ & 64.79$_{\pm 0.14}$ & 79.88$_{\pm 0.09}$ & 76.37$_{\pm 0.06}$
        & 56.45$_{\pm 0.01}$ & \cellcolor{red!20}40.61$_{\pm 0.02}$ & \cellcolor{red!20}57.14$_{\pm 0.01}$ & \cellcolor{red!20}49.75$_{\pm 0.01}$
        & \cellcolor{red!20}25.62$_{\pm 0.01}$ & \cellcolor{red!20}24.89$_{\pm 0.01}$ & \cellcolor{red!20}65.90$_{\pm 0.02}$ & \cellcolor{red!20}66.96$_{\pm 0.02}$ \\

        {\footnotesize w/o CSampE \& StructCL} 

        & \cellcolor{red!20}70.39$_{\pm 0.60}$ & \cellcolor{red!20}62.66$_{\pm 0.60}$ & 79.45$_{\pm 0.37}$ & 77.10$_{\pm 0.42}$
        & \cellcolor{red!20}50.24$_{\pm 0.59}$ & \cellcolor{red!20}35.18$_{\pm 0.76}$ & \cellcolor{red!20}54.57$_{\pm 0.22}$ & \cellcolor{red!20}55.82$_{\pm 0.82}$
        & \cellcolor{red!20}31.07$_{\pm 0.02}$ & 36.76$_{\pm 0.01}$ & 72.34$_{\pm 0.01}$ & 70.57$_{\pm 0.01}$ \\

        \bottomrule

        \multirow{2}*{Variation} 
        & \multicolumn{4}{c|}{\multirow{1}*{\textbf{Ogbn-arxiv}}} 
        & \multicolumn{4}{c|}{\multirow{1}*{\textbf{Reddit}}} 
        & \multicolumn{4}{c}{\multirow{1}*{\textbf{Ogbn-products}}}\\
        
        \cline{2-13}

        & {\textbf{NMI}} & {\textbf{ARI}} & {\textbf{ACC}} & \multicolumn{1}{c|}{\textbf{F1}} 
        & {\textbf{NMI}} & {\textbf{ARI}} & {\textbf{ACC}} & \multicolumn{1}{c|}{\textbf{F1}} 
        & {\textbf{NMI}} & {\textbf{ARI}} & {\textbf{ACC}} & \multicolumn{1}{c}{\textbf{F1}} \\

        \hline

        \rowcolor{gray!20}
        \textbf{\model{}}
        & 48.42$_{\pm 0.08}$ & 37.66$_{\pm 0.47}$ & 44.13$_{\pm 0.59}$ & 31.52$_{\pm 0.41}$
        & 87.92$_{\pm 0.12}$ & 90.97$_{\pm 0.08}$ & 91.75$_{\pm 0.24}$ & 86.48$_{\pm 0.41}$ 
        & 55.61$_{\pm 0.02}$ & 25.65$_{\pm 0.67}$ & 43.88$_{\pm 0.39}$ & 27.82$_{\pm 0.10}$\\

        \hline
                
        w/o SECC
        
        & 48.15$_{\pm 0.07}$ & 36.95$_{\pm 0.37}$ & 44.09$_{\pm 0.21}$ & 31.14$_{\pm 0.37}$
        & 87.45$_{\pm 0.04}$ & 90.72$_{\pm 0.11}$ & 91.14$_{\pm 0.18}$ & 84.51$_{\pm 0.52}$ 
        & 55.33$_{\pm 0.01}$ & 24.27$_{\pm 0.01}$ & 42.89$_{\pm 0.01}$ & 27.63$_{\pm 0.01}$ \\

        w/o CSampE
        
        & \cellcolor{red!20}45.31$_{\pm 0.33}$ & \cellcolor{red!20}21.15$_{\pm 1.11}$ & \cellcolor{red!20}32.73$_{\pm 0.95}$ & \cellcolor{red!20}26.10$_{\pm 0.61}$
        & 85.97$_{\pm 0.16}$ & \cellcolor{red!20}80.19$_{\pm 1.37}$ & \cellcolor{red!20}85.66$_{\pm 1.10}$ & \cellcolor{red!20}83.40$_{\pm 0.57}$ 
        & 53.21$_{\pm 0.67}$ & \cellcolor{red!20}21.76$_{\pm 1.17}$ & \cellcolor{red!20}39.40$_{\pm 1.06}$ & 26.13$_{\pm 0.45}$ \\
        
        w/o StructCL

        & 47.55$_{\pm 0.17}$ & \cellcolor{red!20}32.97$_{\pm 0.31}$ & 41.42$_{\pm 0.45}$ & 30.34$_{\pm 0.73}$
        & 88.20$_{\pm 0.03}$ & 91.27$_{\pm 0.08}$ & 91.57$_{\pm 0.13}$ & 85.82$_{\pm 0.19}$
        & 54.78$_{\pm 0.24}$ & \cellcolor{red!20}22.59$_{\pm 0.26}$ & \cellcolor{red!20}40.62$_{\pm 0.18}$ & 26.71$_{\pm 0.11}$\\

        {\footnotesize w/o SECC \& StructCL}

        & 46.91$_{\pm 0.21}$ & \cellcolor{red!20}28.08$_{\pm 1.77}$ & \cellcolor{red!20}38.35$_{\pm 0.65}$ & 29.82$_{\pm 1.01}$
        & 87.46$_{\pm 0.11}$ & 90.36$_{\pm 0.27}$ & 90.69$_{\pm 0.28}$ & \cellcolor{red!20}83.42$_{\pm 0.90}$
        & 54.80$_{\pm 0.01}$ & \cellcolor{red!20}22.35$_{\pm 0.01}$ & \cellcolor{red!20}40.51$_{\pm 0.01}$ & 26.73$_{\pm 0.02}$\\

        {\footnotesize w/o CSampE \& StructCL}
        
        & \cellcolor{red!20}41.49$_{\pm 0.37}$ & \cellcolor{red!20}15.57$_{\pm 0.10}$ & \cellcolor{red!20}28.25$_{\pm 0.77}$ & \cellcolor{red!20}22.95$_{\pm 0.58}$
        & 85.44$_{\pm 0.04}$ & \cellcolor{red!20}80.09$_{\pm 0.54}$ & \cellcolor{red!20}85.71$_{\pm 0.84}$ & \cellcolor{red!20}82.61$_{\pm 0.50}$
        & \cellcolor{red!20}48.53$_{\pm 0.22}$ & \cellcolor{red!20}15.83$_{\pm 1.65}$ & \cellcolor{red!20}33.92$_{\pm 1.74}$ & \cellcolor{red!20}23.07$_{\pm 0.26}$ \\

        \bottomrule
    \end{tabular}
    \vspace{-3mm}
\end{table*}
\begin{table}[t]
    \centering
    \caption{\revised{End-to-end runtime analysis of \model{}.}}
    \label{tab: time}
    \vspace{-4mm}
    \renewcommand\arraystretch{1.0}
    \setlength{\tabcolsep}{1.8mm}
    \begin{tabular}{l|cccc}
        \toprule
        \textbf{Dataset} & \textbf{SECC}  & \textbf{epoch} & \textbf{clustering} & \textbf{total} \\
        \hline
        
        Photo               & 0.30 s & 1.37 s & 3.34 s & 14.58 min \\
        Computers           & 0.33 s & 4.92 s & 3.48 s & 5.10 min \\
        Pubmed              & 0.35 s & 2.73 s & 3.55 s & 2.87 min \\
        Ogbn-arxiv          & 1.24 s & 73.73 s & 73.26 s & 1.64 h \\
        Reddit              & 2.58 s & 40.97 s & 96.59 s & 1.80 h \\
        Ogbn-products       & 10.75 s & 1028.96 s & 245.21 s & 7.54 h \\
                
        \bottomrule
    \end{tabular}
    \vspace{-1mm}
\end{table}

Table~\ref{tab: results} presents the comprehensive performance evaluation, where \model{} dominates across all benchmarks, achieving state-of-the-art (SOTA) results in 21 out of 24 metrics. On the Ogbn-arxiv dataset, our framework significantly outperforms the strongest baselines by 3.25\% in NMI and a remarkable 17.08\% in F1-score. 
We attribute this localized and global success to three synergistic strengths:
\textbf{First, Countering Structural Drifting and Fragmentation}: Unlike existing structural methods (e.g., DMoN, DeSE) that often suffer from uncontrolled cluster merging or excessive fragmentation, \model{} employs the SECC operator to enforce explicit community-cardinality constraints. This yields partitions that are not only cohesive but also strictly aligned with the network's inherent semantic distribution.
\textbf{Second, Bridging Structural Isolation}: While conventional self-supervised methods (e.g., DGI, SUBLIME) rely on node-local consistency, they frequently lose global community context during batching. \model{} mitigates this via CSampE, which proactively restores community-level connectivity, and StructCL, which recalibrates higher-order dependencies. This combination ensures that the encoder captures holistic topological patterns rather than isolated neighborhood signals.
\textbf{Third, Robust Scalability on Massive Graphs}: On large-scale benchmarks like Reddit and Ogbn-products, many high-performance competitors encounter Out-of-Memory (OOM) errors due to dense graph operations. In contrast, \model{} scales efficiently to millions of nodes by decoupling the community-aware expansion from the backpropagation pipeline. This allows our framework to maintain global topological awareness even under strict memory constraints—a critical advantage for real-world deployment.
Notably, this demonstrates that our framework preserves global topological awareness under memory constraints—a capability lacking in existing scalable methods. 
Collectively, these results demonstrate that the synergy between structural entropy constraints and community-aware sampling is fundamental for high-performance graph clustering at scale.\par

\revised{Furthermore, Figure~\ref{fig: time} reports the average per-epoch runtime of \model{} and MAGI. 
As the graph size increases from thousands to millions of nodes, the runtime of \model{} exhibits a near-linear growth trend, demonstrating good scalability on large-scale graphs. 
Although \model{} incurs a moderate overhead on small datasets due to community-aware batch expansion, it becomes increasingly competitive on larger graphs and achieves lower per-epoch runtime than MAGI on Reddit and Ogbn-products.
To provide a complete end-to-end efficiency analysis, Table~\ref{tab: time} further reports the preprocessing (SECC), training, clustering, and total runtime costs. 
The results show that SECC introduces only a negligible one-time overhead, requiring less than 3 seconds on all datasets except Ogbn-products (10.75 s), which is insignificant compared with the overall training time. 
Even on Ogbn-products with 2.45 million nodes, the entire pipeline can be completed within 7.54 hours.
Table~\ref{tab: memory} reports the memory consumption of both SECC and the training stage. 
Despite operating on large-scale graphs, the memory footprint of SECC remains substantially lower than that of representation learning, confirming that the proposed community discovery step is not the computational bottleneck of the framework. 
These results demonstrate that \model{} achieves strong clustering performance while maintaining practical end-to-end efficiency and scalability for million-scale graph clustering.\par}

\begin{table}[t]
    \centering
    \caption{\revised{Memory usage breakdown of \model{} (GPU A and GPU R denote allocated and reserved GPU memory, respectively).}}
    \label{tab: memory}
    \vspace{-4mm}
    \footnotesize
    \renewcommand\arraystretch{1.2}
    \setlength{\tabcolsep}{0.7mm}
    \begin{tabular}{l|rrr|rrr}
        \toprule
        \multirow{2}*{Dataset} 
        & \multicolumn{3}{c|}{\multirow{1}*{\textbf{SECC}}} 
        & \multicolumn{3}{c}{\multirow{1}*{\textbf{Train}}}\\

        \cline{2-7}

        & \textbf{GPU A} & \textbf{GPU R} & \textbf{CPU}
        & \textbf{GPU A} & \textbf{GPU R} & \textbf{CPU} \\
        \hline
        
        Photo               & 41.77 MB & 78.00 MB & 1.07 MB & 4.30 GB & 14.62 GB & 62.46 MB \\
        Computers           & 87.88 MB & 136.00 MB & 1.30 MB & 3.43 GB & 13.00 GB & 30.34MB \\
        Pubmed              & 18.24 MB & 36.00 MB & 1.50 MB     & 5.79 GB & 16.29 GB & 23.28 MB \\
        Ogbn-arxiv          & 424.40 MB & 591.99 MB & 17.19 MB  & 19.01 GB & 46.84 GB & 356.35 MB \\
        Reddit              & 11.30 GB & 11.90 GB & 2.46 MB     & 11.03 GB & 31.17 GB & 475.02 MB \\
        Ogbn-products       & 18.71 GB & 24.66 GB & 113.06 MB   & 39.43 GB & 46.79 GB & 316.83 MB \\
                
        \bottomrule
    \end{tabular}
    \vspace{-1mm}
\end{table}

\vspace{-4mm}
\subsection{Ablation Study: Dissecting the Synergistic and Complementary Effects}
\label{sec: ablation}

\begin{figure*}[t]
	\centering
    \subfigure[Photo.]{
		\includegraphics[width=0.24\linewidth]{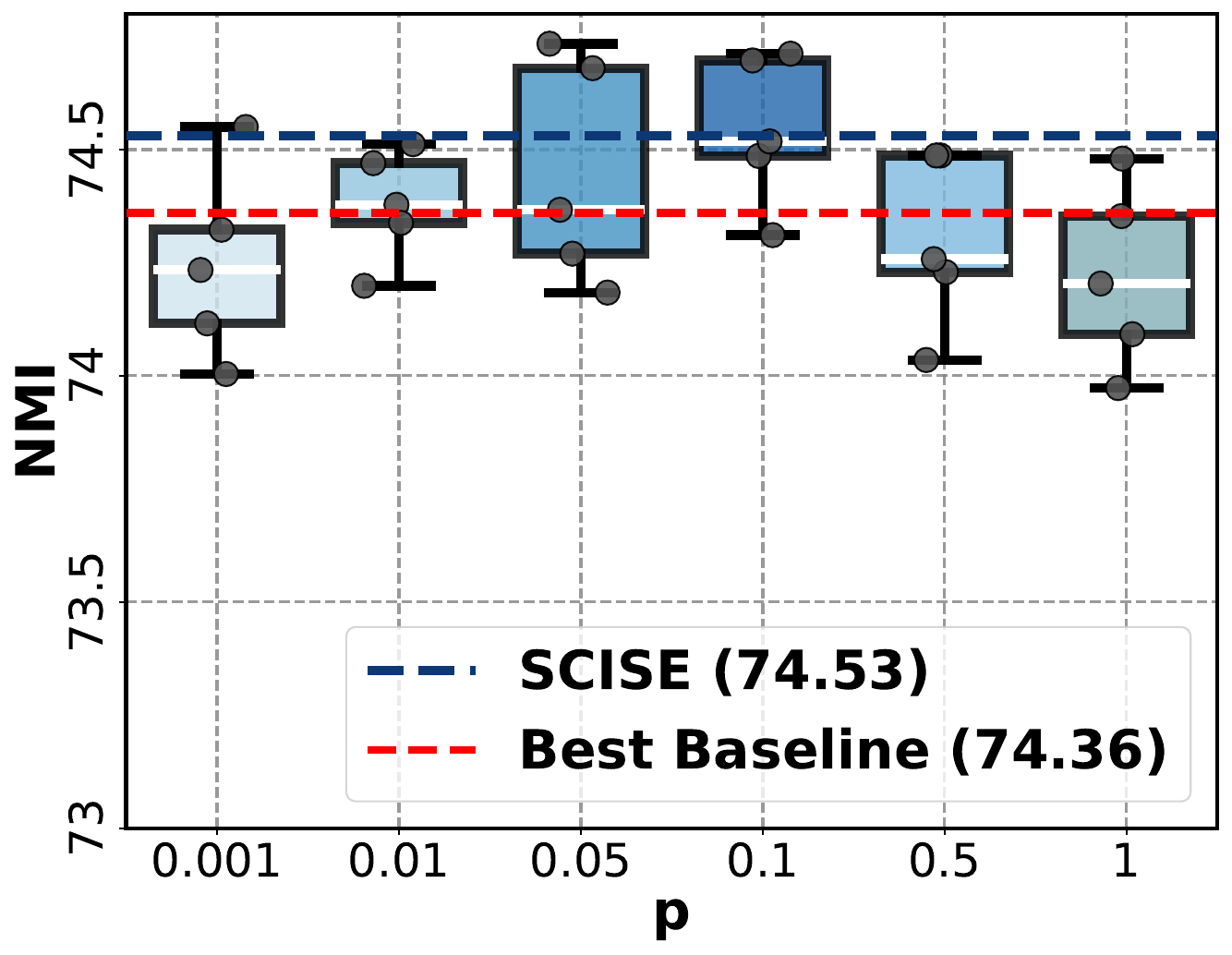}
		\label{fig: hyper_p_photo}
	}\hfill
	\subfigure[Computers.]{
		\includegraphics[width=0.24\linewidth]{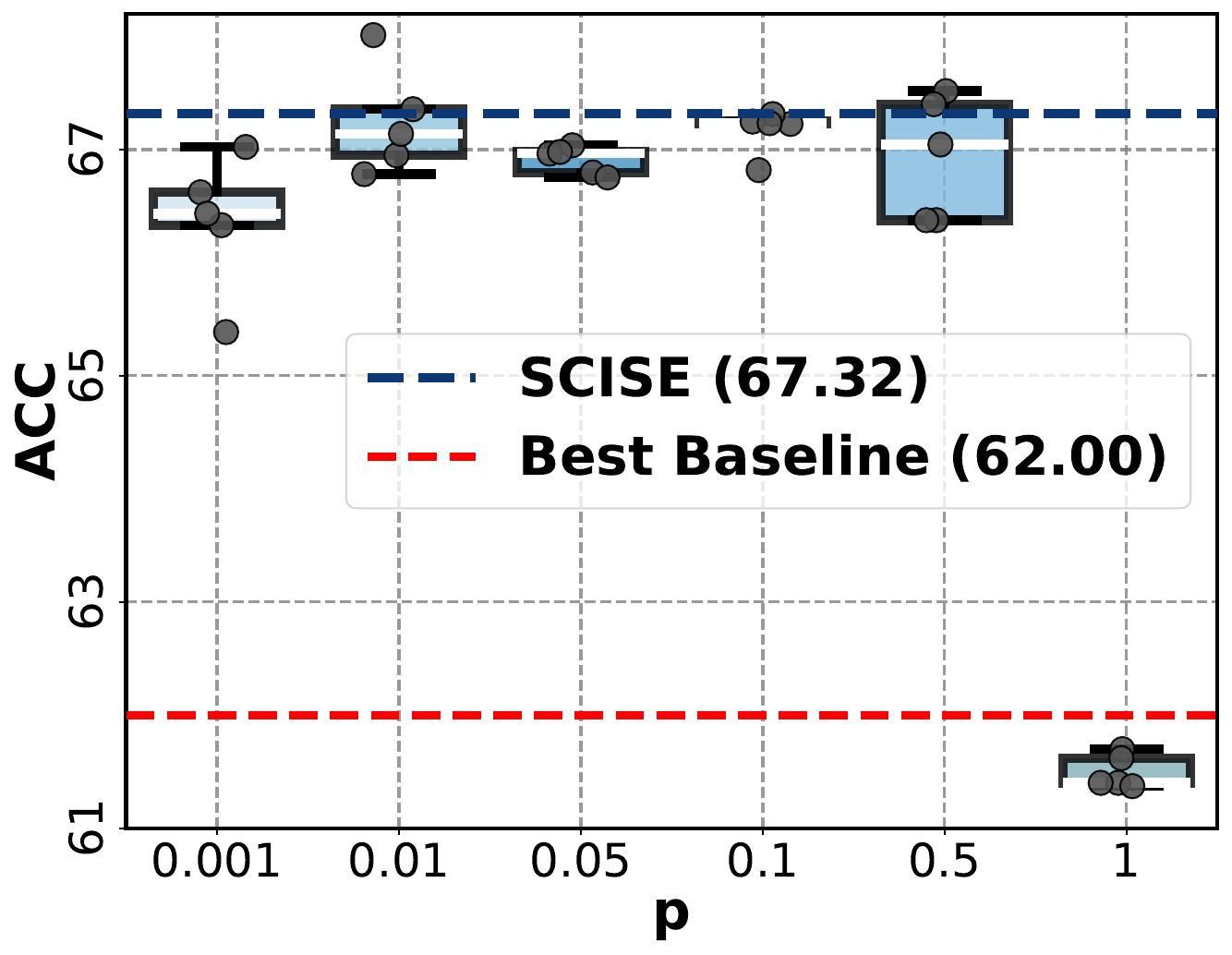}
		\label{fig: hyper_p_computers}
	}\hfill
	\subfigure[Ogbn-arxiv.]{
		\includegraphics[width=0.24\linewidth]{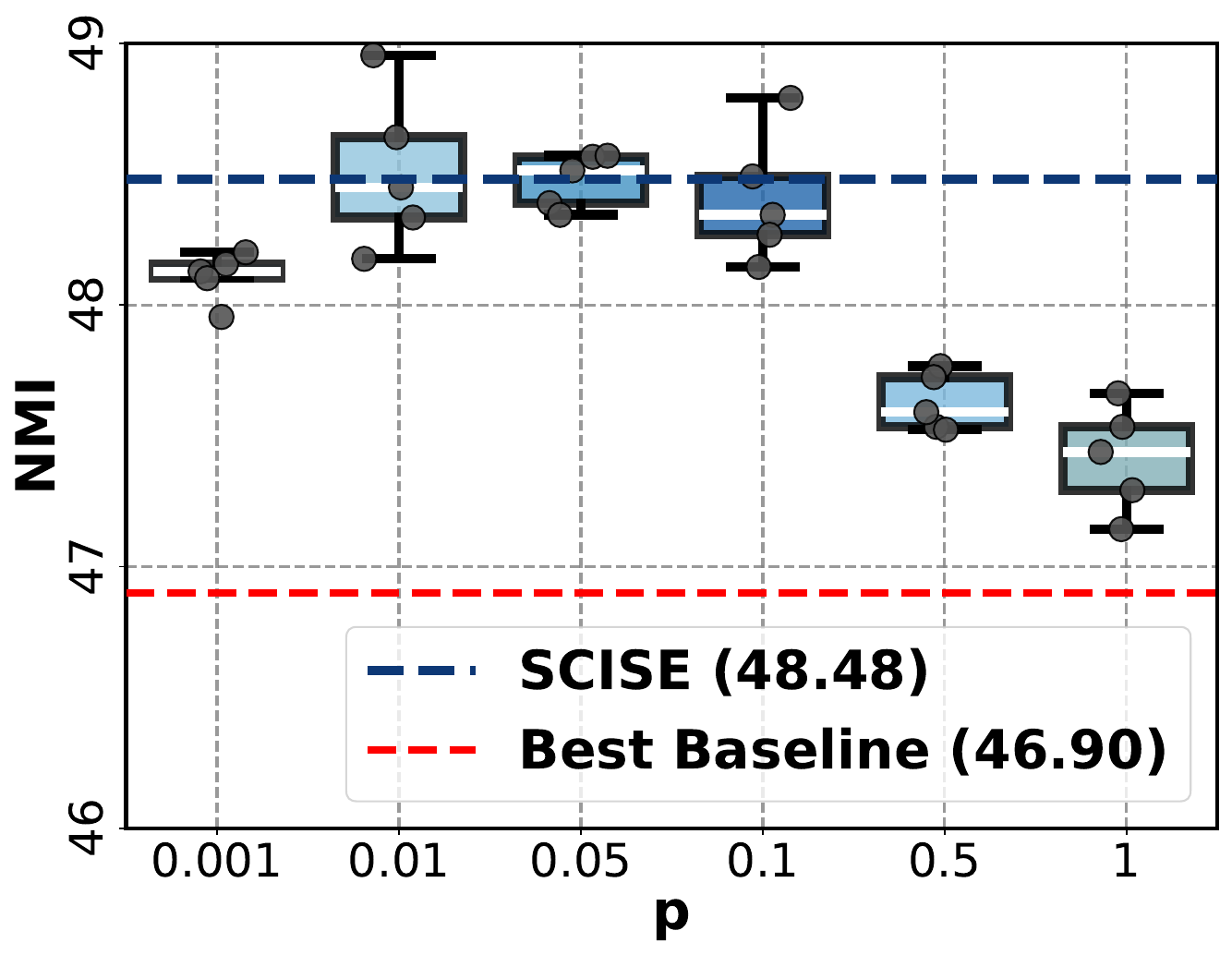}
		\label{fig: hyper_p_arxiv}
	}\hfill
	\subfigure[Reddit.]{
		\includegraphics[width=0.24\linewidth]{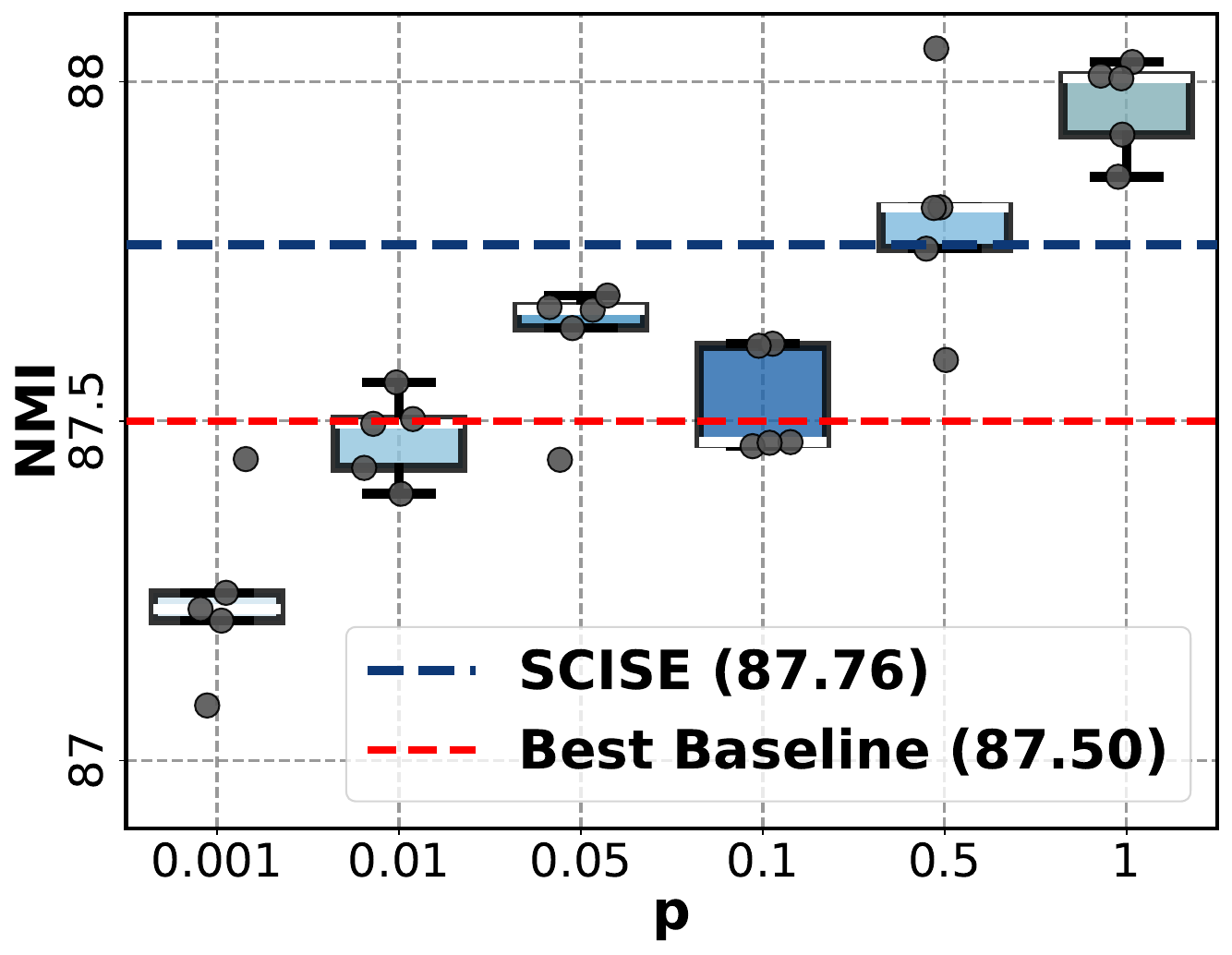}
		\label{fig: hyper_p_reddit}
	}
    \vspace{-5mm}
	\caption{Sensitivity analysis of merge speed $p$ on Photo, Computers, Ogbn-arxiv, and Reddit datasets.}
	\label{fig: hyper_p}
    \vspace{-4mm}
\end{figure*}

\begin{figure*}[t]
	\centering
	\subfigure[Photo.]{
		\includegraphics[width=0.24\linewidth]{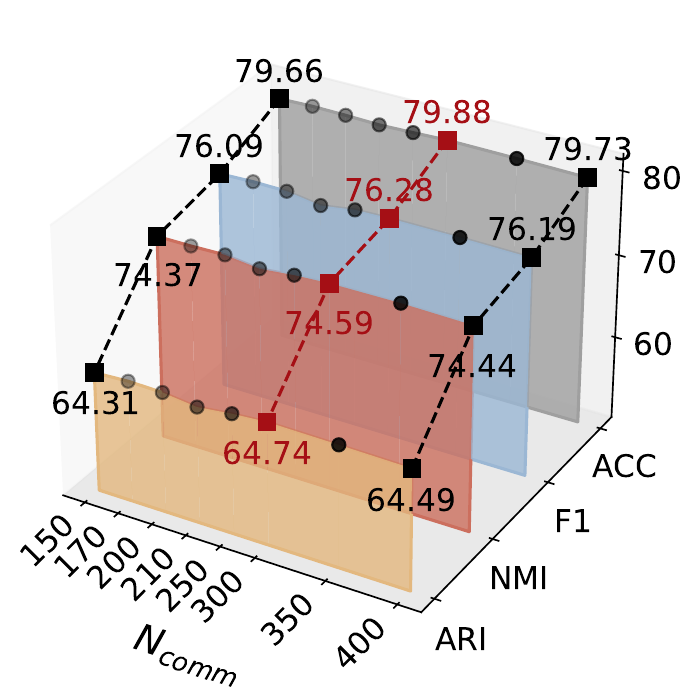}
		\label{fig: hyper_commNum_photo}
	}\hfill
	\subfigure[Computers.]{
		\includegraphics[width=0.24\linewidth]{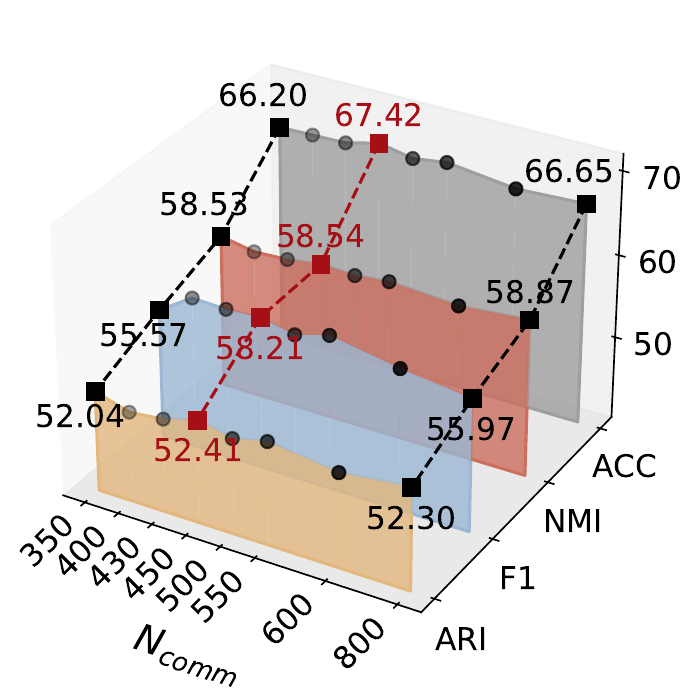}
		\label{fig: hyper_commNum_computers}
	}\hfill
    \subfigure[Ogbn-arxiv.]{
		\includegraphics[width=0.24\linewidth]{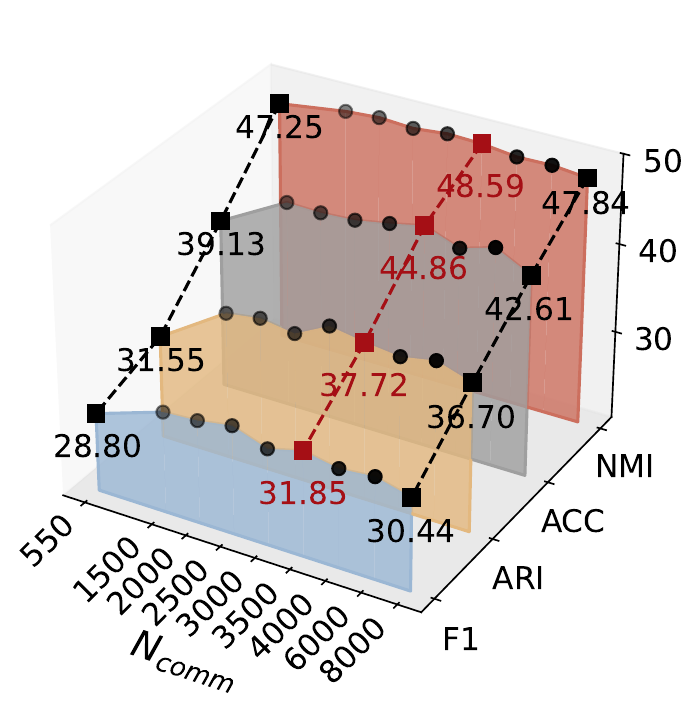}
		\label{fig: hyper_commNum_arxiv}
	}\hfill
    \subfigure[Reddit.]{
		\includegraphics[width=0.24\linewidth]{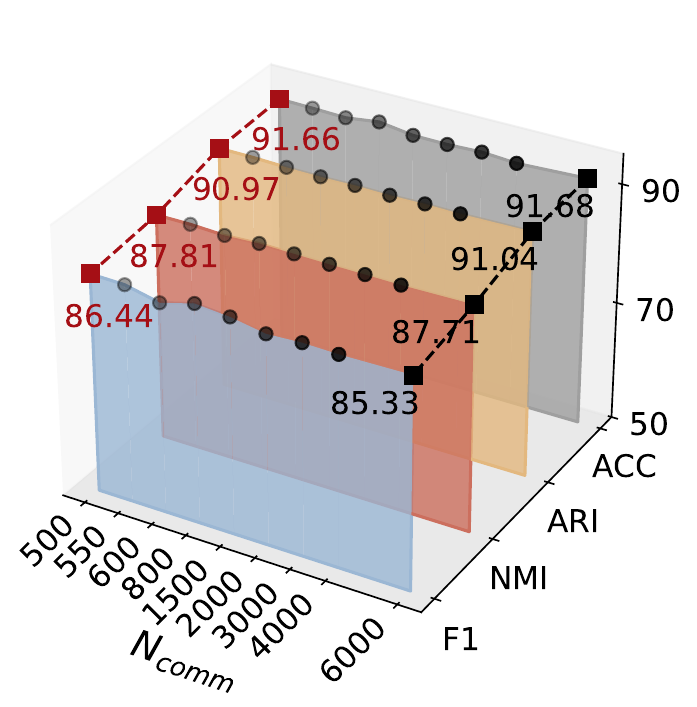}
		\label{fig: hyper_commNum_reddit}
	}   
    \vspace{-5mm}
	\caption{Sensitivity analysis of community number $N_{comm}$ on Photo, Computers, Ogbn-arxiv, and Reddit datasets.}
	\label{fig: hyper_commNum}
    \vspace{-3mm}
\end{figure*}

To evaluate the contribution of each component, we construct three primary variants ("w/o SECC", "w/o CSampE", and "w/o StructCL") and two joint variants ("w/o SECC \& StructCL" and "w/o CSampE \& StructCL"). Specifically, "w/o SECC" replaces community-constrained structural entropy minimization with unconstrained optimization, while "w/o StructCL" computes contrastive loss directly on the raw adjacency matrix. The results in Table~\ref{tab: ablation} show that all components contribute positively to the final performance.
(1) Effect of CSampE. Removing CSampE leads to the largest degradation on large-scale graphs. For example, the ARI on Ogbn-arxiv drops from 37.66\% to 21.15\%, indicating that community-aware batch expansion is crucial for alleviating structural isolation and preserving long-range topology during mini-batch training.
(2) Effect of StructCL. StructCL consistently improves clustering quality, particularly on medium-scale datasets. On Pubmed, removing StructCL decreases ARI from 36.33\% to 30.32\%, demonstrating the benefit of incorporating refined structural affinities into contrastive learning.
(3) Effect of SECC. Although the impact of SECC is generally smaller than that of CSampE, it provides a stable structural prior for downstream learning. Comparing "w/o StructCL" and "w/o SECC \& StructCL", the additional removal of SECC further reduces Pubmed NMI from 28.59\% to 25.62\%, suggesting that community-constrained structural entropy helps maintain globally consistent partitions.

\begin{table}[t]
    \centering
    \caption{Sensitivity analysis of learning rate $lr$ with NMI.}
    \label{tab: hyper_lr}
    \vspace{-3mm}
    \renewcommand\arraystretch{0.9}
    \setlength{\tabcolsep}{1.3mm}
    \begin{tabular}{c|cccccc}
        \toprule
        $lr$ & 0.05 & 0.01 & 0.005 & 0.001 & 0.0005 & 0.0001 \\
        
        \hline

        Photo
        & 71.65 & 74.31 & 74.44 & 72.69 & 72.38 & 72.50 \\

        Computers
        & 57.58 & 58.02 & 56.34 & 58.62 & 58.40 & 57.89 \\

        Pubmed
        & 23.75 & 34.86 & 35.62 & 27.97 & 27.35 & 28.61 \\

        Ogbn-arxiv
        & 48.40 & 48.35 & 48.18 & 48.40 & 48.24 & 47.77 \\

        Reddit
        & 88.31 & 88.34 & 88.06 & 88.26 & 88.36 & 87.91 \\

        Ogbn-products
        & 55.33 & 55.19 & 55.62 & 55.35 & 55.49 & 55.85 \\

        \bottomrule
    \end{tabular}
    \vspace{-2mm}
\end{table}

\vspace{-4mm}
\subsection{Stability Analysis: Assessing the Invariance to Hyper-parameter Variations}
\label{sec: hyperparameter}

To further evaluate the robustness of \model{}, we analyze the sensitivity of seven key hyperparameters, including the merge speed $p$, community number $N_{comm}$, learning rate $lr$, random walk parameters ($w_t$, $w_l$), expansion threshold $\theta$, and initial batch size $B$. 
Overall, \model{} exhibits stable performance across a wide range of settings, indicating low hyperparameter sensitivity and minimal tuning effort. \par

\textbf{Merge speed $p$:}
\revised{Figure~\ref{fig: hyper_p} shows that the performance remains stable across $p\in\{0.001,0.01,0.05,0.1,0.5,1\}$. 
Only extreme values lead to noticeable fluctuations on a few datasets, while most settings consistently outperform the strongest baseline.}

\begin{table}[t]
    \centering
    \caption{\revised{Sensitivity analysis of hyper $w_t$ with NMI.}}
    \label{tab: hyper_wt}
    \vspace{-4mm}
    \renewcommand\arraystretch{1.0}
    \setlength{\tabcolsep}{0.8mm}
    \begin{tabular}{c|cccccccc}
        \toprule
        $w_t$ & 10 & 20 & 30 & 40 & 50 & 60 & 80 & 100 \\
        
        \hline
        
        Photo
        & 73.87 & 74.00 & 74.21 & 74.12 & 74.40 & 74.42 & 74.18 & 74.15 \\

        Computers
        & 56.57 & 58.40 & 57.82 & 56.50 & 56.54 & 56.60 & 56.42 & 56.50 \\

        Pubmed
        & 35.36 & 35.58 & 35.62 & 35.54 & 35.73 & 35.21 & 35.73 & 35.35 \\

        Arxiv
        & 48.28 & 48.34 & 48.35 & 48.50 & 47.98 & 48.53 & 48.35 & 48.27 \\

        Reddit
        & 86.31 & 87.46 & 88.35 & 88.13 & 88.15 & 88.08 & 87.75 & 87.72 \\

        Products
        & 54.63 & 55.34 & 55.52 & OOM & OOM & OOM & OOM & OOM \\

        \bottomrule
    \end{tabular}
    \vspace{-4mm}
\end{table}
\begin{table}[t]
    \centering
    \caption{\revised{Sensitivity analysis of hyper $w_l$ with NMI.}}
    \label{tab: hyper_wl}
    \vspace{-4mm}
    \renewcommand\arraystretch{1.0}
    \setlength{\tabcolsep}{0.85mm}
    \begin{tabular}{c|cccccccc}
        \toprule
        $w_l$ & 2 & 3 & 4 & 5 & 6 & 7 & 8 & 10 \\
        
        \hline

        Photo
        & 74.49 & 74.01 & 73.43 & 72.57 & 71.95 & 71.28 & 70.88 & 69.94\\

        Computers
        & 56.20 & 56.81 & 57.13 & 57.26 & 59.02 & 58.03 & 58.52 & 58.24\\

        Pubmed
        & 35.62 & 33.67 & 31.34 & 30.87 & 33.25 & 33.10 & 33.54 & 33.99 \\

        Ogbn-arxiv
        & 48.60 & 48.40 & 48.46 & 48.01 & 47.93 & 47.97 & 48.02 & 47.98 \\

        Reddit
        & 87.56 & 88.25 & 88.28 & 87.99 & 87.72 & 87.34 & 87.04 & 86.58 \\

        products
        & 54.18 & 55.24 & 55.27 & 55.60 & 55.77 & 55.96 & OOM & OOM \\

        \bottomrule
    \end{tabular}
    \vspace{-2mm}
\end{table}

\textbf{Community number $N_{comm}$:}
\revised{As shown in Figure~\ref{fig: hyper_commNum}, the performance curves remain relatively flat across a broad range of community numbers. 
Except for a localized drop on Ogbn-arxiv, the model maintains stable clustering quality, suggesting that \model{} is largely insensitive to the exact choice of $N_{comm}$.}

\begin{table*}[t]
    \centering
    \caption{\revised{Performance comparison under varying community noise ratios (NMI+ARI+ACC+F1).}}
    \label{tab: noise}
    \small
    \vspace{-3mm}
    \renewcommand\arraystretch{1.0}
    \setlength{\tabcolsep}{1.0mm}
    \begin{tabular}{c|>{\columncolor{gray!25}}c>{\columncolor{gray!10}}c|cccccccc|ccccccc}
        \toprule

        Dataset 
        & \textbf{\model{}}
        & \textbf{MAGI}
        & \multicolumn{8}{c|}{\multirow{1}*{\textbf{random noise}}}  
        & \multicolumn{7}{c}{\multirow{1}*{\textbf{neighbor noise}}} \\
        
        \midrule
        
        Rate
        & 0\% & 0\%
        
        & 1\% & 2\% & 4\% & 6\% & 8\% & 10\% & 20\%  & 100\%
        & 1\% & 2\% & 4\% & 6\% & 8\% & 10\% & 20\%    \\
        
        \midrule

        \footnotesize Photo 
        & 294.96 & 285.00 
        & 290.44 & 289.06 & 290.61 & 294.60 & 293.80 & 288.44 & 292.33 & 291.29 
        & 289.94 & 290.39 & 292.08 & 294.73 & 294.03 & 290.84 & 288.12 \\

        Computers 
        & 237.03 & 224.80 
        & 236.71 & 244.05 & 234.89 & 234.47 & 236.23 & 231.30 & 227.53 & 209.67 
        & 235.22 & 236.31 & 236.47 & 234.46 & 239.82 & 237.28 & 231.89 \\ 

        Pubmed 
        & 216.32 & 189.96 
        & 186.69 & 182.17 & 195.14 & 188.09 & 195.83 & 198.05 & 189.93 & 208.10 
        & 196.47 & 181.42 & 196.02 & 183.79 & 204.18 & 195.00 & 208.01 \\ 

        Ogbn-arxiv 
        & 161.73 & 143.30 
        & 160.24 & 160.91 & 159.38 & 159.05 & 158.88 & 159.27 & 159.76 & 143.42 
        & 159.18 & 161.59 & 157.73 & 158.21 & 161.54 & 158.12 & 158.30 \\ 

        Reddit 
        & 357.12 & 354.60 
        & 359.11 & 358.43 & 357.86 & 357.38 & 357.58 & 357.04 & 357.00 & 330.97 
        & 356.96 & 357.73 & 358.21 & 357.64 & 357.84 & 356.84 & 356.63 \\ 

        Ogbn-products 
        & 152.95 & 146.70 
        & 151.82 & 150.50 & 152.31 & 152.06 & 151.27 & 150.23 & 151.81 & 136.34 
        & 148.87 & 151.62 & 149.11 & 149.82 & 148.34 & 150.03 & 151.27 \\ 
    
        \bottomrule
    \end{tabular}
    \vspace{-2mm}
\end{table*}

\begin{figure}[t]
	\centering
	\includegraphics[width=0.99\linewidth]{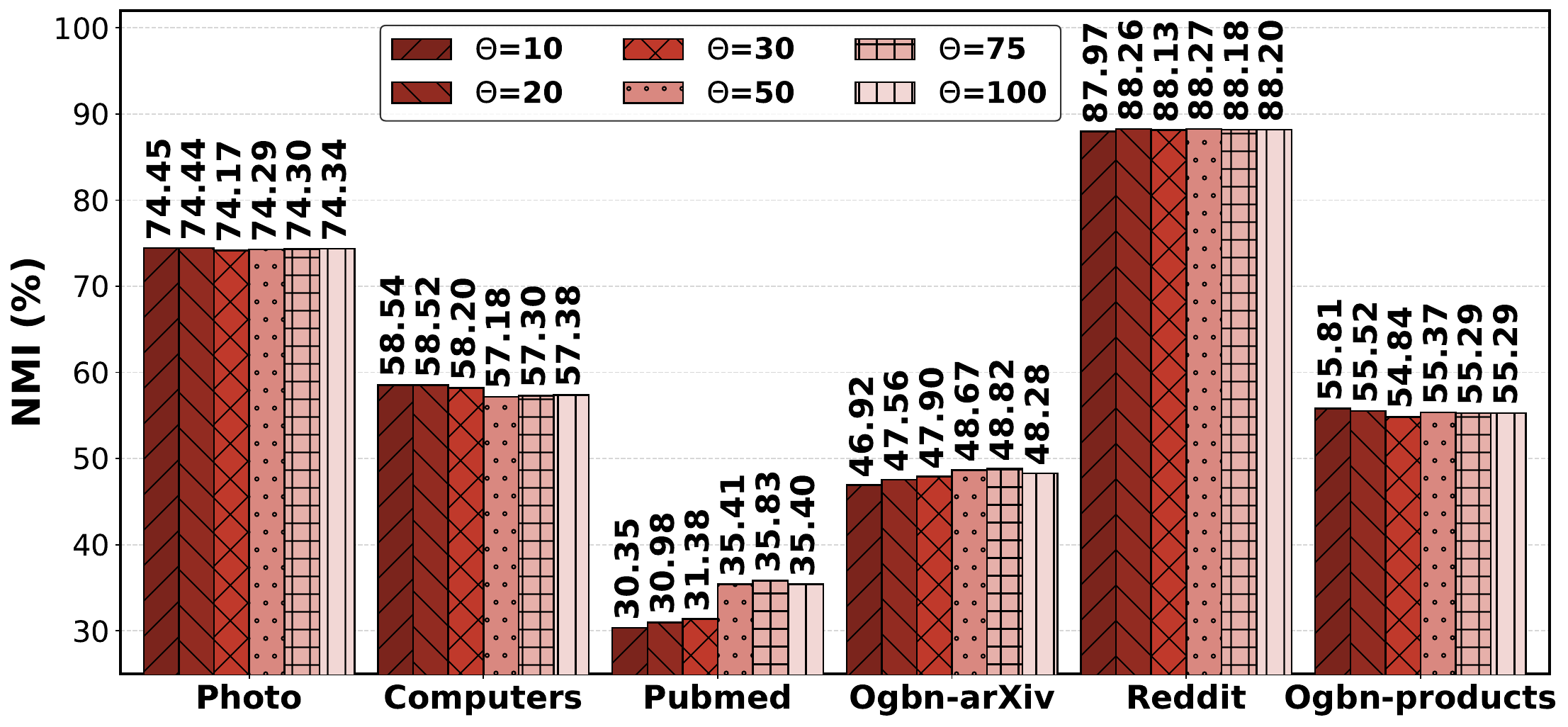}
    \vspace{-3mm}
	\caption{Sensitivity analysis of threshold $\theta$ with NMI.}
	\label{fig: hyper_theta_nmi}
    \vspace{-2mm}
\end{figure}

\textbf{Learning rate $lr$:}
As shown in Table~\ref{tab: hyper_lr}, \model{} demonstrates stable convergence across typical learning rate intervals, \revised{which further confirms that the framework is not prone to overfitting specific parameter combinations.}\par

\revised{\textbf{Random walker $w_t$ and $w_l$:}}
\revised{Tables~\ref{tab: hyper_wt} and \ref{tab: hyper_wl} show only minor performance variations when changing the number and length of random walks, suggesting that the structural affinity construction is robust to different walk configurations.}\par

\textbf{Community size threshold $\theta$:}
\revised{Figure~\ref{fig: hyper_theta_nmi} shows that \model{} maintains stable performance for $\theta\in[10,100]$ on most datasets. 
Only Pubmed exhibits a modest improvement with larger thresholds, while the overall trend remains consistent.}

\begin{figure}[t]
	\centering
	\includegraphics[width=0.99\linewidth]{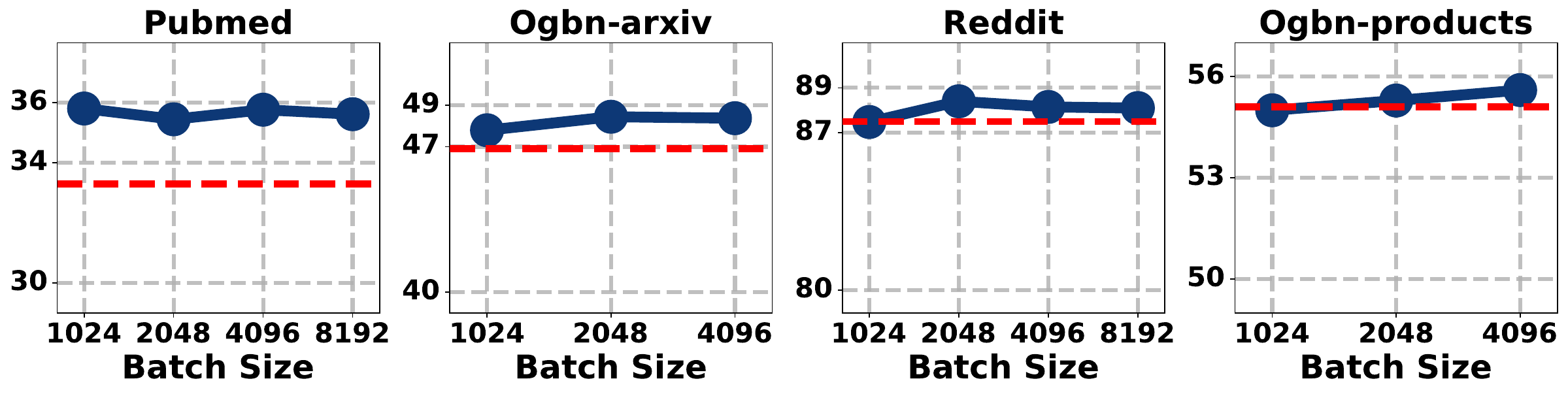}
    \vspace{-3mm}
	\caption{Sensitivity analysis of initial batch size $B$ with NMI.}
	\label{fig: hyper_batchsize_nmi}
    \vspace{-1mm}
\end{figure}

\textbf{Initial Batch size $B$:}
Figure~\ref{fig: hyper_batchsize_nmi} indicates that \model{} consistently outperforms the strongest baseline under different batch sizes. 
This suggests that the framework is not strongly dependent on a specific batch configuration and can adapt to diverse hardware constraints.

Overall, \model{} achieves consistently strong performance across a wide range of hyperparameter settings. 
These results indicate that the framework requires little parameter tuning in practice, making it suitable for large-scale graph clustering scenarios where prior knowledge is limited.

\vspace{-4mm}
\subsection{\revised{Robustness against Noisy Community Priors}}
\label{sec: noise}
\revised{Since \model{} relies on the community partitions generated by SECC, we further evaluate its robustness under imperfect community priors. 
Specifically, we inject two types of perturbations: \textbf{random noise}, which randomly reassigns a fraction of nodes to different communities, and \textbf{neighbor noise}, which reassigns nodes according to the communities of their neighbors.
Table~\ref{tab: noise} reports the results under varying noise ratios using the aggregated score (NMI+ARI+ACC+F1). 
Overall, \model{} exhibits strong robustness to both perturbation strategies. Even when 20\% of the nodes are assigned to incorrect communities, the performance degradation remains marginal on most datasets. 
In particular, the three large-scale benchmarks (Ogbn-arxiv, Reddit, and Ogbn-products) remain highly stable under both random and neighbor noise, indicating that moderate inaccuracies in community assignments have little impact on the learned representations.\par

We further consider an extreme setting with completely random community assignments (100\% random noise). 
Although performance decreases on all datasets, \model{} remains competitive and still achieves performance comparable to or better than MAGI on several benchmarks. 
This observation suggests that community priors serve as beneficial structural guidance rather than a strict prerequisite for effective clustering. 
The relatively small degradation on Photo can be attributed to its limited graph size, where the expanded batches already cover a substantial portion of the graph and therefore preserve sufficient structural context even under random community assignments.
Overall, these results demonstrate that \model{} benefits from community-aware priors but does not critically depend on highly accurate SECC outputs. 
Combined with the hyperparameter sensitivity analysis in Figures~\ref{fig: hyper_p} and \ref{fig: hyper_commNum}, which shows that \model{} remains stable across a wide range of SECC configurations, we conclude that the effectiveness of \model{} is robust to both imperfect community partitions and variations in the SECC preprocessing stage.\par}

\vspace{-4mm}
\subsection{\revised{Robustness against Sparsity}}
\label{sec: sparsity}

\begin{table*}[t]
    \centering
    \caption{Performance comparison (NMI and ACC) under varying edge deletion ratios.}
    \label{tab: del_rate}
    \vspace{-4mm}
    \small
    \renewcommand\arraystretch{0.85}
    \setlength{\tabcolsep}{1.4mm}
    \begin{tabular}{c|c|ccccccccc|ccccccccc}
        \toprule

        \multicolumn{2}{c|}{\multirow{1}*{}}
        & \multicolumn{9}{c|}{\multirow{1}*{\textbf{NMI}}}  
        & \multicolumn{9}{c}{\multirow{1}*{\textbf{ACC}}} \\
        
        \midrule
        
        \multicolumn{2}{c|}{\multirow{1}*{Rate}}
        & 0\% & 10\% & 20\% & 30\% & 40\% & 50\% & 60\% & 70\% & 80\% 
        & 0\% & 10\% & 20\% & 30\% & 40\% & 50\% & 60\% & 70\% & 80\% \\
        
        \midrule

        \multirow{3}*{\rotatebox{90}{\tiny\textbf{Photo}}}
        & \textbf{\model{}} 
        & 74.53
        & 74.42 & 74.29 & 73.90 & 74.09 & 74.33 & 74.14 & 73.87 & 72.82
        & 79.71
        & 79.78 & 79.77 & 79.67 & 79.61 & 79.69 & 79.73 & 79.54 & 78.62 \\

        & \textbf{w/o SECC} 
        & 74.41
        & 74.03 & 74.15 & 74.27 & 74.14 & 74.24 & 74.12 & 73.93 & 72.83 
        & 79.74
        & 79.76 & 79.67 & 79.67 & 79.74 & 79.72 & 79.57 & 79.57 & 78.76 \\

        & MAGI 
        & 71.60
        & 70.06 & 70.03 & 68.76 & 69.67 & 68.68 & 68.86 & 70.17 & 68.22
        & 79.00
        & 79.06 & 78.80 & 77.93 & 78.46 & 77.97 & 78.07 & 79.14 & 77.98 \\

        \midrule

        \multirow{3}*{\rotatebox{90}{\tiny \textbf{Computers}}}

        & \textbf{\model{}} & 58.46
        & 58.24 & 57.84 & 58.22 & 57.97 & 57.67 & 57.08 & 56.95 & 56.72
        & 67.32 
        & 66.79 & 66.98 & 66.99 & 66.84 & 66.37 & 65.98 & 66.06 & 65.10 \\

        & \textbf{w/o SECC} & 58.11
        & 58.37 & 58.11 & 58.27 & 57.66 & 57.46 & 57.13 & 56.19 & 56.24
        & 67.05 
        & 67.26 & 66.96 & 67.16 & 66.71 & 66.05 & 66.09 & 65.41 & 64.95 \\

        & MAGI & 59.20 
        & 59.01 & 59.40 & 59.50 & 59.60 & 58.71 & 58.79 & 59.36 & 59.91
        & 62.00
        & 62.02 & 61.73 & 61.44 & 62.62 & 61.52 & 61.93 & 61.67 & 61.08 \\

        \midrule

        \multirow{3}*{\rotatebox{90}{\tiny \textbf{Pubmed}}}

        & \textbf{\model{}} 
        & 35.77
        & 32.91 & 32.36 & 32.02 & 32.35 & 32.32 & 31.66 & 30.74 & 24.29
        & 72.38
        & 70.62 & 70.04 & 70.22 & 70.30 & 70.03 & 69.87 & 69.24 & 63.34 \\        

        & \textbf{w/o SECC} 
        & 35.59
        & 32.00 & 32.48 & 32.71 & 32.47 & 32.25 & 31.18 & 31.61 & 31.59
        & 72.31
        & 69.96 & 70.66 & 70.57 & 70.27 & 70.37 & 69.61 & 69.70 & 69.15 \\

        & MAGI 
        & 29.37
        & 28.98 & 28.97 & 29.14 & 28.69 & 26.17 & 23.47 & 20.00 & 4.36
        & 66.23
        & 65.94 & 66.14 & 65.90 & 65.59 & 62.14 & 57.53 & 51.61 & 43.96 \\

        \midrule

        \multirow{3}*{\rotatebox{90}{\tiny \textbf{arxiv}}}

        & \textbf{\model{}} 
        & 48.42
        & 48.11 & 48.09 & 46.12 & 45.52 & 45.15 & 44.24 & 43.34 & 42.16
        & 44.13
        & 43.43 & 43.33 & 38.21 & 34.70 & 34.99 & 33.69 & 33.54 & 34.14 \\
        
        & \textbf{w/o SECC} 
        & 48.15
        & 47.87 & 47.55 & 47.42 & 46.28 & 45.99 & 45.13 & 43.30 & 41.46
        & 44.09
        & 43.82 & 43.19 & 43.12 & 41.15 & 39.67 & 37.51 & 34.46 & 30.57 \\
        
        & MAGI 
        & 46.90
        & 45.86 & 46.35 & 46.17 & 45.66 & 44.93 & 43.80 & 42.72 & 40.69
        & 38.80
        & 37.62 & 37.09 & 37.01 & 35.59 & 34.22 & 33.00 & 31.90 & 29.74 \\

        \midrule

        \multirow{3}*{\rotatebox{90}{\tiny \textbf{Reddit}}}

        & \textbf{\model{}} & 87.92
        & 87.62 & 87.47 & 87.51 & 87.45 & 87.41 & 87.29 & 86.98 & 86.82
        & 91.75
        & 91.41 & 91.29 & 91.46 & 91.20 & 91.00 & 90.77 & 90.86 & 90.78 \\
        
        & \textbf{w/o SECC} & 87.45
        & 85.33 & 84.94 & 84.64 & 85.58 & 85.40 & 84.43 & 84.54 & 84.35
        & 91.14
        & 83.38 & 82.70 & 81.14 & 84.64 & 82.20 & 82.35 & 82.14 & 82.35 \\
        
        & MAGI & 87.50
        & 87.84 & 87.34 & 87.29 & 87.09 & 87.44 & 86.81 & 87.15 & 87.12
        & 91.10
        & 91.79 & 90.40 & 89.44 & 89.97 & 90.31 & 88.80 & 90.14 & 90.45 \\
        
        \midrule

        \multirow{3}*{\rotatebox{90}{\tiny \textbf{products}}}

        & \textbf{\model{}} 
        & 55.61
        & 55.88 & 55.75 & 55.48 & 55.33 & 55.95 & 55.69 & 55.11 & 55.43
        & 43.88
        & 43.77 & 42.96 & 44.21 & 42.07 & 42.84 & 42.90 & 42.64 & 41.96 \\

        & \textbf{w/o SECC} 
        & 55.33
        & 56.49 & 55.64 & 55.59 & 55.51 & 55.76 & 55.12 & 55.52 & 55.20
        & 42.89
        & 44.05 & 42.86 & 43.86 & 42.42 & 42.57 & 42.26 & 42.21 & 42.86 \\
        
        & MAGI 
        & 55.10
        & 55.66 & 55.52 & 56.05 & 54.64 & 55.12 & 54.68 & 54.94 & 54.43 
        & 42.50
        & 41.08 & 40.59 & 41.62 & 39.63 & 40.54 & 41.54 & 40.59 & 38.65 \\

        \bottomrule
    \end{tabular}
    \vspace{-2mm}
\end{table*}
\begin{table}[t]
    \centering
    \caption{\revised{Statistics of the batch before and after CSampE.}}
    \label{tab: batch}
    \vspace{-4mm}
    \small
    \renewcommand\arraystretch{1.0}
    \setlength{\tabcolsep}{1.6mm}
    \begin{tabular}{l|lll}
        \toprule
        \textbf{Dataset} & \textbf{AVG. Node}  & \textbf{AVG. Edge} & \textbf{AVG. Degree} \\
        \hline

        Photo           & 2,048 $\to$ 6,609   & 19,475 $\to$ 226,000    & 9.51 $\to$ 34.20 \\
        Computers       & 2,048 $\to$ 9,713   & 12,868 $\to$ 374,281    & 6.28 $\to$ 8.53 \\
        Pubmed          & 2,048 $\to$ 14,863  & 3,002 $\to$ 72,280      & 1.47 $\to$ 4.86 \\
        Ogbn-arxiv      & 2,048 $\to$ 28,549  & 2,384 $\to$ 204,633     & 1.16 $\to$ 7.17 \\
        Reddit          & 2,048 $\to$ 12,127  & 10,906 $\to$ 1,442,950  & 5.33 $\to$ 118.99 \\
        Ogbn-products   & 2,048 $\to$ 22,788  & 2,135 $\to$ 284,661     & 1.04 $\to$ 12.49  \\
                
        \bottomrule
    \end{tabular}
    \vspace{-2mm}
\end{table}

\revised{To further evaluate whether \model{} can alleviate structural isolation under sparse graph conditions, we progressively remove 10\%--80\% of the edges from the original graph and report the results in Table~\ref{tab: del_rate}. 
In addition, Table~\ref{tab: batch} summarizes the statistics of mini-batches before and after CSampE expansion.
The results reveal two important observations. 
\textbf{First, CSampE substantially enriches the structural context available to each mini-batch.} 
Across all datasets, the expanded batches contain significantly more nodes and edges than the original sampled batches. 
For example, on Ogbn-arxiv and Ogbn-products, the average batch size increases from 2,048 nodes to 28,549 and 22,788 nodes, respectively, while the average degree increases by more than 6$\times$ and 12$\times$. 
This demonstrates that CSampE effectively reconnects fragmented local neighborhoods and recovers structural information that would otherwise be lost during mini-batch sampling.
\textbf{Second, \model{} exhibits strong robustness under severe graph sparsification.} Even after removing up to 80\% of the edges, the performance degradation remains relatively moderate on most datasets. 
Compared with MAGI, \model{} consistently maintains higher NMI and ACC scores, with the advantage becoming particularly evident on sparse graphs such as Pubmed, where MAGI suffers a dramatic collapse while \model{} preserves substantially better clustering quality. 
These results indicate that the community-aware structural expansion introduced by \model{} effectively mitigates the loss of connectivity caused by both mini-batch sampling and graph sparsification, thereby improving resilience against structural fragmentation.
Interestingly, \model{} also remains consistently more stable than its w/o SECC variant on several sparse graphs (e.g., Reddit and Ogbn-arxiv), suggesting that the structural prior provided by SECC further improves resilience against graph fragmentation.}

\vspace{-4mm}
\subsection{Case Study: Balancing SE Minimization and Community Cohesion}
\label{sec: case_study}
\begin{figure*}[t]
	\centering
	\subfigure[Heatmap of Structural Entropy on Ogbn-arxiv.]{
		\includegraphics[width=0.49\linewidth]{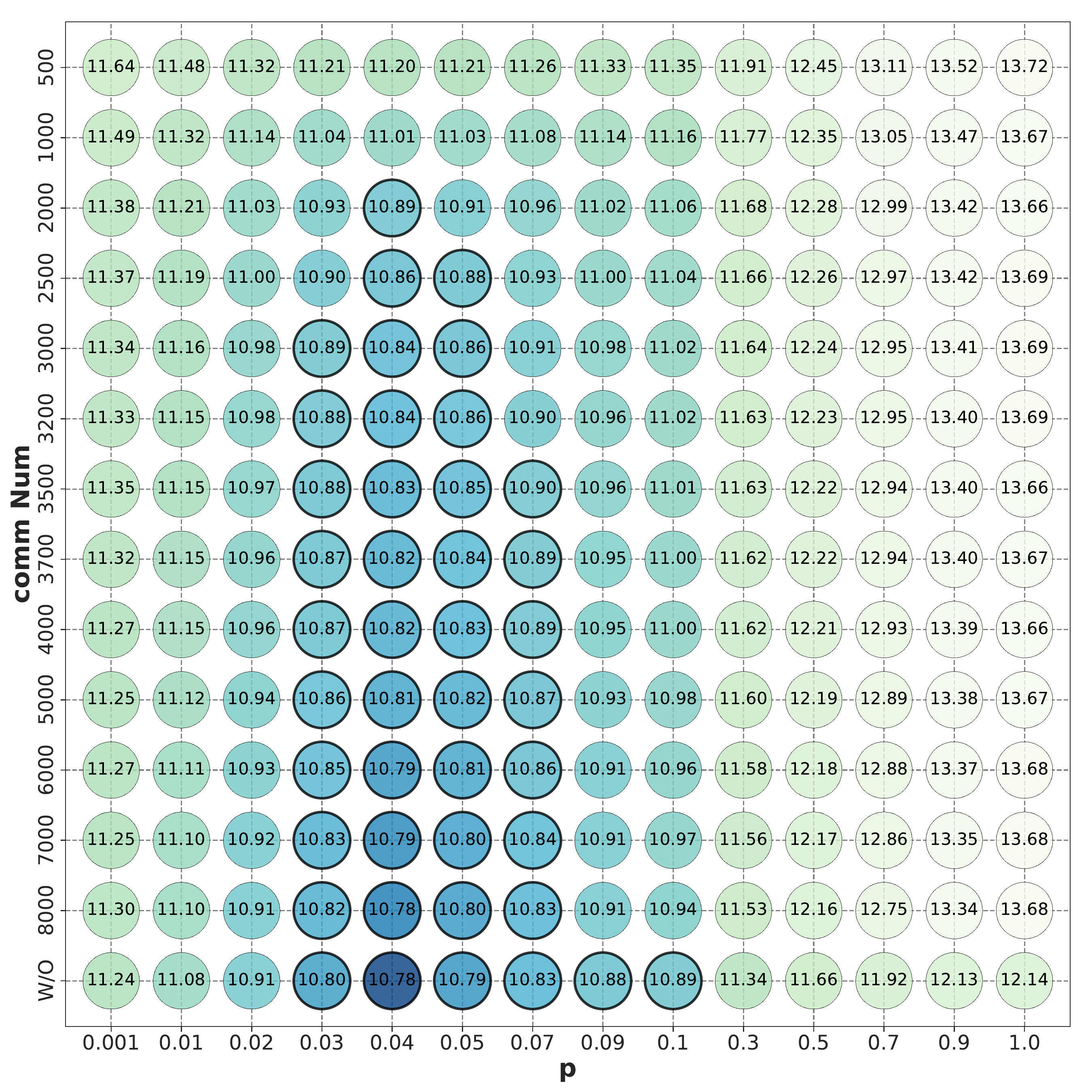}
		\label{fig: SE_hot_arxiv}
	}\hfill
	\subfigure[Histogram of community size on Ogbn-arxiv.]{
		\includegraphics[width=0.49\linewidth]{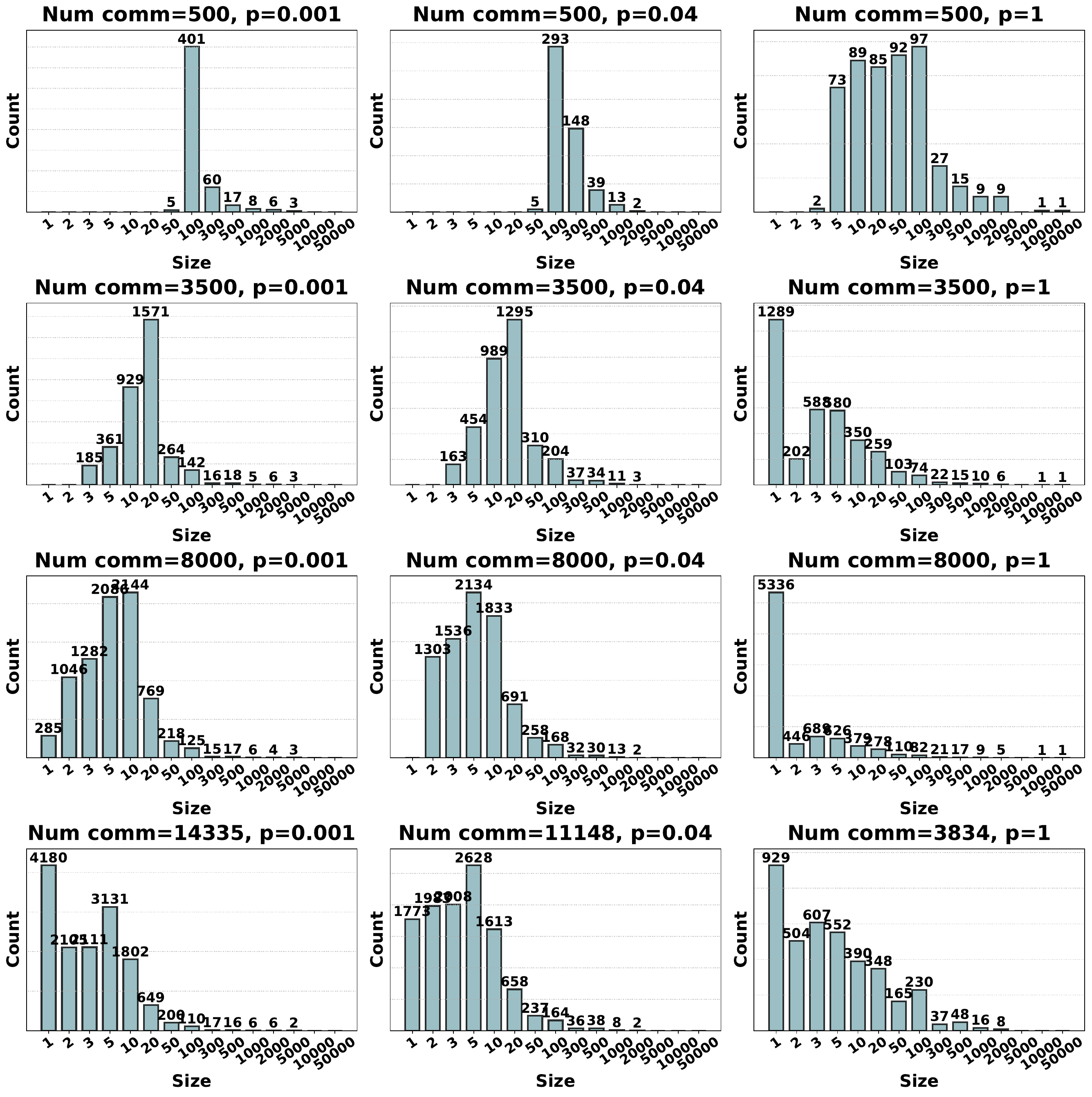}
		\label{fig: SE_bin_arxiv}
	}
    \vspace{-6mm}
	\caption{Evolution of structural entropy and community distribution on Ogbn-arxiv.}
	\label{fig: case_arxiv}
    \vspace{-5mm}
\end{figure*}

To intuitively understand how SECC minimizes structural entropy while enforcing global constraints, \revised{we conduct a comprehensive case study on the Ogbn-arxiv dataset.} 
\revised{Figure~\ref{fig: case_arxiv} illustrates the structural entropy landscapes (Figure~\ref{fig: SE_hot_arxiv}) alongside the corresponding community size distributions (Figure~\ref{fig: SE_bin_arxiv}) under varying $N_{comm}$ and merge speeds $p$.}
Specifically, for Ogbn-arxiv, $N_{comm}$ ranges from 500 to 8,000, including a "$w/o$" setting representing the unconstrained scenario. 
As shown in Figure~\ref{fig: SE_bin_arxiv}, without SECC constraints, the number of formed communities can skyrocket to 14,335, 11,148, or 3,834 for different $p$ values. 
Notably, the minimum constraint of 136 is dictated by the 115 inherent isolated nodes in the dataset and the connectivity between communities, which define the lower bound of partitionable structures.\par

The heatmaps visualize the structural entropy values, where deeper shades represent lower entropy, indicating a more stable and cohesive partition. 
A key observation is that structural entropy remains remarkably low and stable across a wide range of $N_{comm}$ when modulated by $p$. 
Specifically, on the Ogbn-arxiv dataset, when $p$ is tuned between $0.03$ and $0.07$, $N_{comm} > 3,000$ consistently maintains the structural entropy below $10.9$.
This quantitative evidence proves that by executing the SECC module, \model{} sacrifices a negligible amount of structural entropy precision to secure substantial community-level cohesion. 
By constraining the solution space, SECC guides the optimization process toward partitions that are both structurally stable and semantically meaningful.
The grid of histograms further reveals how community boundaries are formed under different configurations.
\textbf{(1) Over-fragmentation without Constraints}: As depicted in the bottom rows of Figure~\ref{fig: case_arxiv}(b), unconstrained optimization continuously favors local entropy reduction, producing a large number of tiny communities (size $<5$). In this case, community boundaries are excessively fragmented and fail to reflect meaningful mesoscopic structures.
\textbf{(2) Community Collapse under Excessive Constraint}: Conversely, overly strict constraints (top rows) force many structurally distinct regions to be merged together. The resulting boundaries become overly coarse, yielding a few dominant communities that obscure the intrinsic organization of the graph.
\revised{\textbf{(3) Balanced Boundary Formation}: In contrast, the partitions generated by \model{} (middle rows) exhibit a much healthier community size distribution, with few fragmented micro-communities and few oversized giant clusters. This indicates that the resulting boundaries emerge near structural bottlenecks where further merging provides little entropy reduction, while preserving sufficient intra-community connectivity. Consequently, SECC achieves a balance between entropy minimization and community cohesion, producing partitions that better align with the underlying graph structure.}\par

\begin{figure}[t]
	\centering
	\subfigure[\model{}.]{
		\includegraphics[width=0.31\linewidth]{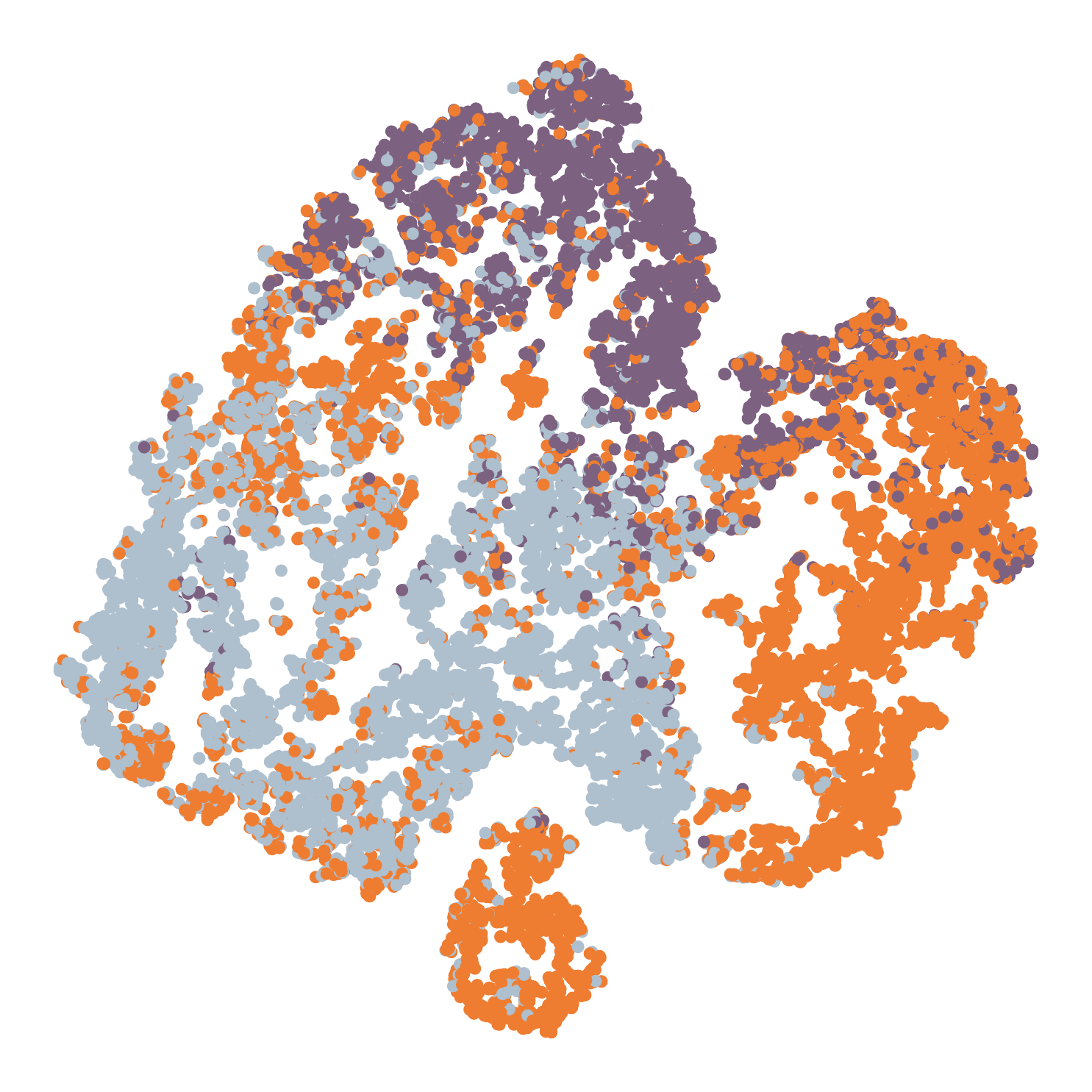}
		\label{fig: tsne_pubmed_SCISE}
	}\hfill
	\subfigure[MAGI.]{
		\includegraphics[width=0.31\linewidth]{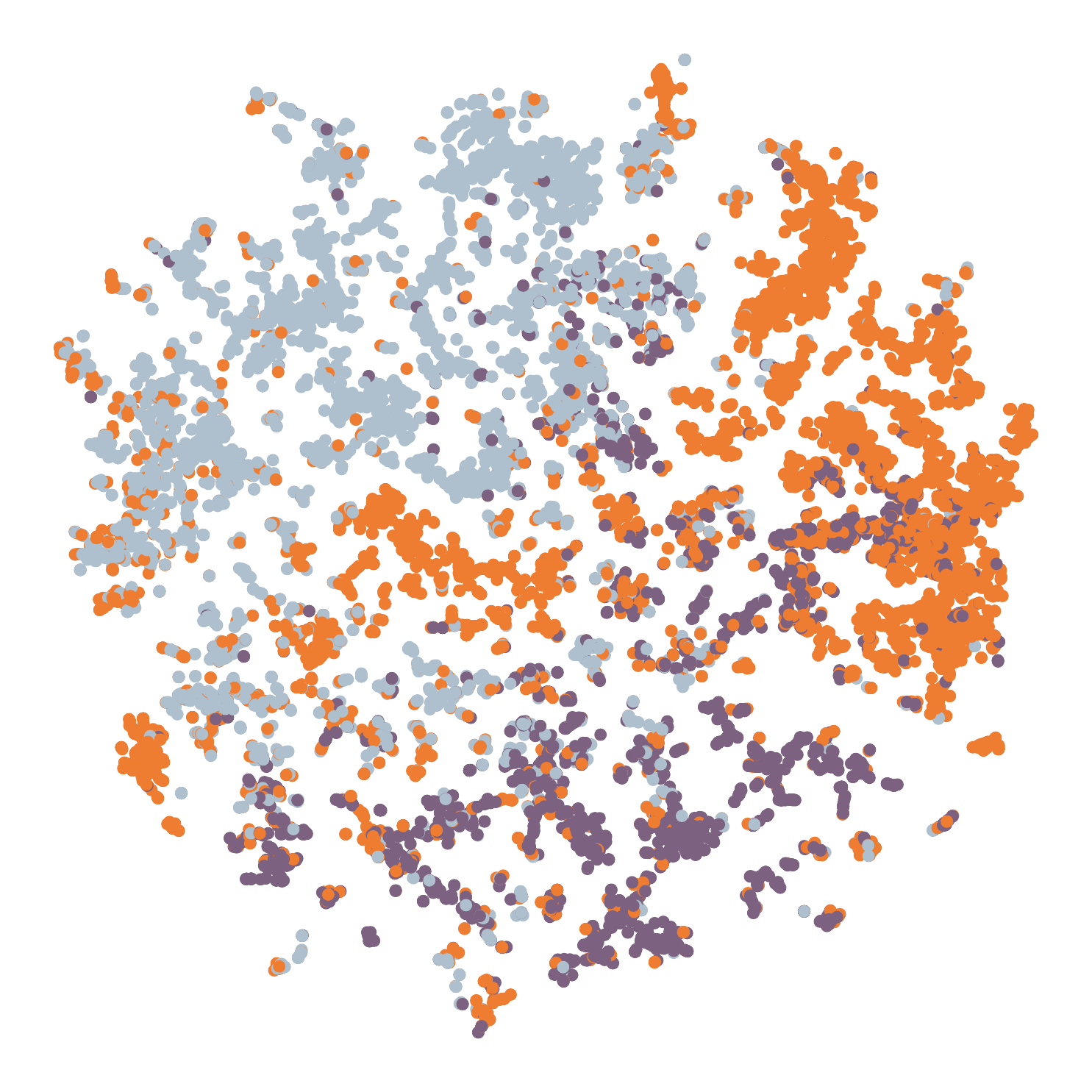}
		\label{fig: tsne_pubmed_MAGI}
	}\hfill
	\subfigure[LSEnet.]{
		\includegraphics[width=0.31\linewidth]{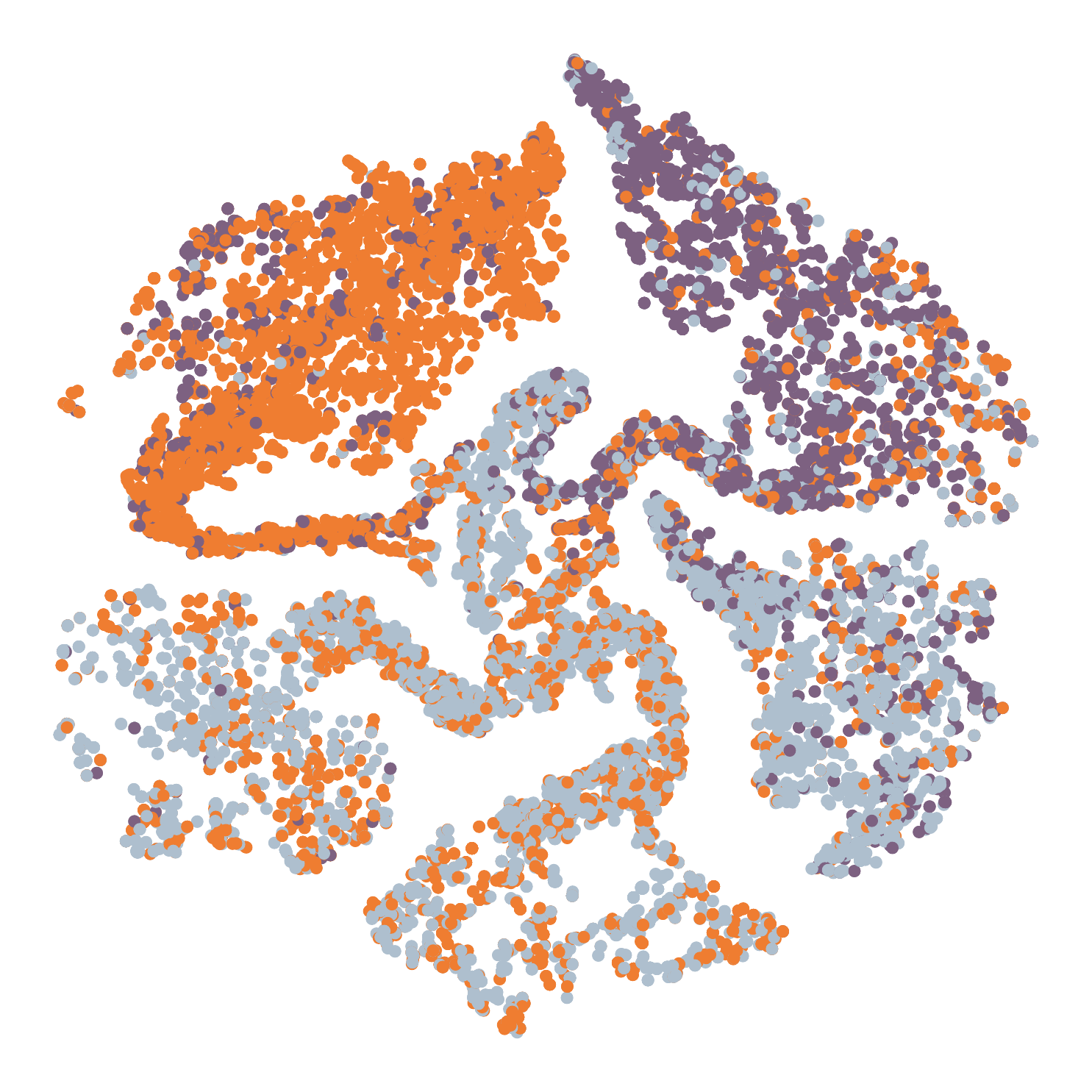}
		\label{fig: tsne_pubmed_LSEnet}
	}
    
    \vspace{-4mm}

    \subfigure[SCGC.]{
		\includegraphics[width=0.31\linewidth]{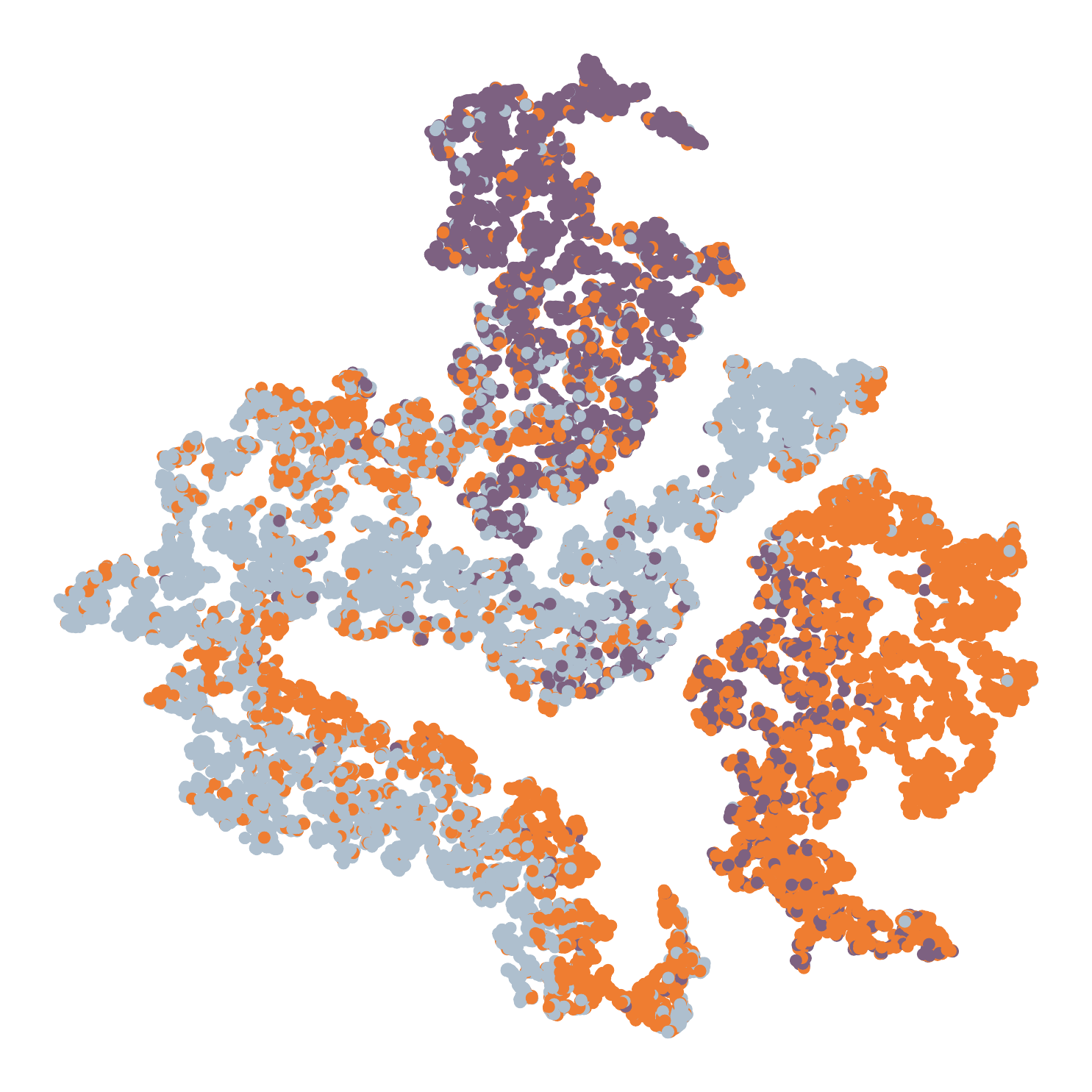}
		\label{fig: tsne_pubmed_SCGC}
	}\hfill
	\subfigure[DeSE.]{
		\includegraphics[width=0.31\linewidth]{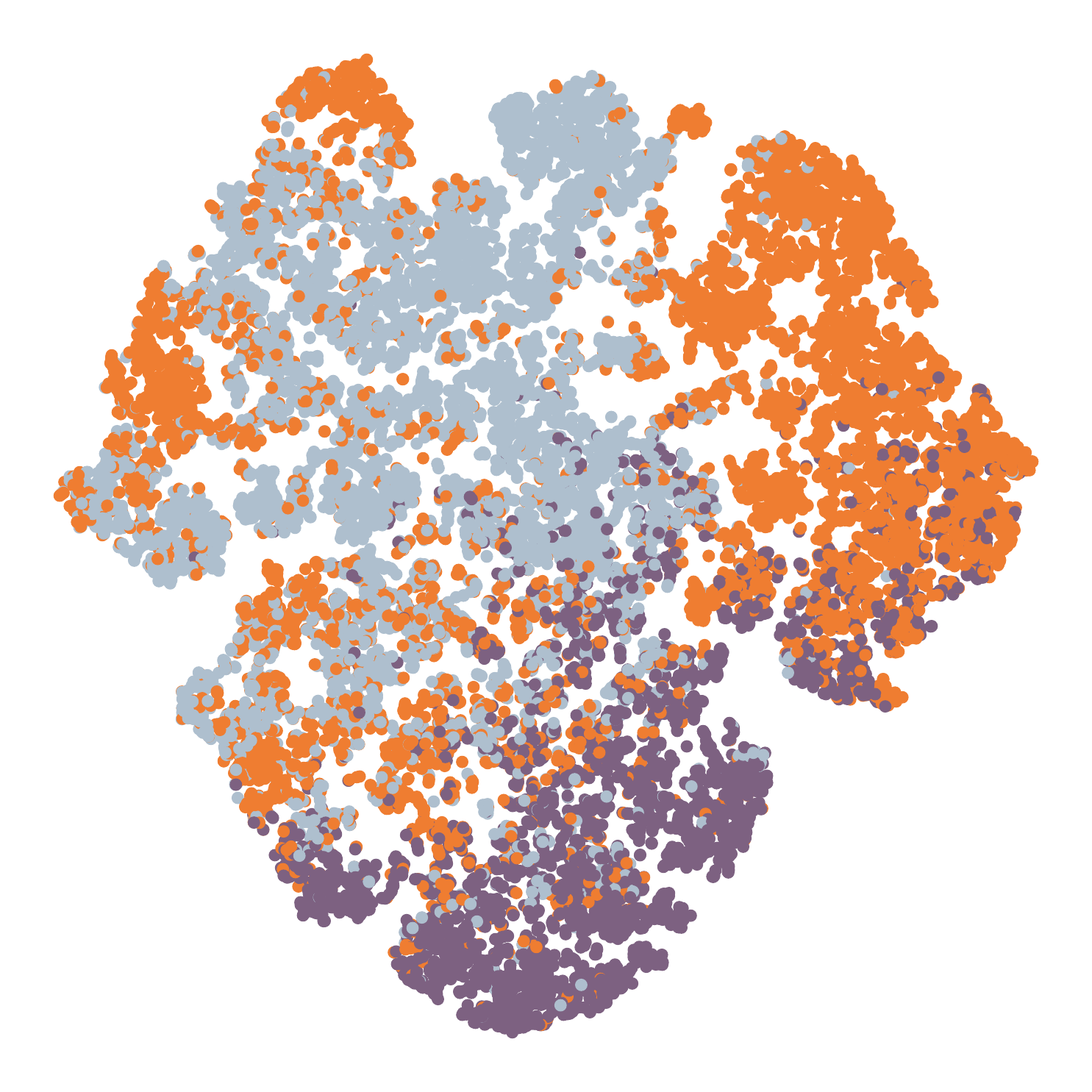}
		\label{fig: tsne_pubmed_DeSE}
	}\hfill
	\subfigure[DinkNet.]{
		\includegraphics[width=0.31\linewidth]{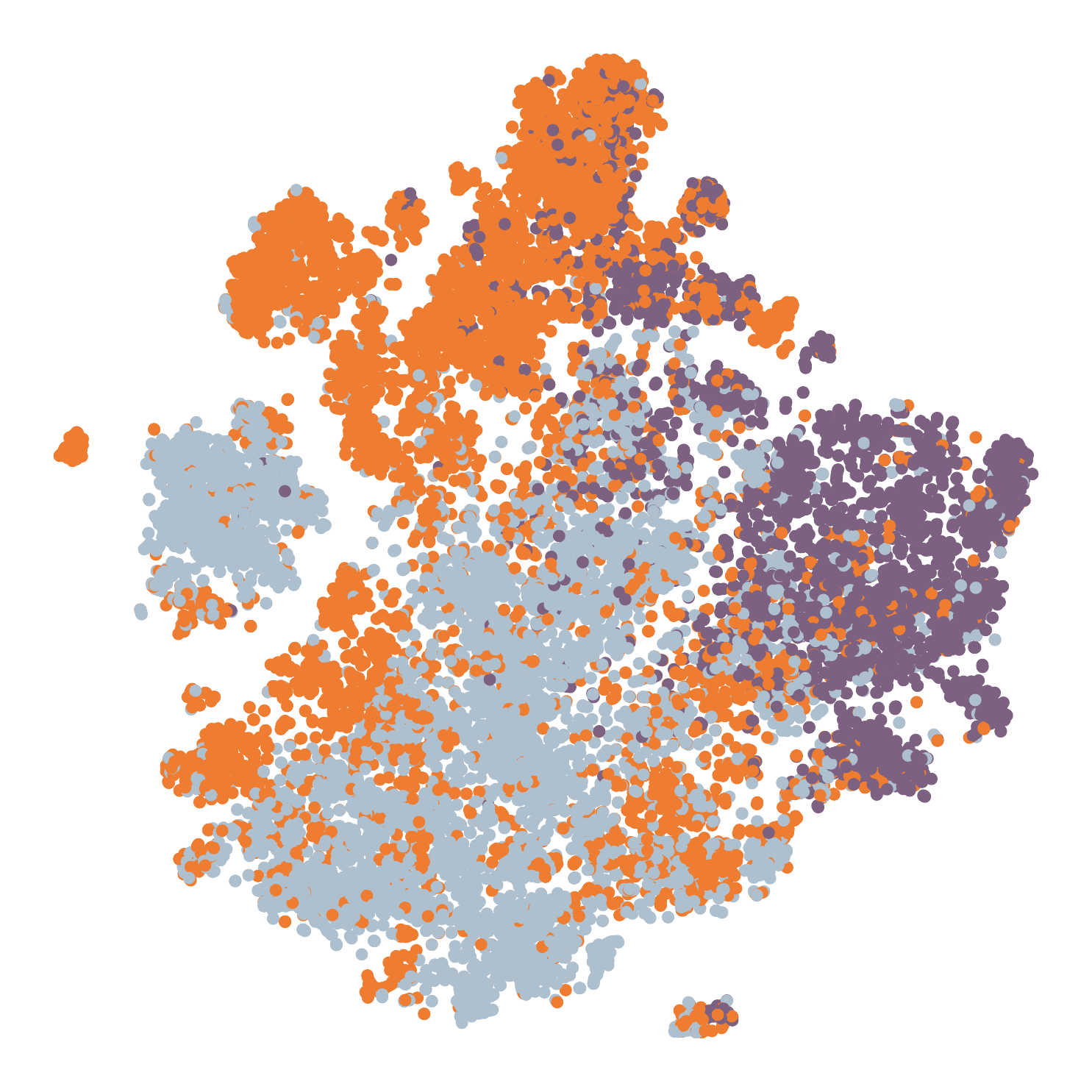}
		\label{fig: tsne_pubmed_DinkNet}
	}
    \vspace{-5mm}
	\caption{Visualization on Pubmed.}
	\label{fig: visualization_pubmed}
    \vspace{-3mm}
\end{figure}

\vspace{-5mm}
\subsection{Visualization}
\label{sec: visualization}

To qualitatively evaluate the learned representations, we apply t-SNE to project node embeddings into a two-dimensional space.
Figure~\ref{fig: visualization_pubmed} presents the visualization results on PubMed. Compared with existing methods, the embeddings generated by \model{} exhibit more compact intra-cluster distributions and clearer inter-cluster boundaries. 
While several baselines produce elongated or overlapping manifolds, \model{} forms well-separated clusters with substantially reduced boundary ambiguity. 
This observation suggests that the proposed community-aware structural expansion and contrastive learning strategy effectively capture global structural semantics, resulting in more discriminative representations for graph clustering.

\vspace{-3.2mm}
\section{Conclusion}
\label{sec: conclusion}


In this paper, we propose \model{}, a scalable unsupervised graph clustering framework designed to alleviate structural isolation in large-scale graph learning. 
By incorporating the Structural Entropy Community Constraint (SECC), Community-Aware Sampling Expansion (CSampE), and Structured Contrastive Learning (StructCL), \model{} effectively preserves both community cohesion and global topological information during mini-batch training.
Extensive experiments on six benchmark datasets demonstrate that \model{} consistently achieves state-of-the-art clustering performance while maintaining strong scalability on graphs containing millions of nodes. 
\revised{Beyond accuracy improvements, our analyses show that \model{} is highly robust to noisy community priors, graph sparsification, and hyperparameter variations. 
Furthermore, the additional time and memory evaluations confirm that the proposed SECC preprocessing introduces only a negligible overhead relative to the overall training pipeline, making \model{} practical for real-world large-scale deployments.
Overall, our findings highlight the importance of integrating community-level structural information into scalable graph clustering and demonstrate that reliable structural priors can effectively mitigate sampling-induced information loss without sacrificing efficiency or robustness. }
As future work, we plan to investigate more scalable structural entropy optimization techniques and extend the proposed framework to emerging ultra-large graph benchmarks, such as Ogbn-papers100M.
\revised{Another promising direction is to develop adaptive community-aware sampling strategies that automatically adjust random-walk length and sampling budgets according to graph structural characteristics.}

\vspace{-3mm}
\begin{acks}
This work is supported by the 
NSFC through grants U25B2029, 62322202, and 62432006, 
Beijing Natural Science Foundation through grant L253021, 
the Pioneer and Leading Goose R\&D Program of Zhejiang through grant 2025C02044, 
Local Science and Technology Development Fund of Hebei Province Guided by the Central Government of China through grant 254Z9902G, Science Research Project of Hebei Higher Education Institutions under grant CYZD2026005,  
Shijiazhuang Science and Technology Plan Project through grant 2511301807A, 
Maior Science and Technology Special Projects of Yunnan Province through grants 202502AD080012 and 202502AD080006,
and the Academic Excellence Foundation of BUAA for PhD Students.
\end{acks}

\clearpage
\bibliographystyle{ACM-Reference-Format}
\bibliography{7_references}

\clearpage
\section{Appendix}
\label{sec: appendix}
\appendix

\section{Notations} 
\label{ap: notations}
The comprehensive list of the primary symbols used throughout this paper is presented in Table~\ref{tab: notation}.

\begin{table}[thp]
    \aboverulesep=0ex
    \belowrulesep=0ex
    \caption{Forms and interpretations of notations.} 
    \vspace{-3mm}
    \centering
    \renewcommand\arraystretch{1.2}
    \setlength{\tabcolsep}{1mm}
    \begin{adjustbox}{width=1.0\columnwidth,center}
    \begin{tabular}{r|l} 
        \toprule
        \textbf{Symbol} & \textbf{Definition} \\
        \hline
        $G=(V, E, X)$ & The original graph.\\
        
        $V$, $E$, $X$ & The node set , the edge set, the node feature matrix.\\

        $N$, $M$ & The number of nodes, the number of edges.\\

        $A \in {\{0, 1\}}^{N \times N}$ & The adjacency matrix of the original graph.\\

        $f$, $d$ & The dimension of feature/embedding. \\

        $\mathcal{Y}\in\{1,\dots,K\}^n$ & Clustering assignments for all nodes. \\

        $K$ & Number of clusters. \\

        \hline
        $\mathcal{T}$; $\lambda$ & Encoding tree; The root vertice of the encoding tree. \\
        
        $\lambda$ & Root node of the encoding tree $\mathcal{T}$. \\

        $\alpha$; $\alpha^-$ & Vertice on encoding tree. Parent vertice of vertice $\alpha$. \\

        $T_\alpha$ & Subset of graph nodes corresponding to tree node $\alpha$. \\

        $g_\alpha$ & Total weight of cut edges between $T_{\alpha}$ and $V \setminus T_{\alpha}$. \\

        $d_{v}$ & Degree of node $v$. \\

        $vol(G); vol(\alpha)$ & Volume of Graph $G$; Volume of vertice $\alpha$.\\

        $H^\mathcal{T}(G)$ & Structural entropy of graph $G$ induced by encoding tree $\mathcal{T}$.\\

        $h$ & The height of the encoding tree. \\

        $\mathcal{C}=\{c_1,\dots,c_N\}$ & Set of communities produced by SECC. \\

        $c(v)$ & Community assignment of node $v$. \\

        $N_{\text{comm}}$ & Target number of communities in SECC. \\

        $p$ & Parallel merge ratio controlling agglomeration rate. \\
        
        $\Delta H(c_i,c_j)$ & Change in SE when merging communities $c_i$ and $c_j$. \\

        $OP$ & Number of candidate community pairs that could merge. \\

        \hline
        $\mathcal{V}_{batch}$ & Initial set of sampled nodes in a mini-batch. \\

        $\mathcal{V}_{ext}$ & Community-guided expansion node set in CSampE. \\

        $\mathcal{V}^*_{batch}$ & Final expanded batch after CSampE. \\

        $\theta$ & Community size threshold for batch expansion strategy. \\
        
        $w_l, w_t$ & Length and number of random walks, respectively. \\

        $f(u;v)$ & Random walk visitation frequency from $v$ to $u$. \\

        $\mathcal{R}_v$ & Reachable node set of $v$ via intra-community walks. \\

        $\mathbf{A}^* \in \mathbb{R}^{B \times B}$ & Structural affinity matrix constructed within a mini-batch. \\

        $\mathbf{z}_i$  & Latent embedding of node $v_i$. \\

        $\mathbf{Z}_{batch}$ & embeddings of nodes in a mini-batch. \\

        $\mathbf{Z}_{global}$ & consolidated global node embeddings. \\

        $\mathcal{P}_i$ & Structural positive set of node $v_i$. \\

        $\mathcal{L}_{struct}$ & Structural contrastive learning objective. \\
        
        \bottomrule
    \end{tabular}
    \end{adjustbox}
    \label{tab: notation}
    \vspace{-3mm}
\end{table}

\section{Dataset}
\label{ap: dataset}
Detailed descriptions of 6 datasets are provided below:

$\bullet$
\textbf{Photo}~\cite{shchur2018pitfalls} is a segment of the Amazon co-purchase graph specifically focusing on the photography category. Nodes represent products (e.g., cameras and lenses), and edges indicate that two items are frequently bought together. Each node is associated with a 745-dimensional feature vector derived from product reviews in a bag-of-words format, reflecting the fine-grained attributes of photography equipment.

$\bullet$
\textbf{Computers}~\cite{shchur2018pitfalls} is also a part of the Amazon co-purchase network, where nodes correspond to computer-related products and edges represent frequent co-purchase relationships. Similar to the Photo dataset, the node features are encoded as bag-of-words vectors based on product descriptions and customer reviews, capturing the semantic similarities between electronic components and peripherals.

$\bullet$
\textbf{Pubmed}~\cite{sen2008pubmed} is a citation network consisting of scientific publications related to diabetes from the PubMed database. Nodes represent documents, and edges denote citation links between them. Each node is characterized by a feature vector represented by weighted TF-IDF scores of unique words from the document's abstract.

$\bullet$
\textbf{Ogbn-arxiv}~\cite{hu2020ogbn} is a directed graph representing the citation network between all Computer Science (CS) arXiv papers indexed by MAG. Each node is an arXiv paper, and each directed edge indicates that one paper cites another. The node features are derived by averaging the word embeddings of the paper's title and abstract, and the task is to classify papers into 40 subject areas.

$\bullet$
\textbf{Reddit}~\cite{hamilton2017inductive} is a large-scale social network dataset constructed from Reddit posts made in September 2014. Nodes represent posts, and an edge exists between two nodes if the same user commented on both posts. The node features are calculated based on GloVe word vectors, and the communities are defined by the subreddits to which the posts belong.

$\bullet$
\textbf{Ogbn-products}~\cite{hu2020ogbn} is an undirected and unweighted graph representing an Amazon product co-purchasing network. Nodes represent products sold on Amazon, and edges indicate that the products are purchased together. Node features are generated by performing Principal Component Analysis (PCA) on the product descriptions, making it a benchmark for large-scale graph clustering and classification.

\section{Baselines} 
\label{ap: baseline}
Detailed descriptions of 10 baselines compared to our work are introduced as follows:

$\bullet$
DGI~\cite{velivckovic2018deep} is a self-supervised node representation learning approach that maximizes the mutual information between local patch representations and high-level graph summaries to capture global structural context.

$\bullet$
S3GC~\cite{devvrit2022s3gc} is a scalable self-supervised graph clustering framework that employs contrastive learning to integrate node features with topological structures, effectively learning cluster-friendly representations.

$\bullet$
SUBLIME~\cite{liu2022sublime} is a structure bootstrapping contrastive Learning framework with the aid of self-supervised contrastive learning, where the learned graph topology is optimized by data itself.

$\bullet$
CONVERT~\cite{yang2023convert} is a contrastive graph clustering network with reliable augmentation and distills reliable semantic information by recovering the perturbed latent embeddings.

$\bullet$
DMoN~\cite{tsitsulin2023graph} introduces a modularity measure of clustering quality to optimize cluster assignment in an end-to-end manner and proposes Deep Modularity Networks.

$\bullet$
Dink-Net~\cite{liu2023dink} is an end-to-end scalable deep graph clustering method that optimizes clustering distributions through adversarial dilation and shrink losses to guide the network in learning discriminative features.

$\bullet$
LSEnet~\cite{sun2024lsenet} is a hyperbolic graph clustering network that formulates a differentiable structural information (DSI) objective, enabling the model to reveal hierarchical cluster structures and determine cluster numbers in a self-supervised manner.

\begin{figure}[thp]
	\centering
    \subfigure[Pubmed.]{
		\includegraphics[width=0.48\linewidth]{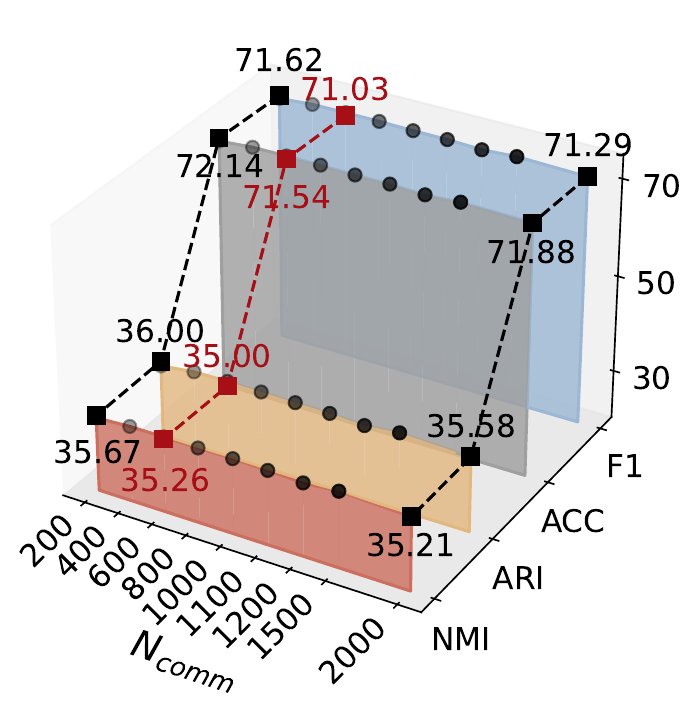}
		\label{fig: hyper_commNum_pubmed}
	}\hfill
    \subfigure[Ogbn-products.]{
		\includegraphics[width=0.48\linewidth]{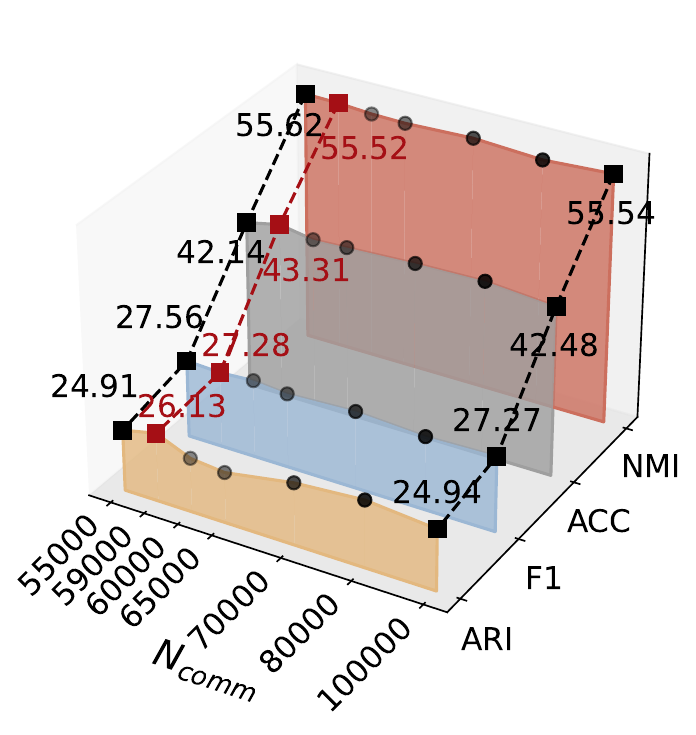}
		\label{fig: hyper_commNum_products}
	}
    \vspace{-2mm}
	\caption{Analysis of hyper $N_{comm}$ on Pubmed and Ogbn-products datasets.}
	\label{fig: hyper_commNum_appendix}
\end{figure}

\begin{figure}[thp]
	\centering
    \subfigure[Pubmed.]{
		\includegraphics[width=0.48\linewidth]{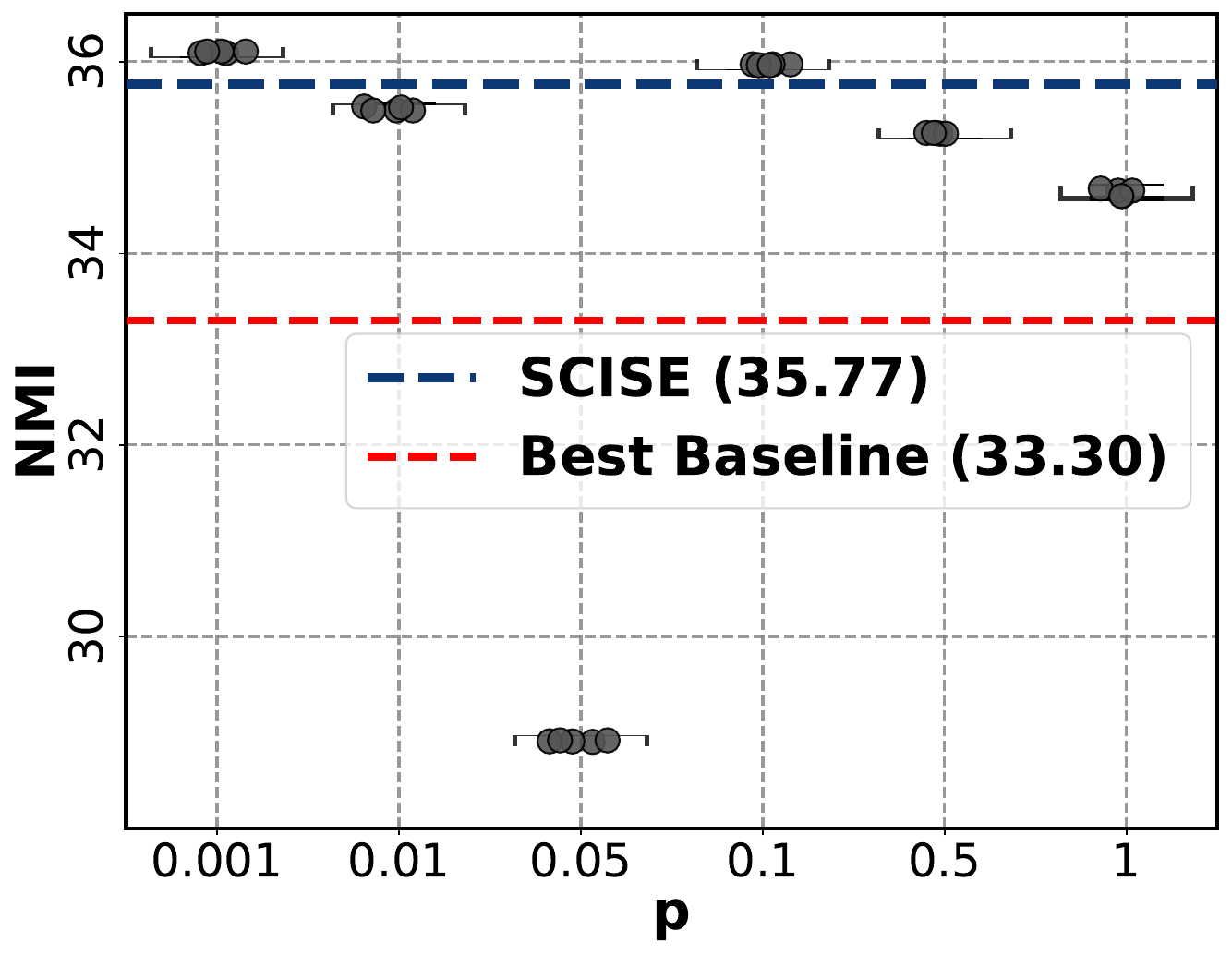}
		\label{fig: hyper_p_pubmed}
	}\hfill
    \subfigure[Ogbn-products.]{
		\includegraphics[width=0.48\linewidth]{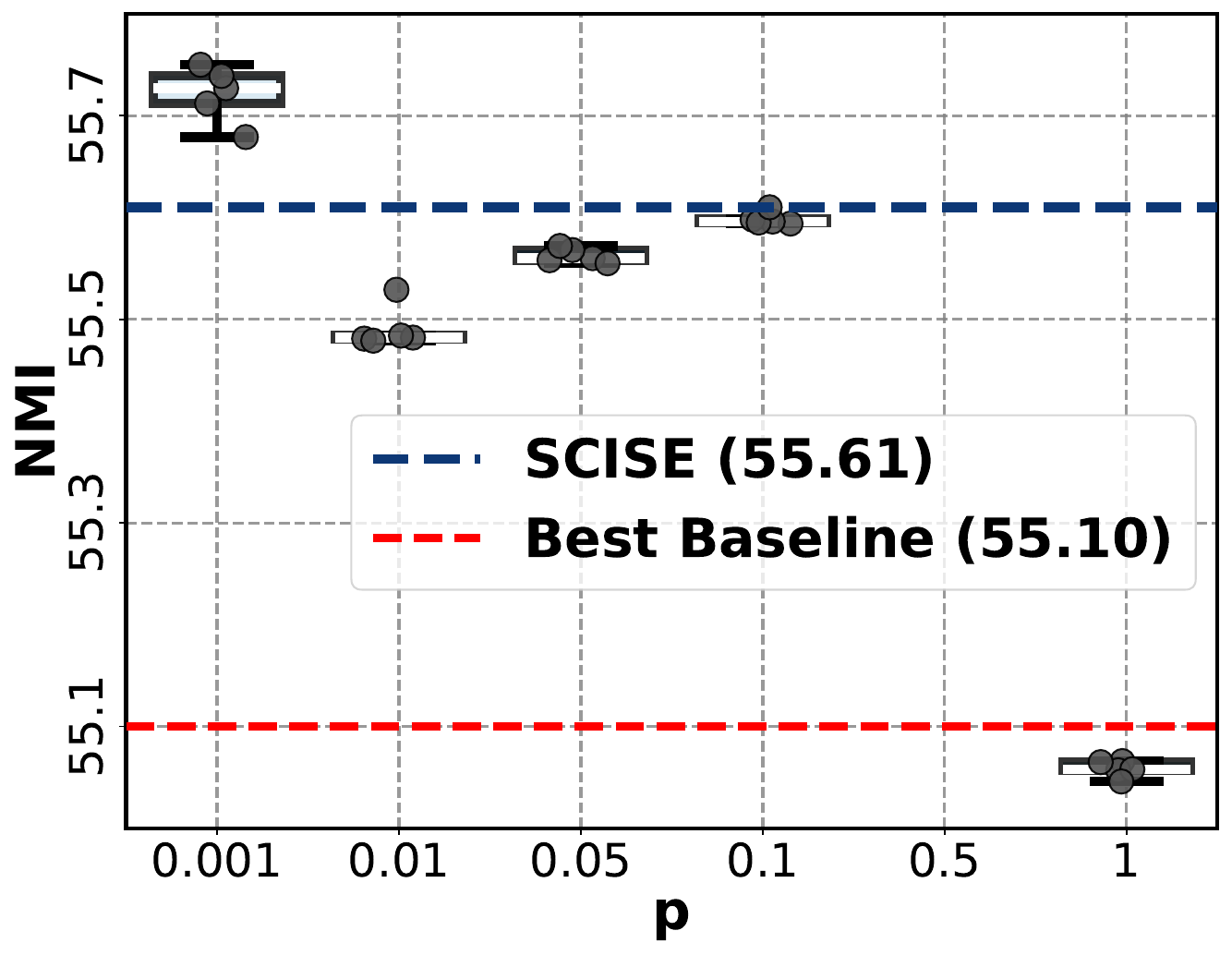}
		\label{fig: hyper_p_products}
	}
    \vspace{-2mm}
	\caption{Analysis of hyper $p$ on Pubmed and Ogbn-products datasets.}
	\label{fig: hyper_p_appendix}
\end{figure}

\begin{figure}[thp]
	\centering
	\includegraphics[width=0.99\linewidth]{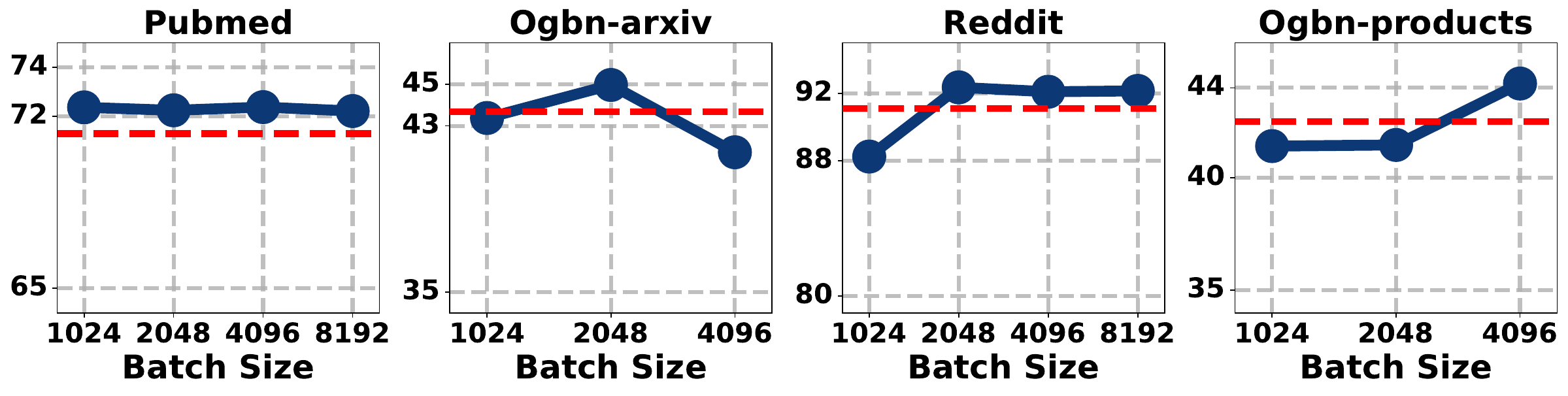}
    \vspace{-2mm}
	\caption{Analysis of hyper $B$ with ACC.}
	\label{fig: hyper_batchsize_acc}
\end{figure}

\begin{figure}[thp]
	\centering
	\includegraphics[width=0.99\linewidth]{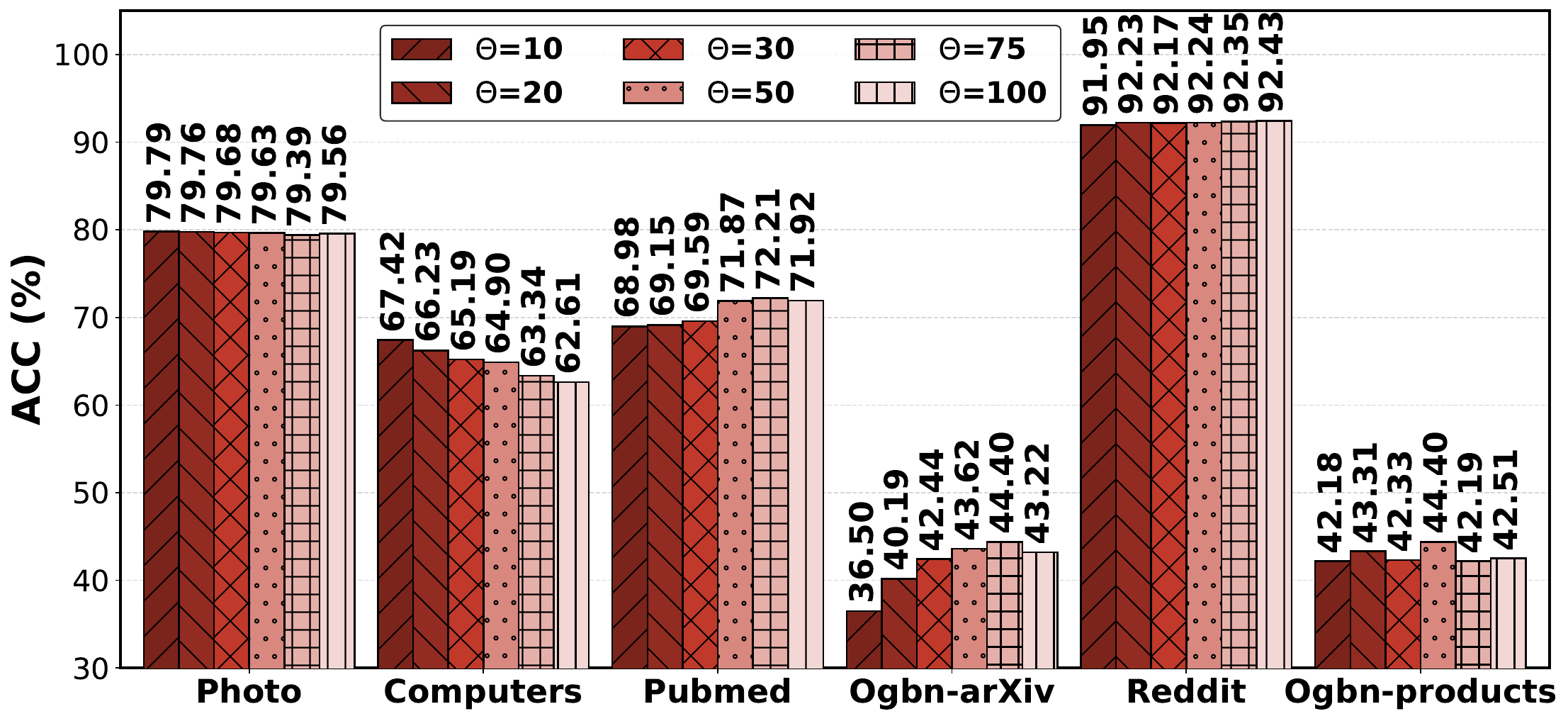}
    \vspace{-2mm}
	\caption{Analysis of hyper $\theta$ with ACC.}
	\label{fig: hyper_theta_acc}
\end{figure}

$\bullet$
MAGI~\cite{liu2024magi} is a graph clustering framework that utilizes modularity maximization as a contrastive pretext task to define positive and negative samples, capturing structural information without manual data augmentations.

\begin{table}[thp]
    \centering
    \caption{Hyperparameter settings.}
    \label{tab: hyperparameter}
    \vspace{-3mm}
    \renewcommand\arraystretch{1.0}
    \setlength{\tabcolsep}{1.2mm}
    \begin{tabular}{l|ccccccc}
        \toprule
        \textbf{Dataset} & \textbf{$N_{comm}$} & \textbf{$p$} & \textbf{$\theta$} & \textbf{batchsize}  & \textbf{lr} & \textbf{$w_t$} & \textbf{$w_l$}\\
        \hline
        
        Photo & 300 & 0.1 & 25 & 2048 & 0.005 & 120 & 2\\
        
        Computers & 450 & 0.1 & 10 & 1024 & 0.0005 & 30 & 10\\
        
        Pubmed & 600 & 0.1 & 70 & 4096 & 0.005 & 30 & 2\\

        Ogbn-arxiv & 3500 & 0.04 & 80 & 2048 & 0.01 & 30 & 3\\
        
        Reddit & 1000 & 1 & 70 & 2048 & 0.01 & 30 & 4\\
        
        Ogbn-products & 59000 & 0.1 & 20 & 4096 & 0.005 & 30 & 5\\
                
        \bottomrule
    \end{tabular}
\end{table}

$\bullet$
DeSE~\cite{zhang2025dese} is an unsupervised graph clustering framework that incorporates differentiable structural entropy with soft assignment, utilizing a structural learning layer and a GNN-based assignment module to optimize clustering on an enhanced graph.

$\bullet$
SCGC~\cite{kulatilleke2025scgc} is a scalable graph clustering framework that eliminates traditional GNN convolutions in favor of simple linear units and dynamic soft structure fusion via an augmentation-less edge-contrastive loss.

\section{Sensitivity Analysis}
\label{ap: sensitivity}

This section provides supplementary results for the hyperparameter sensitivity analysis discussed in Section~\ref{sec: hyperparameter}. \par

$\bullet$
Impact of community number $N_{comm}$: Figure~\ref{fig: hyper_commNum_appendix} illustrates the stability of ~\model{} on the Pubmed and Ogbn-products datasets as the target community number $N_{comm}$ varies. 
The flat curves across wide intervals confirm that the SECC operator effectively anchors the partition regardless of the specific $N_{comm}$ value.

$\bullet$
Impact of merge speed $p$: Figure~\ref{fig: hyper_p_appendix} shows the distribution of performance across different merge speeds $p$ on the Pubmed and Ogbn-products datasets. 
Box plots show that, at different structure entropy minimization granularities, as long as extreme values are not taken, the model performance is always better than the optimal baseline.

$\bullet$
Impact of initial batch size $B$: Figure ~\ref{fig: hyper_batchsize_acc} shows that the model's accuracy is higher than the baseline for most small batch sizes and remains relatively stable, further demonstrating its scalability and decoupling from hardware limitations.

$\bullet$
Impact of community size threshold $\theta$: In Figure~\ref{fig: hyper_theta_acc}, we present the ACC scores across different expansion thresholds $\theta$. 
Similar to the NMI results, the model remains stable for most benchmarks. 
Only on the Ogbn-arxiv dataset does it exhibit a slight trend with increasing $\theta$.\par

\begin{table*}[thp]
    \centering
    \caption{\revised{Comparison between random-walk affinity and PPR-based alternatives.}}
    \label{tab: ppr}
    \vspace{-3mm}
    \renewcommand\arraystretch{1.1}
    \setlength{\tabcolsep}{0.6mm}
    \begin{tabular}{c|cccc|cccc|cccc}
        \toprule
        \multirow{2}*{Variation} 
        & \multicolumn{4}{c|}{\multirow{1}*{\textbf{Photo}}} 
        & \multicolumn{4}{c|}{\multirow{1}*{\textbf{Computers}}} 
        & \multicolumn{4}{c}{\multirow{1}*{\textbf{Pubmed}}}\\
        
        \cline{2-13}

        & {\textbf{NMI}} & {\textbf{ARI}} & {\textbf{ACC}} & \multicolumn{1}{c|}{\textbf{F1}} 
        & {\textbf{NMI}} & {\textbf{ARI}} & {\textbf{ACC}} & \multicolumn{1}{c|}{\textbf{F1}} 
        & {\textbf{NMI}} & {\textbf{ARI}} & {\textbf{ACC}} & \multicolumn{1}{c}{\textbf{F1}} \\

        \hline

        \rowcolor{gray!20}
        \textbf{\model{}} 
        & 74.53$_{\pm 0.14}$ & 64.51$_{\pm 0.29}$ & 79.71$_{\pm 0.15}$ & 76.22$_{\pm 0.12}$
        & 58.46$_{\pm 0.24}$ & 51.78$_{\pm 0.80}$ & 67.32$_{\pm 0.36}$ & 59.48$_{\pm 0.84}$
        & 35.77$_{\pm 0.02}$ & 36.33$_{\pm 0.02}$ & 72.38$_{\pm 0.01}$ & 71.83$_{\pm 0.01}$ \\

        \hline

        \textbf{batch PPR}
        & 51.69$_{\pm 0.71}$ & 37.93$_{\pm 2.12}$ & 62.96$_{\pm 0.98}$ & 65.25$_{\pm 1.13}$ 
        & 50.75$_{\pm 0.33}$ & 35.12$_{\pm 2.87}$ & 52.43$_{\pm 0.62}$ & 46.43$_{\pm 2.99}$ 
        & 25.95$_{\pm 1.28}$ & 24.57$_{\pm 0.88}$ & 63.87$_{\pm 0.85}$ & 64.17$_{\pm 1.47}$ \\

        \textbf{global PPR}
        & 74.26$_{\pm 0.19}$ & 64.10$_{\pm 0.29}$ & 79.44$_{\pm 0.20}$ & 76.15$_{\pm 0.14}$ 
        & 56.81$_{\pm 1.13}$ & 40.29$_{\pm 2.79}$ & 56.34$_{\pm 1.59}$ & 51.37$_{\pm 2.26}$ 
        & 31.52$_{\pm 2.43}$ & 30.57$_{\pm 3.27}$ & 68.56$_{\pm 2.80}$ & 68.16$_{\pm 2.52}$ \\

        \bottomrule

        \multirow{2}*{Variation} 
        & \multicolumn{4}{c|}{\multirow{1}*{\textbf{Ogbn-arxiv}}} 
        & \multicolumn{4}{c|}{\multirow{1}*{\textbf{Reddit}}} 
        & \multicolumn{4}{c}{\multirow{1}*{\textbf{Ogbn-products}}}\\
        
        \cline{2-13}

        & {\textbf{NMI}} & {\textbf{ARI}} & {\textbf{ACC}} & \multicolumn{1}{c|}{\textbf{F1}} 
        & {\textbf{NMI}} & {\textbf{ARI}} & {\textbf{ACC}} & \multicolumn{1}{c|}{\textbf{F1}} 
        & {\textbf{NMI}} & {\textbf{ARI}} & {\textbf{ACC}} & \multicolumn{1}{c}{\textbf{F1}} \\

        \hline

        \rowcolor{gray!20}
        \textbf{\model{}}
        & 48.42$_{\pm 0.08}$ & 37.66$_{\pm 0.47}$ & 44.13$_{\pm 0.59}$ & 31.52$_{\pm 0.41}$
        & 87.92$_{\pm 0.12}$ & 90.97$_{\pm 0.08}$ & 91.75$_{\pm 0.24}$ & 86.48$_{\pm 0.41}$ 
        & 55.61$_{\pm 0.02}$ & 25.65$_{\pm 0.67}$ & 43.88$_{\pm 0.39}$ & 27.82$_{\pm 0.10}$\\

        \hline
                
        \textbf{batch PPR}
        & 30.23$_{\pm 0.72}$ & 10.47$_{\pm 0.53}$ & 20.96$_{\pm 0.22}$ & 17.13$_{\pm 0.11}$ 
        & 54.95$_{\pm 1.01}$ & 35.52$_{\pm 0.62}$ & 51.08$_{\pm 1.12}$ & 44.57$_{\pm 1.20}$ 
        & 27.80$_{\pm 0.98}$ &  7.38$_{\pm 0.54}$ & 20.14$_{\pm 1.07}$ & 13.57$_{\pm 0.89}$ \\

        \textbf{global PPR}
        & 45.67$_{\pm 0.18}$ & 20.33$_{\pm 0.12}$ & 32.77$_{\pm 0.18}$ & 27.73$_{\pm 0.75}$ 
        & \multicolumn{4}{c|}{OOM}
        & \multicolumn{4}{c}{OOM} \\
        
        \bottomrule
    \end{tabular}
    \vspace{-3mm}
\end{table*}
\begin{table}[thp]
    \centering
    \caption{\revised{Comparison of \model{} and MAGI on Directed and Undirected Versions of Ogbn-arxiv.}}
    \label{tab: direct}
    \vspace{-2mm}
    \renewcommand\arraystretch{1.3}
    \setlength{\tabcolsep}{1.8mm}
    \begin{tabular}{l|ccccc}
        \toprule
         & NMI & ARI & ACC & F1 & ALL\\
        \hline

        SCISE-direct & 42.94 & 27.18 & 37.37 & 27.40 & 134.90  \\
        MAGI-direct & 42.24 & 21.97 & 31.07 & 22.35 & 117.63  \\

        \hline
        
        SCISE-undirect & \textbf{48.42} & \textbf{37.66} & \textbf{44.13} & \textbf{31.52} & \textbf{161.73}  \\
        MAGI-undirect & 46.90 & 31.00 & 38.80 & 26.60 & 143.30  \\
                
        \bottomrule
    \end{tabular}
\end{table}

\section{Hyperparameter Configurations}
For the purpose of reproducibility and experimental transparency, we provide the specific hyperparameter settings for \model{} across six datasets. 
These configurations are consistent with the experimental results reported in Section~\ref{sec: cluster}. The detailed parameter assignments are summarized in Table~\ref{tab: hyperparameter}.

\revised{\section{Test on Direct Graph}}

\revised{Although the main paper focuses on undirected homogeneous graphs, \model{} is not inherently restricted to such settings. 
To further evaluate its applicability on directed networks, we conduct an additional experiment on Ogbn-arxiv, which is originally a directed citation graph.
Specifically, we compare \model{} and the strongest baseline MAGI under two settings: (1) the original directed graph and (2) the commonly used undirected graph obtained by symmetrizing the adjacency matrix. The results are reported in Table~\ref{tab: direct}.
Several observations can be made. 
First, \model{} consistently outperforms MAGI under both directed and undirected settings, demonstrating that the proposed framework remains effective even when edge directions are preserved. 
Second, both \model{} and MAGI achieve better performance on the undirected graph. This suggests that bidirectional neighborhood information provides more complete structural signals for community discovery and representation learning. 
Third, despite the performance reduction caused by preserving edge directions, \model{} still maintains a clear advantage over MAGI across all evaluation metrics.
Overall, these results indicate that \model{} is not limited to undirected graphs and can be directly applied to directed networks. 
Nevertheless, the current work mainly targets homogeneous graph clustering, while extending the structural entropy constrained community framework to heterogeneous and more complex graph settings remains an important direction for future research.}

\vspace{-3mm}
\revised{\section{Test on Personalized PageRank}}
\revised{To further validate our technical design regarding the affinity matrix computation, we conducted an ablation study replacing our random walk mechanism with Personalized PageRank (PPR), a well-established metric in the graph clustering literature. We introduced two variants for comparison: (1) \textbf{global PPR}, which computes the exact Personalized PageRank over the entire original graph, and (2) \textbf{batch PPR}, which computes the Personalized PageRank restricted strictly within the sampled mini-batch. 
The quantitative results across six datasets are presented in Table \ref{tab: ppr}. The empirical findings strongly support our choice of a restricted random walk over PPR due to two critical limitations of the latter:
\textbf{The Scalability Bottleneck of Global PPR:} While global PPR maintains relatively competitive performance on smaller datasets (e.g., Photo), it fundamentally lacks scalability. On massive datasets such as Reddit and Ogbn-products, calculating and storing the dense global affinity matrix for millions of nodes immediately triggers Out-of-Memory (OOM) errors. This extreme memory consumption violates the core scalability requirements of our framework, making global PPR completely infeasible for large-scale graph clustering.
\textbf{The Performance Degradation of Batch PPR:} To circumvent the OOM issue, batch PPR limits the computation to the local subgraphs within the batch. However, as demonstrated in Table \ref{tab: ppr}, this localized approach leads to a severe degradation in clustering effectiveness across all datasets. For instance, on the Ogbn-arxiv dataset, the NMI score drops dramatically from 48.42 (SCISE) to 30.23 (batch PPR). Similarly, on the Reddit dataset, the NMI plummets from 87.92 to 54.95. This sharp decline occurs because calculating PPR within a strictly truncated batch suffers from severe boundary effects, isolating nodes from their broader topological environment and losing the crucial global structural context necessary for accurate affinity estimation.
In contrast, our proposed random walk mechanism dynamically captures structural affinity without requiring dense global matrix computations. It effectively circumvents OOM errors while preserving the global structural context, thereby achieving an optimal balance between state-of-the-art clustering effectiveness and strict scalability.}

\begin{figure*}[thp]
	\centering
	\subfigure[Heatmap of Structural Entropy on Photo.]{
		\includegraphics[width=0.49\linewidth]{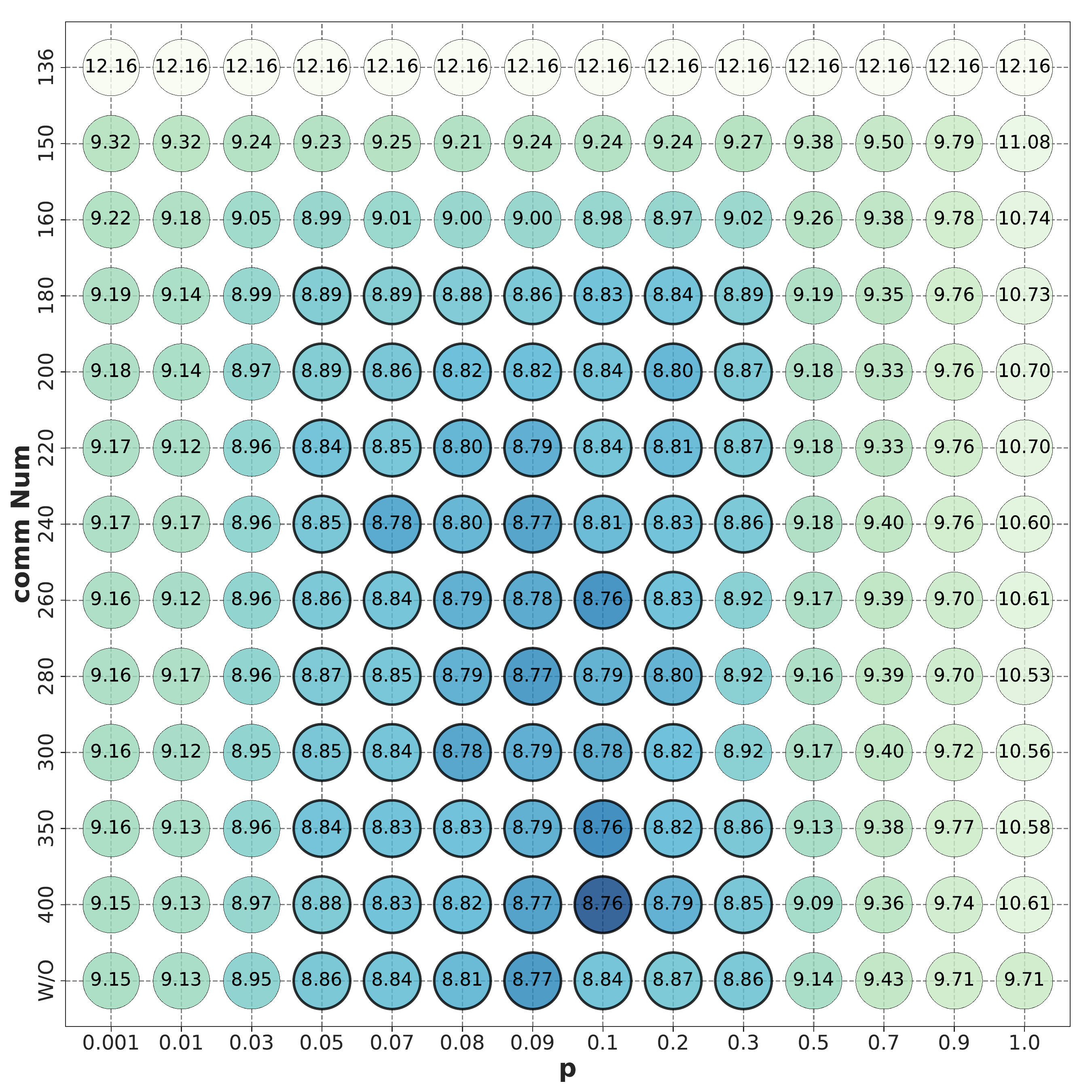}
		\label{fig: SE_hot_photo}
	}\hfill
	\subfigure[Histogram of community size on Photo.]{
		\includegraphics[width=0.49\linewidth]{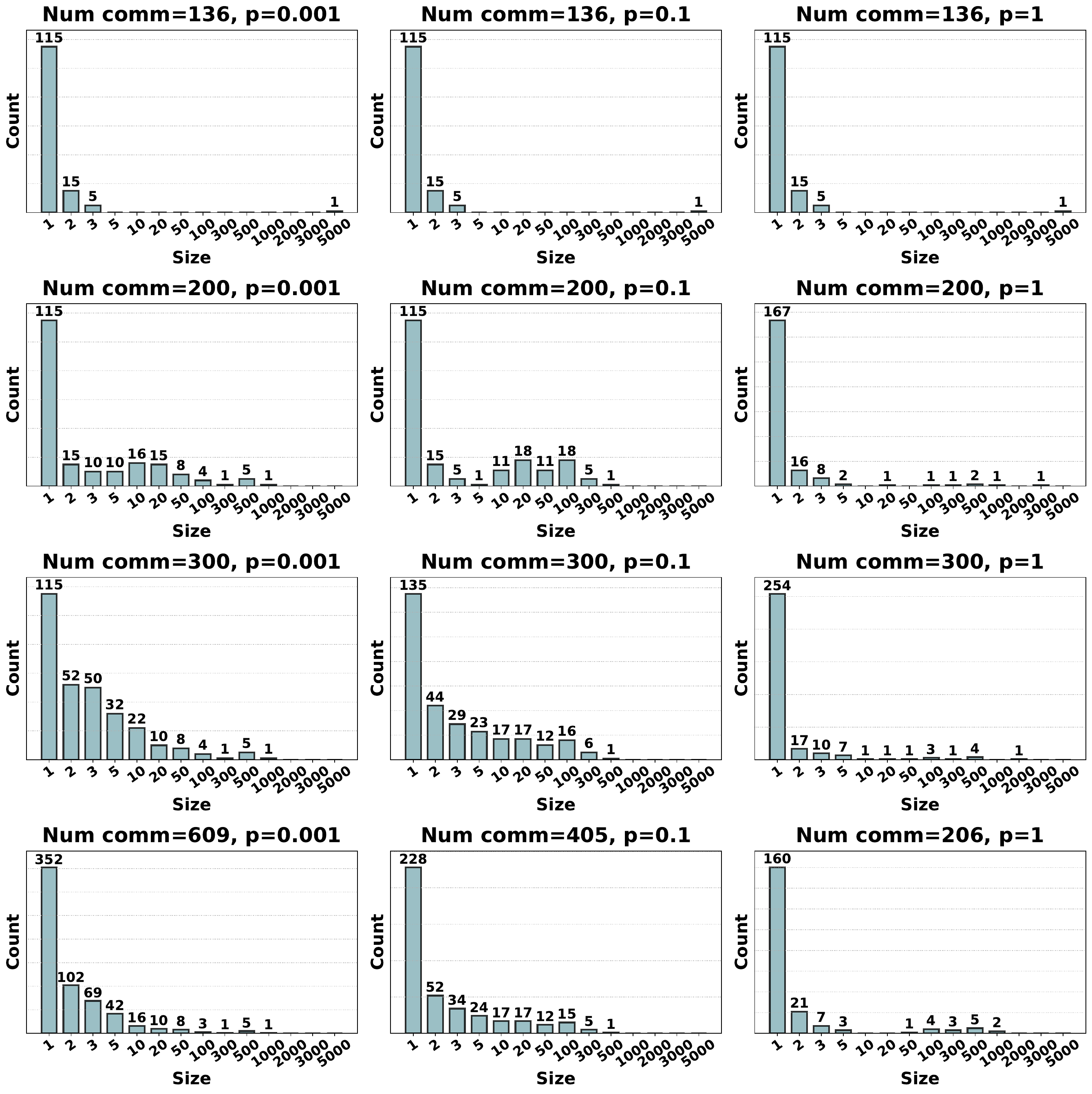}
		\label{fig: SE_bin_photo}
	}
    \vspace{-4mm}
	\caption{Evolution of structural entropy and community distribution on Photo.}
	\label{fig: case_photo}
    \vspace{-4mm}
\end{figure*}

\begin{figure*}[thp]
	\centering
	\subfigure[Heatmap of Structural Entropy on Ogbn-products.]{
		\includegraphics[width=0.49\linewidth]{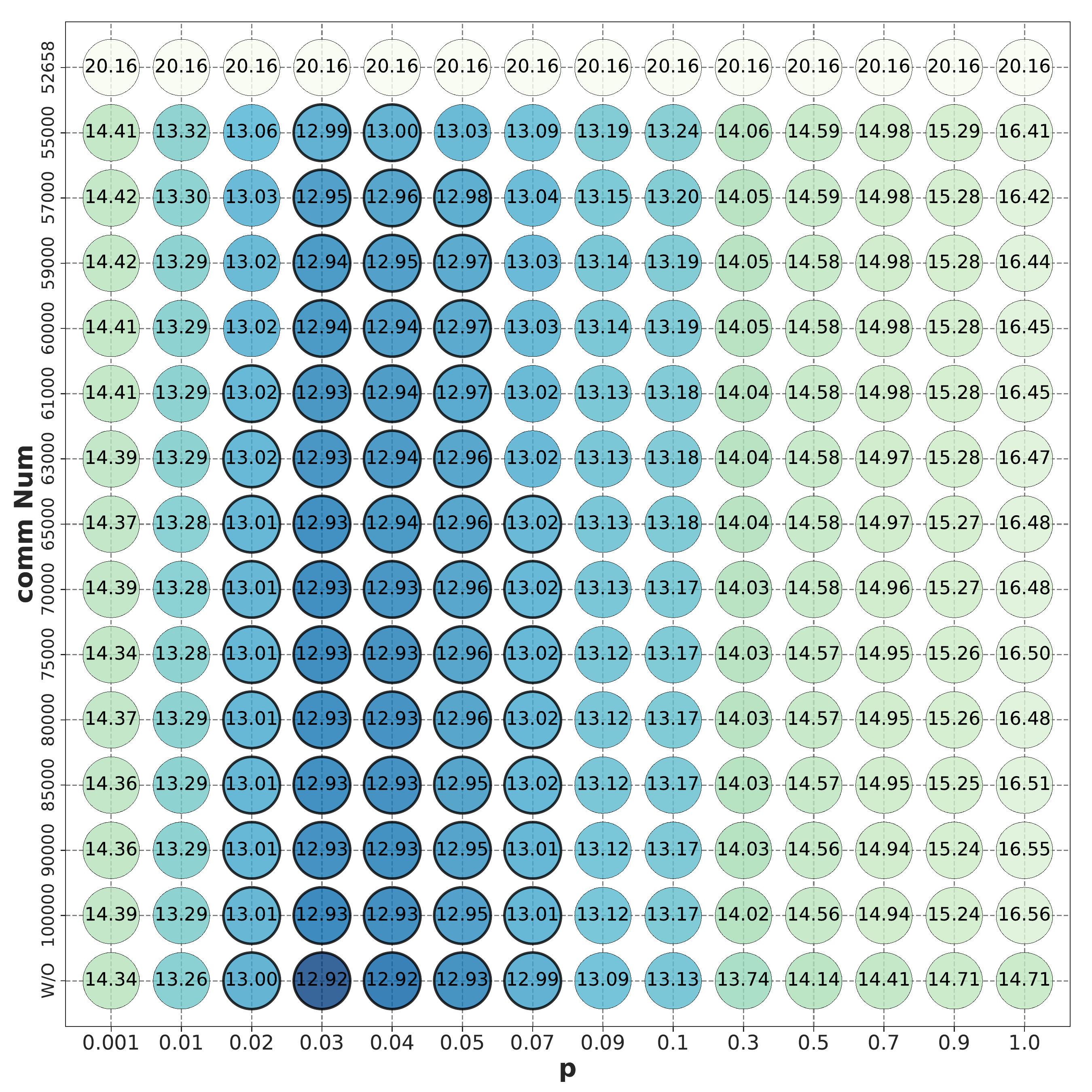}
		\label{fig: SE_hot_products}
	}\hfill
	\subfigure[Histogram of community size on Ogbn-products.]{
		\includegraphics[width=0.49\linewidth]{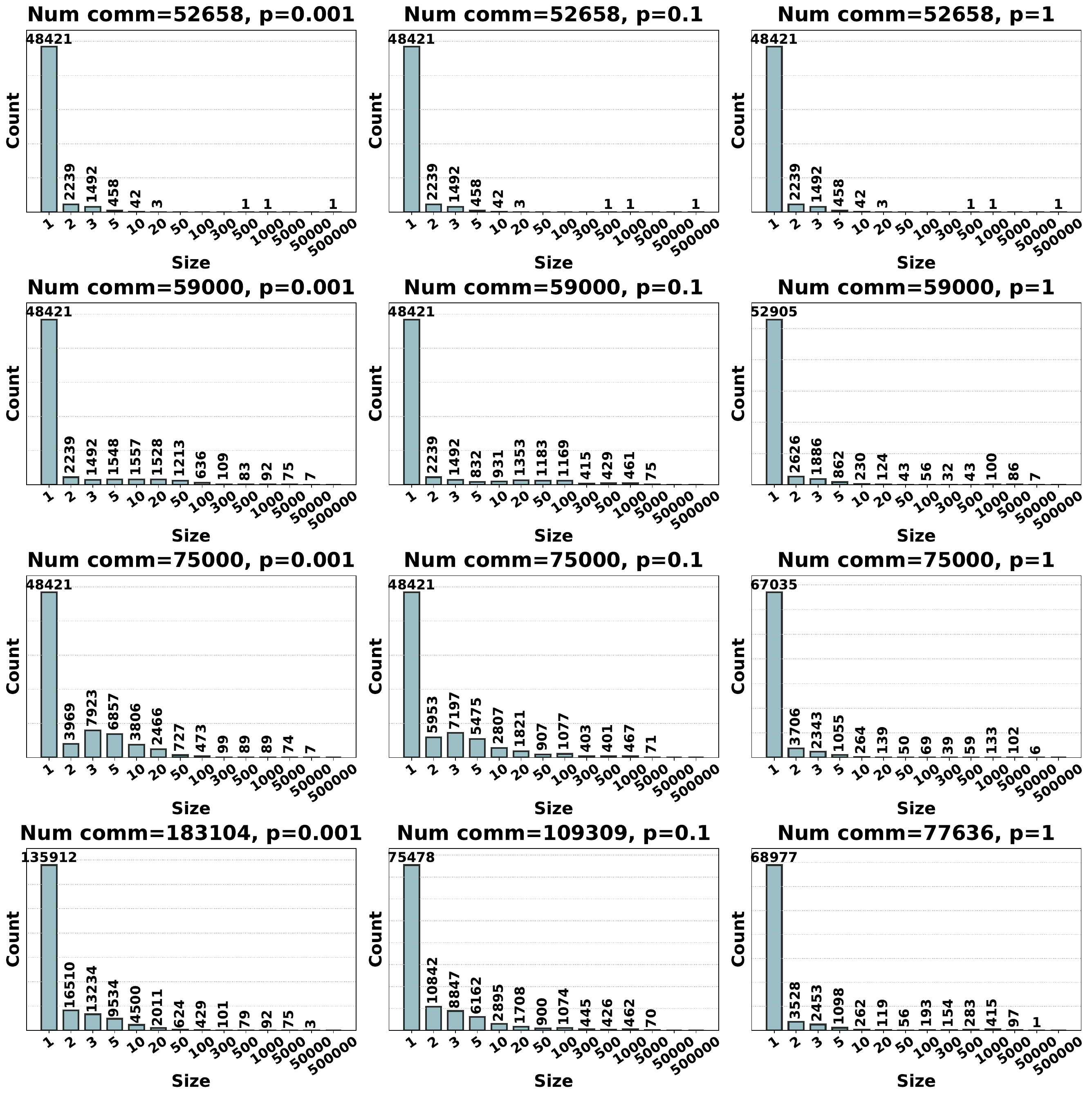}
		\label{fig: SE_bin_products}
	}\hfill
    \vspace{-4mm}
	\caption{Evolution of structural entropy and community distribution on Ogbn-products.}
    \vspace{-4mm}
	\label{fig: case_products}
\end{figure*}

\end{document}